\documentclass[10pt,twocolumn,letterpaper]{article}

\usepackage{cvpr} 

\usepackage[
  pagebackref,
  breaklinks,
  colorlinks,
  bookmarks=false]{hyperref}
\usepackage{microtype}
\usepackage{kotex}
\usepackage{amsmath,amssymb}
\usepackage{cleveref}
\usepackage{xcolor}
\usepackage{multirow}
\usepackage{colortbl}
\usepackage{adjustbox}
\usepackage{tabularx, booktabs}
\usepackage{caption}
\usepackage{subcaption}
\usepackage{bbm}

\usepackage{mathtools}

\definecolor{cblue} {HTML}{1B45F5}
\definecolor{cgreen}{HTML}{489761}
\definecolor{cdarkpurple}{HTML}{242038}
\definecolor{cpostechpink}{HTML}{C80150}
\definecolor{cbrown}{HTML}{7A4419}
\definecolor{c 
red}  {HTML}{E23126}

\newcommand\SP[1]{{\color{magenta}{[SP: #1]}}}

\newcommand{\realnum}{\mathbb{R}}
\newcommand{\batch}{\mathcal{B}}

\newcommand{\mean}{\textnormal{mean}}

\usepackage[linesnumbered,ruled,vlined]{algorithm2e}
\usepackage {algpseudocode}
\usepackage{algorithmicx}
\usepackage{algcompatible}

\usepackage[accsupp]{axessibility}

\usepackage{amsmath,amsfonts,bm}

\newtheorem{theorem}{Theorem}

\def\1{\bm{1}}

\DeclareMathAlphabet{\mathsfit}{\encodingdefault}{\sfdefault}{m}{sl}
\SetMathAlphabet{\mathsfit}{bold}{\encodingdefault}{\sfdefault}{bx}{n}

\def\gD{{\mathcal{D}}}

\def\gI{{\mathcal{I}}}

\def\gX{{\mathcal{X}}}

\DeclareMathOperator*{\argmax}{\arg\max}
\DeclareMathOperator*{\argmin}{\arg\min}

\def\cvprPaperID{14307} 
\def\confName{CVPR}
\def\confYear{2024}

\title{MedBN: Robust Test-Time Adaptation against Malicious Test Samples}

\author{Hyejin Park$^{*}$ \qquad Jeongyeon Hwang$^{*}$ \qquad Sunung Mun \qquad Sangdon Park \qquad Jungseul Ok$^{\dagger}$ \\
Pohang University of Science and Technology (POSTECH), South Korea \\
{\tt\small \{parkebbi2, jeongyeon.hwang, mtablo, sangdon, jungseul\}@postech.ac.kr} \\
}

\date{\today}

\begin{document}

\maketitle

\def\thefootnote{}\footnotetext{$^*$: Equal contribution; {$^\dagger$}: Correspondence to \href{mailto:jungseul@postech.ac.kr}{jungseul@postech.ac.kr}}

\begin{abstract}
Test-time adaptation (TTA) has emerged as a promising solution to address performance decay due to unforeseen distribution shifts between training and test data. While recent TTA methods excel in adapting to test data variations, such adaptability exposes a model to vulnerability against malicious examples.
Indeed, previous studies have uncovered security vulnerabilities within TTA even when a small proportion of the test batch is maliciously manipulated. In response to the emerging threat, we propose median batch normalization (MedBN), leveraging the robustness of the median for statistics estimation within the batch normalization layer during test-time inference. Our method is algorithm-agnostic, thus allowing seamless integration with existing TTA frameworks. 
Our experimental results on benchmark datasets, including CIFAR10-C, CIFAR100-C, and ImageNet-C, consistently demonstrate that MedBN outperforms existing approaches in maintaining robust performance across different attack scenarios, encompassing both instant and cumulative attacks. Through extensive experiments, we show that our approach sustains the performance even in the absence of attacks, achieving a practical balance between robustness and performance.
Our code is available at \href{https://github.com/ml-postech/MedBN-robust-test-time-adaptation}{https://github.com/ml-postech/MedBN-robust-test-time-adaptation}.

\end{abstract}
\vspace{-0.6em}
\section{Introduction}\label{sec:intro}
Deep neural networks (DNNs) have shown noticeable advances in benchmarks across diverse recognition tasks, assuming virtually no distribution shift between training and test data. However, distribution shifts are inevitable in practice mainly due to time-varying environments (e.g., lighting variations and changing weather conditions), and severely degenerate the model performance \cite{hendrycks2018benchmarking, koh2021wilds}. It is infeasible to forecast and prepare for every potential test domain in advance. In response, test-time adaptation (TTA) has been extensively studied \cite{niu2022towards, sun2020test, gong2023sotta, goyal2022test, wang2022continual, chen2022contrastive, dobler2023robust}, where TTA aims at adapting a pre-trained model to test data, which is unlabeled and from latent domain, in an online manner.

\begin{figure}[t!]
     \centering
     \vspace{-0.4em}
     \includegraphics[trim={0 0 0 0}, clip, width=0.425\textwidth]{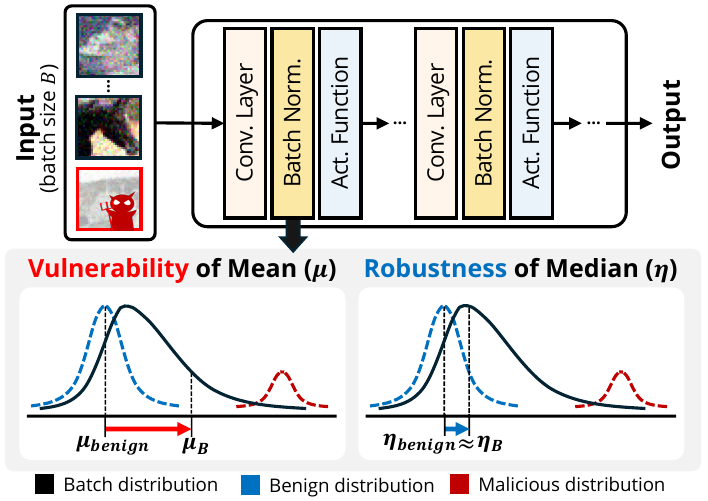}
     \vspace{-0.7em}
    \caption{An illustrative example of the vulnerability of mean in a batch normalization layer to manipulation by malicious sample (left), contrasted with the robustness of median such manipulation (right), when dealing with malicious samples within the batch. }
    \label{fig:illust}
    \vspace{-2.0em}
\end{figure}

The major advantage of TTA stems from leveraging the statistics of the test batch. 
A prominent technique is to use test batch statistics in each batch normalization (BN) layer \cite{nado2020evaluating, schneider2020improving} before adjusting model parameters.
Hence, it is crucial to reliably estimate the test batch statistics and make necessary adjustments. Most of the recent advances have focused on robust estimations of the test batch statistics in a variety of scenarios, including continual distribution shifts  \cite{wang2022continual}, small test batches \cite{koh2021wilds, khurana2021sita}, temporally correlated stream of test data \cite{gong2022note}, and out-of-distribution test data \cite{gong2023sotta}, where the exponential moving averaging (EMA) \cite{yuan2023robust,gong2023sotta} or interpolating source and test statistics \cite{kang2023leveraging, lim2022ttn, wu2023uncovering}
are proposed for robust statistics estimation.

Despite such efforts to build robust TTA methods, recent works \cite{wu2023uncovering, cong2023test} have revealed the vulnerability of TTA methods that use the test batch statistics.
By injecting small portions of malicious samples into the test batch, an adversary can easily manipulate the test batch statistics and also predictions on other (benign) samples, constituting a data poisoning attack.
As we cannot presume the distribution of test samples in the real world, verifying the robustness of TTA methods against the data poisoning attack is essential since it can be considered as a worst-case study.
Although the initial studies have proposed heuristics 
to partially address the vulnerability, it still remains a potential threat, posing a challenge even to state-of-the-art TTA methods.

This paper examines the potential vulnerabilities of existing TTA methods to data poisoning attacks through both theoretical (Section~\ref{sec:mean:med}) and empirical (Section~\ref{sec:vul}) investigations,
including the state-of-the-art techniques \cite{gong2023sotta, niu2022efficient, niu2022towards, wang2020tent}.
Our theoretical analysis reveals that relying on the mean of test batch statistics creates a loophole that adversaries can exploit.
This arises because the mean can be easily manipulated by even a single malicious sample, whereas the median proves to be robust against manipulation by a number of malicious samples, as illustrated in Figure~\ref{fig:illust}.
Furthermore, despite the integration of various modules for enhancing TTA robustness, 
our experiments show that state-of-the-art methods exhibit notable vulnerabilities to malicious samples.

Consequently, to address the adversarial risks in BN updates, we propose {\bf Med}ian {\bf B}atch {\bf N}ormalization (MedBN) method that uses the median for estimating test batch statistics.
Our approach stands out compared to existing defenses \cite{wu2023uncovering, cong2023test}, as the model not only maintains model performance but also successfully defends against data poisoning attacks. Given the substantial vulnerability of state-of-the-art TTA methods \cite{gong2023sotta, niu2022efficient, niu2022towards, wang2020tent} to malicious samples, 
we demonstrate that integrating MedBN into each method consistently improves robustness against malicious samples.

\smallskip
\indent Our main contributions are summarized as follows:
\begin{itemize}
    \vspace{-0.5em}
    \item Inspired by a theoretical analysis comparing mean and median, 
    we propose MedBN, a simple and effective robust batch normalization method,
    which uses the median instead of the mean to estimate the batch statistics.
    We note that our method effortlessly integrates into existing TTA methods without additional training.
    
    \vspace{-0.5em}
    \item  Our experiments show that even sophisticated TTA methods are susceptible to data poisoning attacks, despite extensive efforts to enhance the robustness of TTA. This vulnerability arises from relying on the mean for estimation, which creates a potential loophole exploitable by adversaries. 
    
    \vspace{-0.5em}
    \item The robustness of the proposed MedBN is empirically justified by evaluating it over three standard benchmarks for TTA, seven TTA methods, and four different attack scenarios. 
    Notably, MedBN outperforms comparing methods in robustness under attacks by a significant margin in all considered cases. 
\end{itemize}

\section{Related Works}

\noindent\textbf{Robust test-time adaptation methods.} 
TTA methods have evolved to ensure robust performance under various scenarios in practice, including a single distribution shift in data distribution \cite{wang2020tent}, continual distribution shifts  \cite{wang2022continual}, small batches of test data \cite{koh2021wilds, khurana2021sita},  test data with temporal correlations \cite{gong2022note}, and out-of-distribution test data \cite{gong2023sotta}.
While significant efforts have been devoted to robustifying TTA methods, their robustness against malicious samples at test time has been relatively under-explored.
Recent works \cite{wu2023uncovering, cong2023test} have introduced data poisoning attack methods that generate malicious samples to sabotage TTA and demonstrated the vulnerability of a few TTA baselines \cite{nado2020evaluating, wang2020tent, lee2013pseudo, rusak2021if, goyal2022test}.
In this work, we properly investigate the robustness of various state-of-the-art TTA methods against data poisoning attacks 
and also present a simple yet effective defense mechanism, which can be effortlessly added to most TTA methods.

\smallskip
\noindent{\textbf{Data poisoning attacks and defense mechanisms.}}
There has been an extensive line of work on data poisoning attacks and defenses, but existing defense mechanisms are not applicable to TTA scenarios.
For instance, adversarial training \cite{geiping2021doesn}, a representative method, necessitates access to the training process, making it impractical for TTA where such access is unavailable.
While some studies have proposed defense mechanisms specifically for data poisoning attacks in TTA \cite{wu2023uncovering}, our experiments in Section~\ref{sec:vul} demonstrate that their effectiveness is limited.
In contrast, our proposed method not only outperforms these defenses but also seamlessly integrates with any prior TTA methods.
Additional discussion on related works is presented in Appendix~\ref{appendix:related}.

\section{Preliminary}
\label{sec:prelim}

Let 
$\mathcal{X}$ be a sample space,
and $\mathcal{Y}$ be a label space. 
Let $\gD_{\textnormal{src}} := \{(x_i, y_i) \}_{i\in [N_\textnormal{src}]} \subseteq \mathcal{X} \times \mathcal{Y}$ be the training dataset
of $N_\textnormal{src}$ labeled samples
and $\gX_{\textnormal{test}}= \{x'_i\}_{i \in [N_\textnormal{test}]} \subseteq \mathcal{X}$ be the test dataset of $N_\textnormal{test}$ unlabeled test samples.
A model $f(\cdot; \theta)$ of parameters $\theta$ 
is pre-trained on $\gD_{\textnormal{src}}$, 
while
it predicts a label $y \in \mathcal{Y}$ given a test sample $x \in \gX_{\textnormal{test}}$ in the presence of unknown domain shift. Depending on the context, a model $f$ can output a distribution over labels.

TTA adjusts parameters while processing test data batch by batch where a test batch at time $t$ is denoted by $\batch^t \subseteq \mathcal{X}_\text{test}$.
To address the domain shift, 
TTA methods that involve the adaptation of BN layers focus on adjusting BN layers, e.g., statistics and affine parameters of BN layers.

\smallskip
\noindent{\textbf{Batch normalization layers \cite{ioffe2015batch}.}} 
Noting that adapting parameters of BN layers is effective for TTA \cite{niu2022towards, sun2020test, gong2023sotta, goyal2022test}, we describe the procedure of a BN layer converting input
$z \in \realnum^{B \times C \times H \times W}$ 
to normalized $z' \in \realnum^{B \times C \times H \times W}$,
where $B, C, H,$ and $W$ are 
the dimensions of batch, channel, height, and width, respectively. 
The normalization is performed channel-wisely
with estimated BN statistics $(\hat{\mu}_c,\hat{\sigma}^2_c)$
and learnable affine 
parameters $(\beta_c,\gamma_c)$
as follows: 
\begin{align} \label{eq:BN-transform}
z'_{bchw} = \gamma_{c}
 \cdot \frac{z_{bchw}-\hat{\mu}_{c}}{\sqrt{\hat{\sigma}^{2}_{c}+\varepsilon}}
+\beta_{c} \;, 
\end{align}
where $\varepsilon$ is a small positive constant to avoid divided-by-zero.
In the training, the BN statistics $(\hat{\mu}_c,\hat{\sigma}^2_c)$
are typically estimated by the EMA of the mean and variance of batches from source dataset $\gD_\textnormal{src}$, denoted by $\mu_\textnormal{src}$ and $\sigma^2_\textnormal{src}$, respectively.
Then, for every test batch $\batch^{t}$, a traditional BN layer uses the same statistics $\mu_\textnormal{src}$ and $\sigma^2_\textnormal{src}$
for $\hat{\mu}_c$ and $\hat{\sigma}^2_c$.

\smallskip
\noindent{\textbf{TTA with batch normalization.}} 
To tackle distribution shifts of test samples,
a standard approach is TeBN \cite{nado2020evaluating} that estimates the test BN statistics $(\mu_{c}, \sigma_{c}^{2})$ for $(\hat{\mu}_c, \hat{\sigma}^2_c)$ as follows: 
\vspace{-4mm}
\begin{align} 
    \mu_{c} &= \mean  \left\{z_{bchw} \right\}_{bhw} \;, \; \label{eq:mean}
\text{and}
    \\
    \sigma_{c}^{2} &= \mean  \left\{(z_{bchw} - \mu_c)^2\right\}_{bhw} \label{eq:insta-bn},
\end{align}

\noindent where $z$ is the input to the BN layer given test batch $\batch^t$ 
and we denote $\mean \{z_i\}_{i\in \gI} := \frac{1}{|\gI|} \sum_{i \in \gI} z_i$ is the average of 
$z_i$'s over $i \in \gI$.
TENT \cite{wang2020tent} modulates the affine parameters $(\gamma_c, \beta_c)$ in the BN layer \eqref{eq:BN-transform} using TeBN by minimizing the entropy of model predictions on test samples. This simple strategy achieves excellent performance for distribution shifts and is commonly employed in TTA with adapted BN layers \cite{niu2022towards, sun2020test, gong2023sotta, goyal2022test}. However, it poses an adversarial risk because it adapts the test samples before making predictions, potentially including malicious samples. 
Section~\ref{sec:pf:attack_obj} describes our problem on TTA with malicious samples, followed by our method in Section~\ref{sec:method}.
A detailed explanation of the vulnerability of TeBN is provided in Section~\ref{sec:mean:med}, and a comprehensive analysis of the vulnerabilities in the state-of-the-art TTA methods with BN is presented in Section~\ref{sec:vul}.

\section{Problem Formulation} \label{sec:pf:attack_obj}

We have a batch $\batch^t \subseteq \mathcal{X}_\text{test}$ at time $t$, 
part of which can be maliciously manipulated. We denote the malicious set by $\batch^{t}_{\textrm{mal}}$ and the benign set by $\batch^{t}_{\textrm{ben}}$ such that 
$\batch^t = \batch^{t}_{\textrm{mal}} \cup \batch^{t}_{\textrm{ben}}$.
We denote a tuple of labels of $\batch^{t}$ as $\mathcal{Y}^{t} \subseteq \mathcal{Y}$ (and $\mathcal{Y}^{t}_{\textrm{ben}}$ is similarly defined).
For simplicity, we denote 
a batch of labeled samples by
$\mathcal{Z}^{t}$, i.e.,
$\mathcal{Z}^{t}$ is ${\batch}^{t}$ with corresponding labels in $\mathcal{Y}^{t}$ (and $\mathcal{Z}^{t}_{\textrm{ben}}$ is similarly defined).

Our objective is to find a performant TTA method 
that is robust to malicious samples $\hat{\batch}^{t}_{\textrm{mal}}$, which can be 
maliciously generated by solving the following bi-level optimization:
\begin{equation}
\hat{\batch}^{t}_{\textrm{mal}} = \argmax_{\batch^{t}_{\textrm{mal}}} \mathcal{L}_{\textrm{attack}}(f(\cdot \,; \hat{\theta}(\batch^t)), \mathcal{Y}^t) \;,
\end{equation}
\vspace{-3mm}

\noindent where $\hat{\theta}(\batch^t)$ is updated parameters via the TTA method, i.e., $\hat{\theta}(\batch^t) = \argmin_{\theta}\mathcal{L}_{\textrm{TTA}}(\batch^{t}; \theta)$, and $\mathcal{L}_{\textrm{attack}}$ is an attack objective function.
For the attack objective, we consider both targeted attacks and indiscriminate attacks, as used in \cite{wu2023uncovering}. Solving bi-level optimization exactly is computationally expensive.
However, TTA methods only perform a single-step update on $\theta$ for each $\batch^{t}$, so we can approximate $\hat{\theta}$ as $\theta$, as done in \cite{wu2023uncovering}.
A detailed description of the attack algorithm 
and examples of malicious samples 
are presented in Appendix~\ref{appendix:attack:algo} and \ref{appendix:vis:malicious}, respectively. We confirm that TTA methods are vulnerable to these attacks. The detailed vulnerability of TTA methods can be found in Section~\ref{sec:vul}.              
In the following, we consider two different attack types used to find $\hat\batch_\text{mal}^t$.

\smallskip
\noindent{\textbf{Targeted attack.}}\label{sec:obj:tar}
The goal of a targeted attack is to manipulate $\batch_\text{mal}^t$ fed into the TTA method such that the adapted model predicts a targeted label $y^t_{\textrm{target}}$ on a targeted sample $x^t_{\textrm{target}} \in \batch^{t}_{\textrm{ben}}$ as follows:
\vspace{-.4mm}
\begin{equation}
\hat{\batch}^{t}_{\textrm{mal}} = \argmax_{\batch^{t}_{\textrm{mal}}} - \mathcal{L}_{\text{CE}}(f(x^t_{\textrm{target}}; \hat{\theta}(\batch^t)), y^t_{\textrm{target}})  \;,
\end{equation}
\vspace{-1.4mm}

\noindent where $\mathcal{L}_{\text{CE}}$ is the cross-entropy loss.

\smallskip
\noindent{\textbf{Indiscriminate attack.}}\label{sec:obj:indis}
The objective of an indiscriminate attack is to degrade the performance of benign samples $\batch^{t}_{\textrm{ben}}$
by manipulating $\batch_\text{mal}^t$ as follows:
\vspace{-.4mm}
\begin{equation}
\hat{\batch}^{t}_{\textrm{mal}} = \argmax_{\batch^{t}_{\textrm{mal}}} \sum_{(x, y) \in \mathcal{Z}^{t}_{\textrm{ben}}}\mathcal{L}_{\text{CE}}(f(x; \hat{\theta}(\batch^t)), y) \;.
\end{equation}
\vspace{-1.2mm}

\noindent{\textbf{Adversary's knowledge.}}
We mainly consider a white box attack scenario where an adversary possesses knowledge of a pre-trained model, a TTA algorithm (including defense mechanism), a batch, and even the labels of samples in the batch. 
Our study against such a mighty adversary can be interpreted as a worst-case analysis, while 
we also consider more practicable (yet milder) attack scenarios with limited adversaries' knowledge and adaptive attack which obfuscates defense mechanisms in Appendix~\ref{appendix:extended:attack}.

\section{Methodology}
\label{sec:method}

\begin{figure}[!t]
     \vspace{-0.4em}
     \centering
     \includegraphics[trim={0 0 0 0}, clip, width=0.42\textwidth]{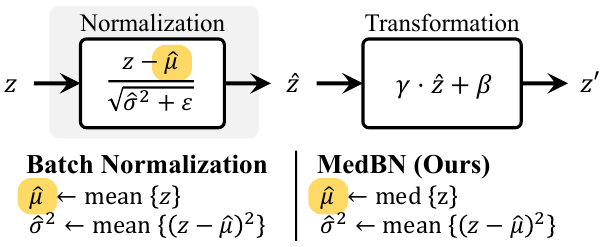}
     \vspace{-0.6em}
     \caption{\textbf{An overview of MedBN.} (Top) TTA methods adapted with BN layers normalize the features ($z$) by estimating normalization statistics $\hat{\mu}$ and $\hat{\sigma}^{2}$, and optimize transformation parameters $\gamma$ and $\beta$.
     (Bottom) In contrast to conventional BN, which computes the statistics based on the mean of inputs, our proposed MedBN utilizes the median value for estimating the statistics, $\hat{\mu}$ and $\hat{\sigma}^{2}$.}
    \label{fig:medbn}      
    \vspace{-1.4em}
\end{figure}

We propose our robust TTA method, Median Batch Normalization (MedBN), followed by its robustness analysis.

\subsection{Median Batch Normalization (MedBN)}\label{sec:method:medbn}
 
Test statistics calculated by mean can be contaminated by data poisoning attacks, as demonstrated by Theorem~\ref{thm:1} in the following section, which in turn, disrupt the model's adaptation and lead to incorrect predictions. To mitigate the effect of malicious samples, we propose a simple approach, called Median Batch Normalization (MedBN). MedBN uses the median instead of the mean for the standardization \eqref{eq:BN-transform} as follows, i.e., $(\eta_{c}, \rho^{2}_{c})$ instead of $({\mu}_{c}, \sigma_{c}^{2})$ for $(\hat{\mu}_{c}, \hat{\sigma}_{c}^{2})$:
\begin{align}
    \eta_c &= \textrm{med}\left\{ z_{bchw} \right\}_{bhw} \;, \;\label{eq:med} \text{and} \\
    \rho_c^2 &= \textrm{mean}\{(z_{bchw} - \eta_c)^2 \}_{bhw} \;,
\end{align}
where $\textrm{med}\{A\} := \min \{a \in A : |\{x \in A: a  > x\}|\ge \frac{|A|}{2}\} $ for a set $A \subseteq \mathbb{R}$.
Here, MedBN standardizes an input $z$ using $(\eta_c, \rho^2_c)$.
Our method is surprisingly effective for the defense against attacks with negligible degradation of model performance. Also, its simplicity allows for easy integration within any existing TTA methods that adjust BN layers.

Note that $\rho_c$ takes the mean of the squared deviations $(z_{bchw} - \eta_c)^2$'s, we can instead take the median of the deviations, which corresponds to the median absolute deviation (MAD), as a part of further robustifying the estimation of BN statistics.
According to our study, 
the use of MAD shows strong defense but a substantial performance drop. 
Hence,
we choose the mean of the squared deviations $(z_{bchw} - \eta_c)^2$'s for our method, see Appendix~\ref{appendix:mad} for results using MAD.

\subsection{Illustrative analysis: mean vs. median}\label{sec:mean:med}

The main idea of our method is to replace the use of means with that of medians when computing BN statistics.
We provide an illustrative analysis comparing the 
robustness of using median instead of mean.

\vspace{-1mm}
\begin{theorem} \label{thm:1}
Consider a set of $n$ numbers $\batch = \{x_i \in \mathbb{R}: i \in [n]\}$ and $1 \le m \le n$ where the first $m$ numbers are possibly manipulated by adversaries. Let $\batch_{\textnormal{mal}} = \{x_i : i \in [m]\}$, and $\batch_{\textnormal{ben}} = \batch \setminus \batch_{\textnormal{mal}}$.

\noindent(i) The mean can be arbitrarily manipulated by a single malicious sample, i.e.,
for any $1 \le m \le n$, 
\vspace{-.8mm}
\begin{align}
\label{eq:meanmean}
\sup_{\batch_\textnormal{mal}} |
\textnormal{mean}(\batch_\textnormal{mal} \cup \batch_\textnormal{ben})
- \textnormal{mean}(\batch_\textnormal{ben})
| &= \infty \;.
\end{align}
\vspace{-1mm}
\noindent(ii) The median is robust against malicious samples unless they are not the majority, i.e.,
for any $1 \le m < n/2$, 
\vspace{-.8mm}
\begin{align}
\sup_{\batch_\textnormal{mal}} |
\textnormal{med}(\batch_\textnormal{mal}\cup \batch_\textnormal{ben})
-\textnormal{med}(\batch_\textnormal{ben})
| &<
\infty \;, \;\text{and}
\label{eq:medmed}
\\
\sup_{\batch_\textnormal{mal}} |
\textnormal{med}(\batch_\textnormal{mal}\cup \batch_\textnormal{ben})
-\textnormal{mean}(\batch_\textnormal{ben})
| &< \infty \;.
\label{eq:medmean}
\end{align}
\end{theorem}
\vspace{-.5mm}

The first part of Theorem~\ref{thm:1}
implies the risk of using mean in the presence of malicious samples. In particular, it says that just a single malicious sample can arbitrarily manipulate
the estimation of mean statistics.
However, as the second part of Theorem~\ref{thm:1}
suggests, such an arbitrary manipulation
by malicious samples is not possible unless the attacker modifies more than half of the batch.
It is noteworthy that the robustness of the median for scalars in Theorem~\ref{thm:1}
can be extended for coordinate-wise or geometric median for vectors as well.
We provide this extension in Appendix~\ref{appendix:multi-dim-median}.





\smallskip
\noindent\textbf{Proof of 
Theorem~\ref{thm:1}.~} 
For
the first part of the vulnerability of mean 
\eqref{eq:meanmean},
we consider a specific choice of $\batch_\textnormal{mal}'$
consisting of $m$-many
$(\textnormal{mean}(\batch_\textnormal{ben}) + L)$'s
for $L \in \mathbb{R}$.
Then, we have

\vspace{-4mm}
\begin{align}
&\sup_{\batch_{\textnormal{mal}}}
|\textnormal{mean}(\batch_{\textnormal{mal}} \cup \batch_\textnormal{ben}) - \textnormal{mean}(\batch_\textnormal{ben}) | \nonumber \\
&\ge
|\textnormal{mean}(\batch_{\textnormal{mal}}' \cup \batch_\textnormal{ben}) - \textnormal{mean}(\batch_\textnormal{ben}) |   = 
\frac{m}{n} L \;,
\end{align}
where the last equality is from the choice of $\batch_\textnormal{mal}'$ such that 
\vspace{-6mm}
\begin{align}
&n \cdot \textnormal{mean}(\batch_{\textnormal{mal}}' \cup \batch_\textnormal{ben})  \nonumber \\
&= 
(n-m) \cdot \textnormal{mean}(\batch_\textnormal{ben}) 
+ m \cdot \textnormal{mean}(\batch_\textnormal{ben}) +
m L \;.
\end{align}

\vspace{-2mm}
\noindent This directly leads to \eqref{eq:meanmean} as the choice of $L$ is arbitrary. 

For the second part on the robustness of median,
we focus on \eqref{eq:medmed}
as the proof of \eqref{eq:medmean} follows similarly.
For \eqref{eq:medmed}, 
let $k = \textrm{med}(\batch_{\textnormal{mal}} \cup \batch_{\textnormal{ben}})$.
If $k \in \batch_{\textnormal{ben}}$, it is trivial. If $k \in \batch_{\textnormal{mal}}$,
given that $1 \le m < n/2$, it follows that $ \min (\batch_{\textnormal{ben}}) \leq k \leq  \max (\batch_{\textnormal{ben}})$. 
Then, $| \textrm{med}(\batch_{\textnormal{mal}} \cup \batch_{\textnormal{ben}}) -\textrm{med}(\batch_{\textnormal{ben}})| \le \max_{x, x' \in \batch_{\textnormal{ben}}} |x - x'| < \infty$. Therefore, this shows \eqref{eq:medmed} and completes the proof of Theorem~\ref{thm:1}.

\begin{figure*}[t!]
     \centering
     \begin{subfigure}[b]{0.31\textwidth}
         \centering
         \includegraphics[width=\textwidth, trim={3.2cm 1.2cm 3.25cm 1.2cm}, clip]{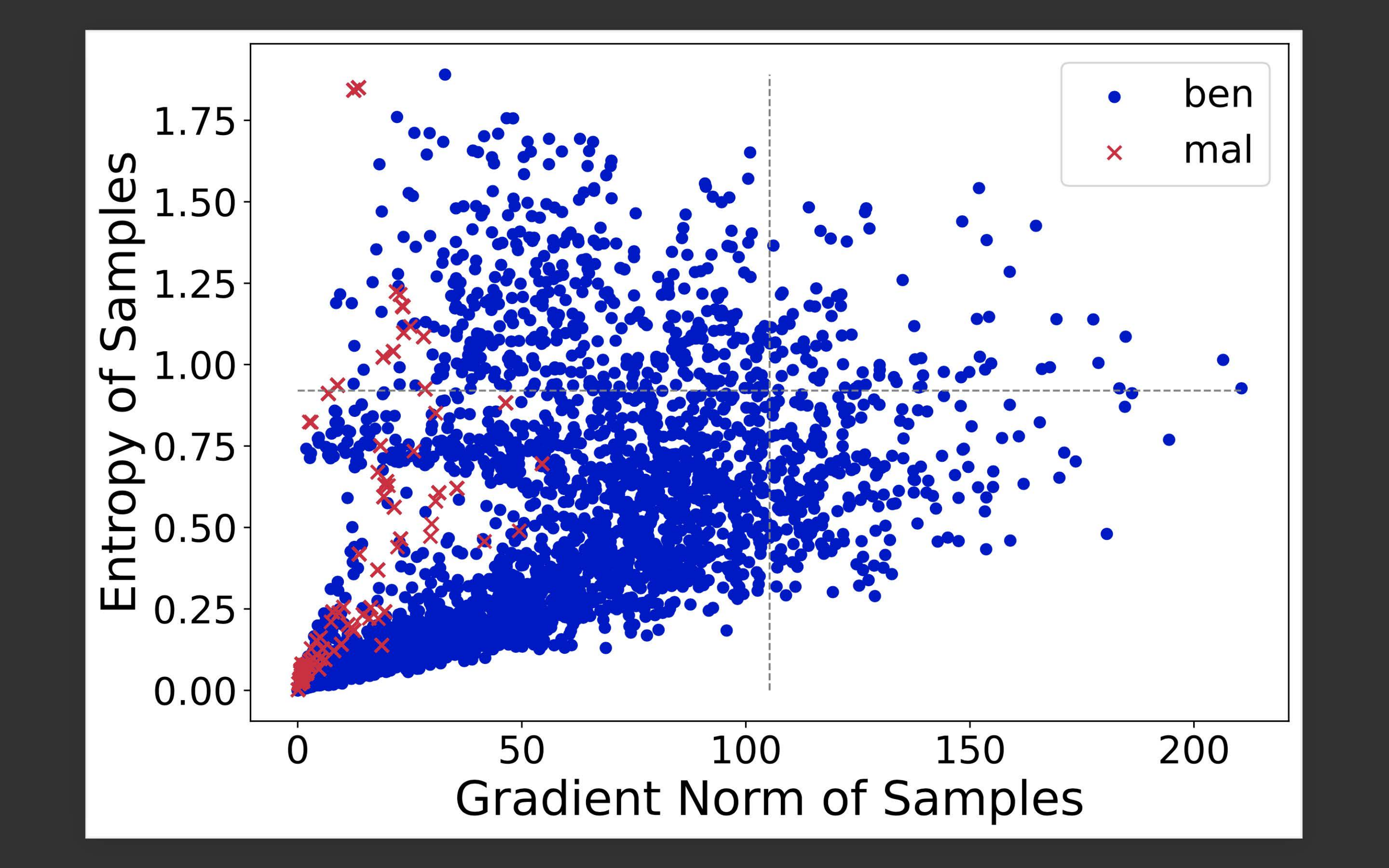}
         \caption{Sample entropy and gradient norm distribution under targeted attack.}\label{fig:vul:target}
     \end{subfigure}
     \hspace{1em}
     \begin{subfigure}[b]{0.31\textwidth}
         \centering
         \includegraphics[width=\textwidth, trim={3.2cm 1.2cm 3.25cm 1.2cm}, clip]{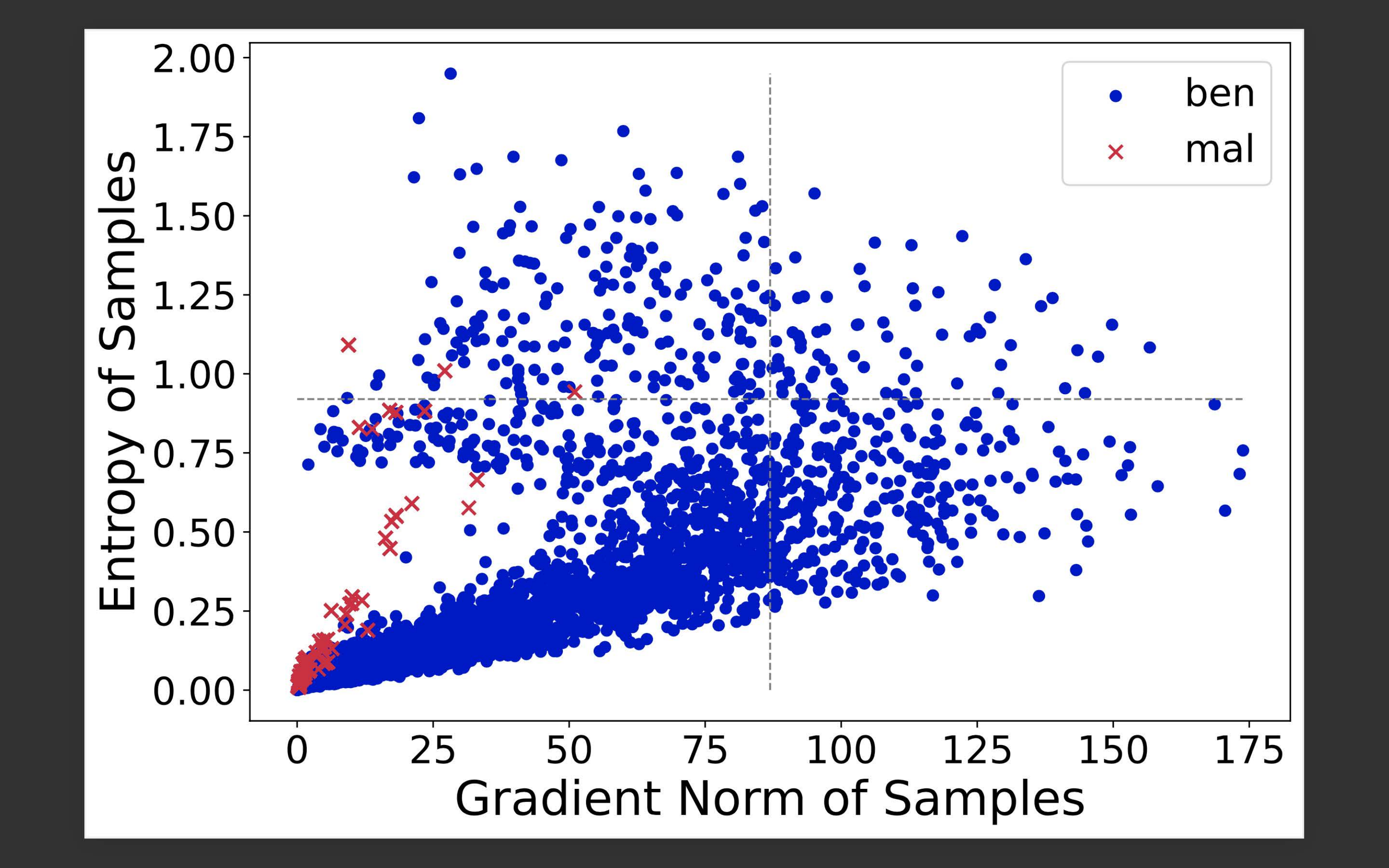}
         \caption{Sample entropy and gradient norm distribution under indiscriminate attack.}\label{fig:vul:indis}
     \end{subfigure}
     \hspace{1em}
     \begin{subfigure}[b]{0.31\textwidth}
         \centering
         \includegraphics[width=\textwidth]{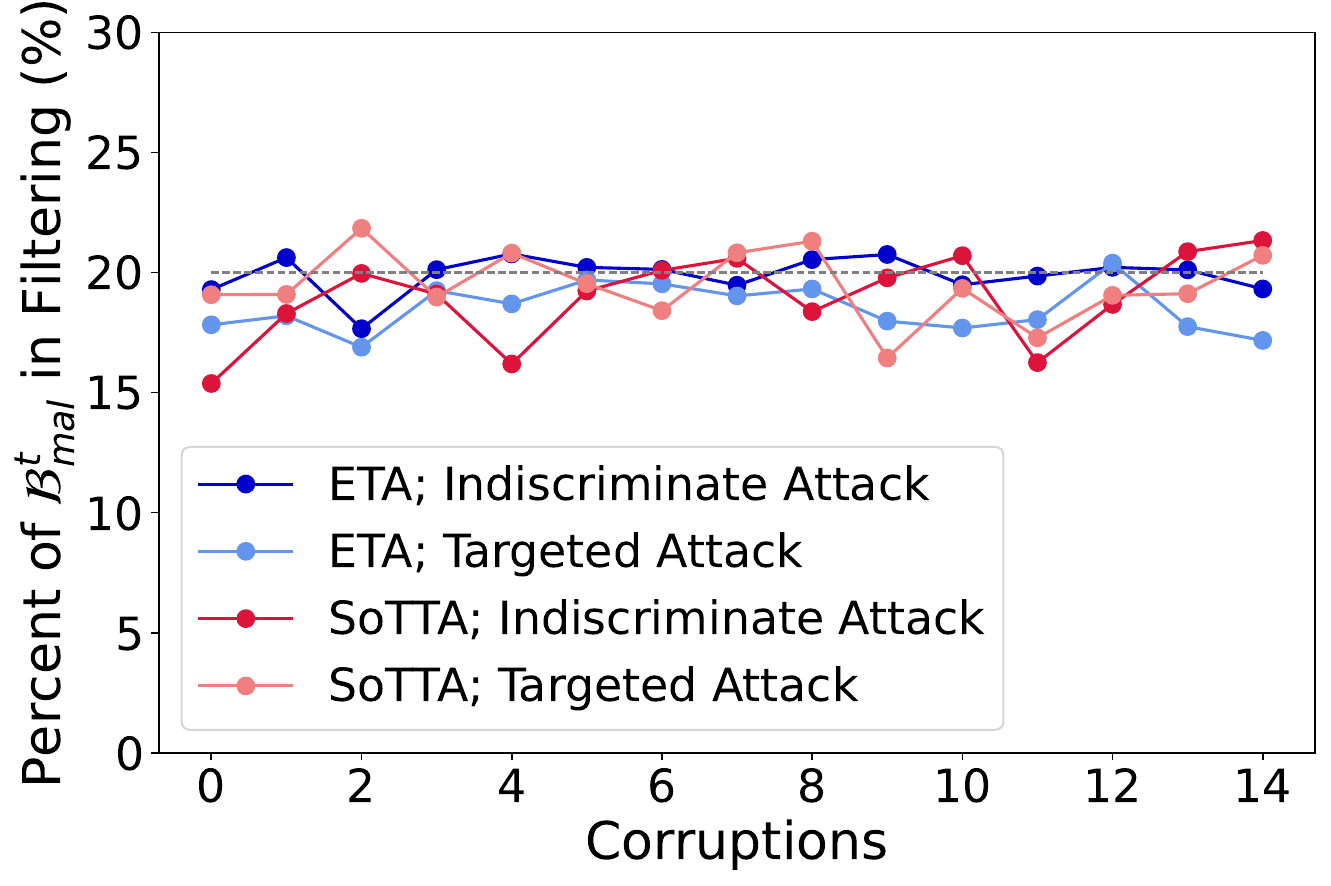}
         \caption{The ratio of malicious samples $\batch_{mal}^{t}$ in filtering over corruptions.}\label{fig:vul:filter}
     \end{subfigure}
     \vspace{-0.6em}
        \caption{Analysis of vulnerability of existing TTA Methods against attacks. Figure~\ref{fig:vul:target}~and~Figure~\ref{fig:vul:indis} represent the relation between entropy and gradient norm of benign and malicious samples in targeted attack and indiscriminate attack, respectively. Figure~\ref{fig:vul:filter} illustrates the proportion of malicious samples $\batch_{mal}$ 
        among the total remaining samples after filtering over the type of corruption, considering an initial condition where 20\% of the samples in the batch were malicious.
        All experiments are performed on CIFAR10-C dataset with Gaussian noise, using a ResNet26, at the highest severity of distribution shift, i.e., level 5.}
        \label{fig:vul}
        \vspace{-1.5em}
\end{figure*}

\section{Vulnerability of Existing TTA Methods against Attacks}\label{sec:vul}

In this section, 
we delve into the effectiveness of TTA methods against malicious samples.
For stabilizing adaptation to test data, many TTA methods propose a variety of modules, including screening out samples to remove noisy ones, optimizing model weights to resist large and noisy gradients, and employing exponential moving averages (EMA) for stable updates of batch normalization statistics.
Hence, we study the influence of these TTA modules against malicious samples across three schemes: (i) filtering, (ii) sharpness-aware learning, and (iii) EMA.

\smallskip
\noindent{\textbf{Filtering scheme.}}
Several research works \cite{niu2022efficient, niu2022towards, gong2023sotta} have proposed the use of filtering modules. The purpose of these modules is to eliminate noisy samples from the adaptation process, based on evaluating the entropy or softmax predictions of model outputs, e.g., screening out samples with high entropy \cite{niu2022efficient, niu2022towards} or low confidence \cite{gong2023sotta}.
By filtering out these potentially problematic samples, the model can be more stably adapted to test data. To identify the malicious samples filtered out by the module using entropy or softmax confidence, we observe the distribution of malicious samples in the entropy-gradient space in two attack scenarios: targeted and indiscriminate attacks with 100 attack steps, a batch size of 200, and 40 malicious samples in each batch.
As illustrated in Figure~\ref{fig:vul:target}~and~Figure~\ref{fig:vul:indis},
malicious samples tend to be clustered with low entropy values, making it challenging to exclude the malicious samples. 
To verify this finding, we investigate the proportion of malicious samples actually filtered out by ETA \cite{niu2022efficient} and SoTTA \cite{gong2023sotta}.
ETA filters samples with high entropy, i.e., $f(x;\theta^t) \log f(x;\theta^t)$, while SoTTA screens out samples with low softmax confidence of model outputs, 
i.e., $\max_{i\in [c]} (e^{f(x;\theta^t)}/\sum_{j=1}^{c} e^{f(x;\theta^t)})$, 
where $c$ denotes the number of classes. 
As shown in Figure~\ref{fig:vul:filter}, we 
observe that malicious samples still exist in the filtered batch (at least $15\%$ of the filtered batch are malicious samples).
Considering that malicious samples constitute $20\%$ of the batch, these results demonstrate that entropy or softmax confidence-based filtering mechanisms are unable to completely remove all malicious samples and allow a high percentage of malicious samples to pass through.



\smallskip
\noindent{\textbf{Sharpness-aware learning scheme.}}
Sharpness-aware learning \cite{niu2022towards}, following Sharpness-Aware Minimization (SAM) \cite{foret2020sharpness}, focuses on the stability of model parameters by guiding them towards a flat minimum in the loss surface. This approach is based on the understanding that a flat minimum is more desirable for model robustness, especially in the presence of noisy or large gradients.
However, as shown in Figure~\ref{fig:vul:target} and Figure~\ref{fig:vul:indis}, the gradient norm of the malicious samples, indicated by the $x$-axis,
is concentrated in regions with small gradients. This indicates that the SAM does not make the model to be robust against malicious samples.


\smallskip
\noindent{\textbf{Exponential moving averages (EMA) scheme.}}
Exponential Moving Averages (EMA) scheme controls the statistics of BN layers, starting with the source statistics ($\mu_{\textrm{src}}$ and $\sigma^{2}_{\textrm{src}}$) from the training phase \cite{singh2019evalnorm, summers2019four}. This differs from approaches like TeBN, which solely rely on test batch statistics. The EMA scheme is defined as follows: 
\vspace{-0.5mm}
\begin{align}
        \hat{\mu}_{t}=\alpha\hat{\mu}_{t-1} + (1-\alpha) \mu_{t} \;, \label{eq:ema:mu}\\
        \hat{\sigma}^2_{t} = \alpha\hat{\sigma}^2_{t-1} + (1-\alpha)
        \sigma^2_{t} \label{eq:ema:sigma} \;,
\end{align}
where $\mu_{0}=\mu_{\textrm{src}}$, $\sigma^2_{0}=\sigma^2_{\textrm{src}}$, and $\alpha \in [0,1]$ is a momentum parameter. 
Leveraging a larger proportion $(\alpha > 0.5)$ of previous statistics $(t-1)$ can mitigate the influence of malicious samples but there exists potential performance degradation of the model to target distribution.
Conversely, utilizing a larger proportion ($\alpha < 0.5$) of current statistics $(t)$ allows for adaptation to the target distribution, but it compromises the robustness against malicious samples. This presents that there is a trade-off requiring strategic consideration for choosing $\alpha$. 


\begin{table*}[t!]\centering
\vspace{-1.2em}
\caption{Attack Success Rate (\%) of the targeted and instant attack scenario.
See Table~\ref{tab:ins:target} in Appendix~\ref{appendix:ins}
for a comprehensive comparison in ASRs over different corruptions. The rightmost column refer the error rates for TeBN without attacks.
See Table~\ref{tab:ins:bf} in 
Appendix~\ref{appendix:ins:bf-attack} for the error rates without attacks of all methods.
}\label{tab:ins:target:avg}
\vspace{-0.3em}
\begin{adjustbox}{width=0.87\textwidth}
\begin{tabular}{c|c|c|ccccccc|c}
\toprule
\multicolumn{1}{l|}{} & \multicolumn{1}{l|}{} & \multicolumn{1}{l|}{} & \multicolumn{7}{c}{Method} & \multicolumn{1}{|c}{$m=0$} \\ \cmidrule{4-11} 
\multicolumn{1}{c|}{\multirow{-2}{*}{Dataset}} & \multicolumn{1}{c|}{\multirow{-2}{*}{$B$ / $m$}} & \multicolumn{1}{c|}{\multirow{-2}{*}{Normalization}} & TeBN & TENT & ETA & SAR & SoTTA & sEMA & mDIA & TeBN (ER \%) \\ 
\midrule
& & BatchNorm & 83.91 & 72.36 & 75.07 & 77.42 & 21.47 & 18.18 & 33.91 & 14.92\\
\multirow{-2}{*}{CIFAR10-C} & \multirow{-2}{*}{\begin{tabular}[c|]{@{}c@{}}200 / 40\\ (20\%)\end{tabular}} & 
\cellcolor[HTML]{EFEFEF}MedBN (Ours) & \cellcolor[HTML]{EFEFEF}\textbf{19.16} & \cellcolor[HTML]{EFEFEF}\textbf{18.36} & \cellcolor[HTML]{EFEFEF}\textbf{18.00} & \cellcolor[HTML]{EFEFEF}\textbf{18.04} & \cellcolor[HTML]{EFEFEF}\textbf{7.82} & \cellcolor[HTML]{EFEFEF}\textbf{8.67} & \cellcolor[HTML]{EFEFEF}\textbf{8.76} & \cellcolor[HTML]{EFEFEF}15.19\\

\midrule
& & BatchNorm & 91.78 & 79.29 & 79.96 & 81.64 & 7.60 & 8.71 & 16.62 & 40.08 \\
\multirow{-2}{*}{CIFAR100-C} & \multirow{-2}{*}{\begin{tabular}[c]{@{}c@{}}200 / 40\\ (20\%)\end{tabular}} & \cellcolor[HTML]{EFEFEF}MedBN (Ours) & \cellcolor[HTML]{EFEFEF}\textbf{2.80} & \cellcolor[HTML]{EFEFEF}\textbf{4.18} & \cellcolor[HTML]{EFEFEF}\textbf{3.02} & \cellcolor[HTML]{EFEFEF}\textbf{3.02} & \cellcolor[HTML]{EFEFEF}\textbf{2.58} & \cellcolor[HTML]{EFEFEF}\textbf{1.60} & \cellcolor[HTML]{EFEFEF}\textbf{2.00} & \cellcolor[HTML]{EFEFEF}40.77\\ 

\midrule
 & & BatchNorm & 97.78 & 91.47 & 94.49 & 64.53 & 15.29 & 11.02 & 32.18 & 66.62 \\
\multirow{-2}{*}{ImageNet-C} & \multirow{-2}{*}{\begin{tabular}[c]{@{}c@{}}200 / 20\\ (10\%)\end{tabular}} & \cellcolor[HTML]{EFEFEF}MedBN (Ours) & \cellcolor[HTML]{EFEFEF}\textbf{0.36} & \cellcolor[HTML]{EFEFEF}\textbf{0.44} & \cellcolor[HTML]{EFEFEF}\textbf{0.44} & \cellcolor[HTML]{EFEFEF}\textbf{0.44} & \cellcolor[HTML]{EFEFEF}\textbf{0.80} & \cellcolor[HTML]{EFEFEF}\textbf{0.27} & \cellcolor[HTML]{EFEFEF}\textbf{1.07} & \cellcolor[HTML]{EFEFEF}69.55 \\
\bottomrule
\end{tabular}
\end{adjustbox}
\vspace{-0.2em}
\end{table*}

\begin{table*}[t!]\centering
\caption{Error Rate (\%) of the indiscriminate and instant attack scenario. 
See 
Table~\ref{tab:ins:indis} in Appendix~\ref{appendix:ins} 
for a comprehensive comparison in ERs over different corruptions. The rightmost column refer the error rates for TeBN without attacks.
See 
Table~\ref{tab:ins:bf} in 
Appendix~\ref{appendix:ins:bf-attack} for the error rates without attacks of all methods.
}\label{tab:ins:indis:avg}
\vspace{-0.3em}
\begin{adjustbox}{width=0.87\textwidth}
\begin{tabular}{c|c|c|ccccccc|c}
\toprule
\multicolumn{1}{l|}{} & \multicolumn{1}{l|}{} & \multicolumn{1}{l|}{} & \multicolumn{7}{c}{Method} & \multicolumn{1}{|c}{$m=0$} \\ \cmidrule{4-11} 
\multicolumn{1}{c|}{\multirow{-2}{*}{Dataset}} & \multicolumn{1}{c|}{\multirow{-2}{*}{$B$ / $m$}} & \multicolumn{1}{c|}{\multirow{-2}{*}{Normalization}} & TeBN & TENT & ETA & SAR & SoTTA & sEMA & mDIA & TeBN (ER \%) \\
\midrule
 & & BatchNorm & 31.02 & 28.13 & 27.42 & 27.56 & 20.40 & 21.65 & 27.96 & 14.92 \\
\multirow{-2}{*}{CIFAR10-C} & \multirow{-2}{*}{\begin{tabular}[c]{@{}c@{}}200 / 40\\ (20\%)\end{tabular}} & \cellcolor[HTML]{EFEFEF}MedBN (Ours) & \cellcolor[HTML]{EFEFEF}\textbf{22.34} & \cellcolor[HTML]{EFEFEF}\textbf{20.30} & \cellcolor[HTML]{EFEFEF}\textbf{19.81} & \cellcolor[HTML]{EFEFEF}\textbf{19.60} & \cellcolor[HTML]{EFEFEF}\textbf{16.49} & \cellcolor[HTML]{EFEFEF}\textbf{17.77} & \cellcolor[HTML]{EFEFEF}\textbf{19.06} & \cellcolor[HTML]{EFEFEF}15.19 \\
\midrule

 & & BatchNorm & 59.80 & 55.10 & 54.45 & 56.40 & 48.33 & 46.89 & 55.43 & 40.08 \\
\multirow{-2}{*}{CIFAR100-C} & \multirow{-2}{*}{\begin{tabular}[c]{@{}c@{}}200 / 40\\ (20\%)\end{tabular}} & \cellcolor[HTML]{EFEFEF}MedBN (Ours) & \cellcolor[HTML]{EFEFEF}\textbf{48.55} & \cellcolor[HTML]{EFEFEF}\textbf{46.96} & \cellcolor[HTML]{EFEFEF}\textbf{46.59} & \cellcolor[HTML]{EFEFEF}\textbf{48.00} & \cellcolor[HTML]{EFEFEF}\textbf{45.38} & \cellcolor[HTML]{EFEFEF}\textbf{43.35} & \cellcolor[HTML]{EFEFEF}\textbf{47.84} &  \cellcolor[HTML]{EFEFEF}40.77 \\ 
\midrule
 & & BatchNorm & 81.46 & 72.82 & 74.15 & 77.74 & 66.05 & 73.21 & 77.28 & 66.62 \\
\multirow{-2}{*}{ImageNet-C} & \multirow{-2}{*}{\begin{tabular}[c]{@{}c@{}}200 / 20\\ (10\%)\end{tabular}} & \cellcolor[HTML]{EFEFEF}{MedBN (Ours)} & \cellcolor[HTML]{EFEFEF}\textbf{69.74} & \cellcolor[HTML]{EFEFEF}\textbf{68.01} & \cellcolor[HTML]{EFEFEF}\textbf{68.47} & \cellcolor[HTML]{EFEFEF}\textbf{69.54} & \cellcolor[HTML]{EFEFEF}\textbf{64.22} & \cellcolor[HTML]{EFEFEF}\textbf{70.22} & \cellcolor[HTML]{EFEFEF}\textbf{69.24} & \cellcolor[HTML]{EFEFEF}69.55 \\
\bottomrule
\end{tabular}
\end{adjustbox}
\vspace{-0.2em}
\end{table*}


\begin{table*}[t!]\centering
\caption{Attack Success Rate (\%) of the targeted and cumulative attack scenario on CIFAR10-C and Error Rate (\%) of the indiscriminate and cumulative attack scenario on CIFAR10-C. See Table~\ref{tab:cum:target} and Table~\ref{tab:cum:indis} in 
Appendix~\ref{appendix:cum:attack} over different corruptions and other TTA benchmarks.} \label{tab:cum:avg}
\vspace{-0.3em}
\begin{adjustbox}{width=0.99\textwidth}
\begin{tabular}{c|c|c|c|ccccccc|c}
\toprule
\multicolumn{1}{l|}{} & \multicolumn{1}{l|}{} & \multicolumn{1}{l|}{} & \multicolumn{1}{l|}{} & \multicolumn{7}{c}{Method} & \multicolumn{1}{|c}{$m=0$} \\ \cmidrule{5-12} 
\multicolumn{1}{c|}{\multirow{-2}{*}{Objective}} &
\multicolumn{1}{c|}{\multirow{-2}{*}{Dataset}} & \multicolumn{1}{c|}{\multirow{-2}{*}{$B$ / $m$}} & \multicolumn{1}{c|}{\multirow{-2}{*}{Normalization}} & TeBN & TENT & EATA & SAR & SoTTA & sEMA & mDIA & TeBN (ER \%) \\
\midrule
 & & & BatchNorm & 84.04 &	74.18 &	75.73 &	76.80 &	21.16 &	16.13 &	34.09 & 14.92 \\
\multicolumn{1}{c|}{\multirow{-2}{*}{\begin{tabular}[c]{@{}c@{}}\textit{Targeted}\\ \textit{Attack}\end{tabular}}} & \multirow{-2}{*}{CIFAR10-C} & \multirow{-2}{*}{\begin{tabular}[c]{@{}c@{}}200 / 40\\ (20\%)\end{tabular}} & \cellcolor[HTML]{EFEFEF}MedBN (Ours) & \cellcolor[HTML]{EFEFEF}\textbf{19.20} & \cellcolor[HTML]{EFEFEF}\textbf{18.80} & \cellcolor[HTML]{EFEFEF}\textbf{21.02} & \cellcolor[HTML]{EFEFEF}\textbf{8.76} & \cellcolor[HTML]{EFEFEF}\textbf{8.13} & \cellcolor[HTML]{EFEFEF}\textbf{8.89} & \cellcolor[HTML]{EFEFEF}\textbf{19.06} & \cellcolor[HTML]{EFEFEF}15.19 \\
\midrule
 & & & BatchNorm & 35.30 & 35.70 & 35.30 & 31.25 & 26.10 &	28.79 & 32.05 & 14.92 \\
\multicolumn{1}{c|}{\multirow{-2}{*}{\begin{tabular}[c]{@{}c@{}}\textit{Indisctiminate}\\ \textit{Attack}\end{tabular}}} & \multirow{-2}{*}{CIFAR10-C} & \multirow{-2}{*}{\begin{tabular}[c]{@{}c@{}}200 / 40\\ (20\%)\end{tabular}} & \cellcolor[HTML]{EFEFEF}MedBN (Ours) & \cellcolor[HTML]{EFEFEF}\textbf{27.22} & \cellcolor[HTML]{EFEFEF}\textbf{25.84} & \cellcolor[HTML]{EFEFEF}\textbf{26.84} & \cellcolor[HTML]{EFEFEF}\textbf{24.29} & \cellcolor[HTML]{EFEFEF}\textbf{22.52} & \cellcolor[HTML]{EFEFEF}\textbf{25.62} & \cellcolor[HTML]{EFEFEF}\textbf{23.96} &  \cellcolor[HTML]{EFEFEF}15.19 \\ 
\bottomrule
\end{tabular}
\end{adjustbox}
\vspace{-1.1em}
\end{table*}

\vspace{-0.1em}
\section{Experiments}\label{sec:exp}
\vspace{-0.1em}

In this section, we provide the results of experimental evaluations of MedBN. 
A detailed description of the experimental setup is presented in Section~\ref{sec:exp:setup}. 
The results on various attack scenarios for both image classification and semantic segmentation are presented in Section~\ref{sec:exp:main}~and~\ref{sec:exp:seg}, respectively.
We investigate the reasons behind the robustness of MedBN against in Section~\ref{sec:exp:why}.
Lastly, Section~\ref{sec:exp:ab} presents an ablation study of hyper-parameters such as the number of malicious samples and the test batch size.
More details of the experiments are provided in Appendix~\ref{appendix:exp:details}.

\subsection{Experimental setup}\label{sec:exp:setup}
\noindent{\textbf{Datasets and model architectures.}
We evaluate our approach using three major benchmarks for TTA \cite{hendrycks2018benchmarking}: CIFAR10-C, CIFAR100-C, and ImageNet-C, which represent perturbed versions of the original CIFAR10, CIFAR100, and ImageNet datasets, respectively.
We use ResNet-26 \cite{he2016deep} for CIFAR10-C and CIFAR100-C experiments, and ResNet-50 \cite{he2016deep} for ImageNet-C experiments. 
The models are pre-trained on clean CIFAR10, CIFAR100, and ImageNet training sets from \cite{croce2021robustbench}, respectively, and then evaluated on the aforementioned corrupted test sets.
We additionally demonstrate the effectiveness of MedBN for various model architectures in Appendix~\ref{appendix:arch}.

\smallskip
\noindent{\textbf{Test-time adaptation baselines.}}\label{sec:exp:base}
We consider seven TTA methods as baselines, 
that update batch statistics or the affine parameters of BN layers. 
Test-time normalization (TeBN) \cite{nado2020evaluating} updates BN statistics for each test batch.
TENT \cite{wang2020tent} updates the affine parameters in BN layers using entropy minimization. 
Efficient anti-forgetting test-time adaptation (EATA) \cite{niu2022efficient} improves a sample-efficient entropy minimization and Fisher regularizer to prevent knowledge loss from pre-trained model. 
ETA denotes EATA without Fisher regularization.
Sharpness-aware and reliable optimization (SAR) \cite{niu2022towards} with BN layers and screening-out test-time adaptation (SoTTA) \cite{gong2023sotta} leverage sample filtering and sharpness-aware minimization \cite{foret2020sharpness} to reduce the negative effects caused by large gradients.
Source-initialized exponential moving average (sEMA) \cite{singh2019evalnorm, summers2019four, yuan2023robust, gong2023sotta} manages BN layers' statistics using EMA with the source statistics from the training phase as the initial value in \eqref{eq:ema:mu} and \eqref{eq:ema:sigma}. We use $\alpha=0.8$ for stable update.
Lastly, mitigating Distribution Invading Attacks (mDIA) \cite{wu2023uncovering} interpolates source statistics and test batch statistics in BN layers, except terminal BN layers.

\smallskip
\noindent{\textbf{Attack scenarios.}}
We consider four different attack scenarios
over two purposes of attacks and two frequencies of attacks. 
In particular, 
\emph{targeted} and \emph{indiscriminate} attacks 
are two purposes of attacks
as outlined in Section~\ref{sec:pf:attack_obj}. 
For each purpose of attack, 
we additionally consider two types of attack: an instant attack and a cumulative attack. 
In \emph{the instant attack scenario}, the attacker injects a set of malicious data into the $t$-th batch 
after adapting to the previous $(t-1)$ benign batches \cite{wu2023uncovering}.
On the other hand, \emph{the cumulative attack scenario} involves an attack across all batches, from the first batch up to $T$-th batch, where
$T$ is the total number of batches.
For the number of malicious samples $m$ per batch, 
we use $40$, $40$, and $20$ for CIFAR10-C, CIFAR100-C, and ImageNet-C, respectively, out of 200 samples in a batch. 

\begin{figure*}[!t]
    \centering
    \begin{subfigure}[b]{0.8\textwidth}
         \centering
        \includegraphics[width=\textwidth]{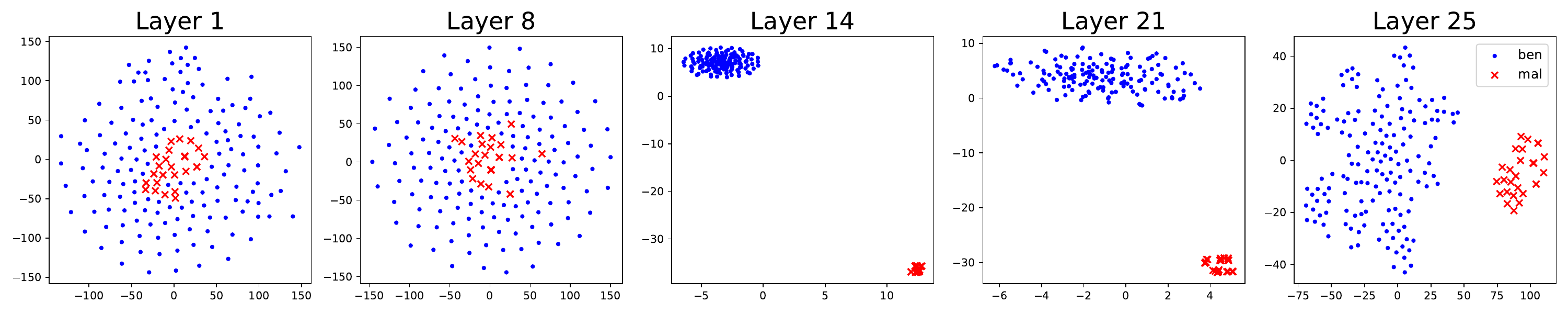}
         \caption{t-SNE visualization of representative BN layers in each block.}\label{fig:tsne:bn}
     \end{subfigure}
    \begin{subfigure}[b]{0.8\textwidth}
         \centering
         \includegraphics[width=\textwidth]{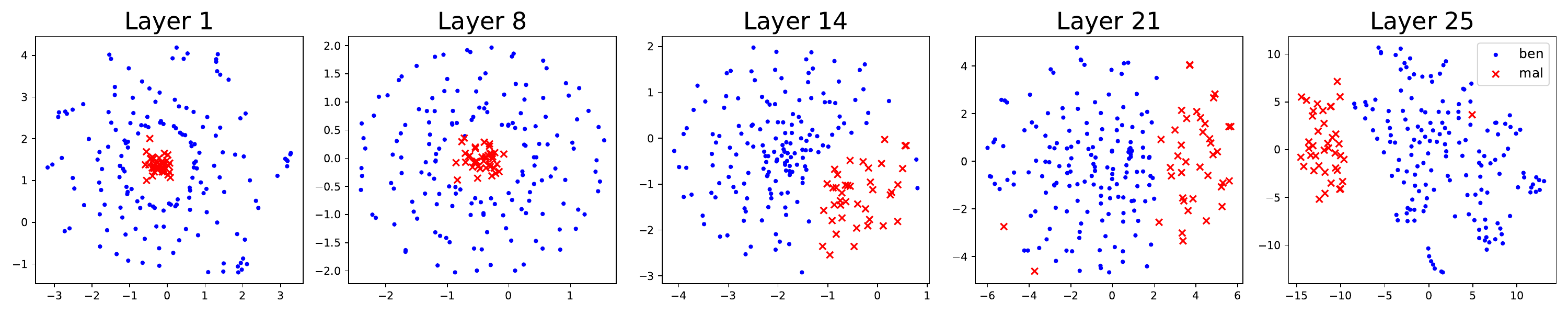}
         \caption{t-SNE visualization of representative MedBN layers in each block.}\label{fig:tsne:med}
     \end{subfigure}
     \vspace{-0.5em}
     \caption{t-SNE visualizations in representative layers of BN and MedBN, across ResNet26 blocks, with benign samples (blue dots) and malicious samples (red crosses).}
    \label{fig:bn}
    \vspace{-0.8em}
\end{figure*}

\begin{figure*}[!t]
    \centering
    \vspace{0.3em}
    \begin{subfigure}[b]{0.45\textwidth}
         \centering
         \includegraphics[width=\textwidth]{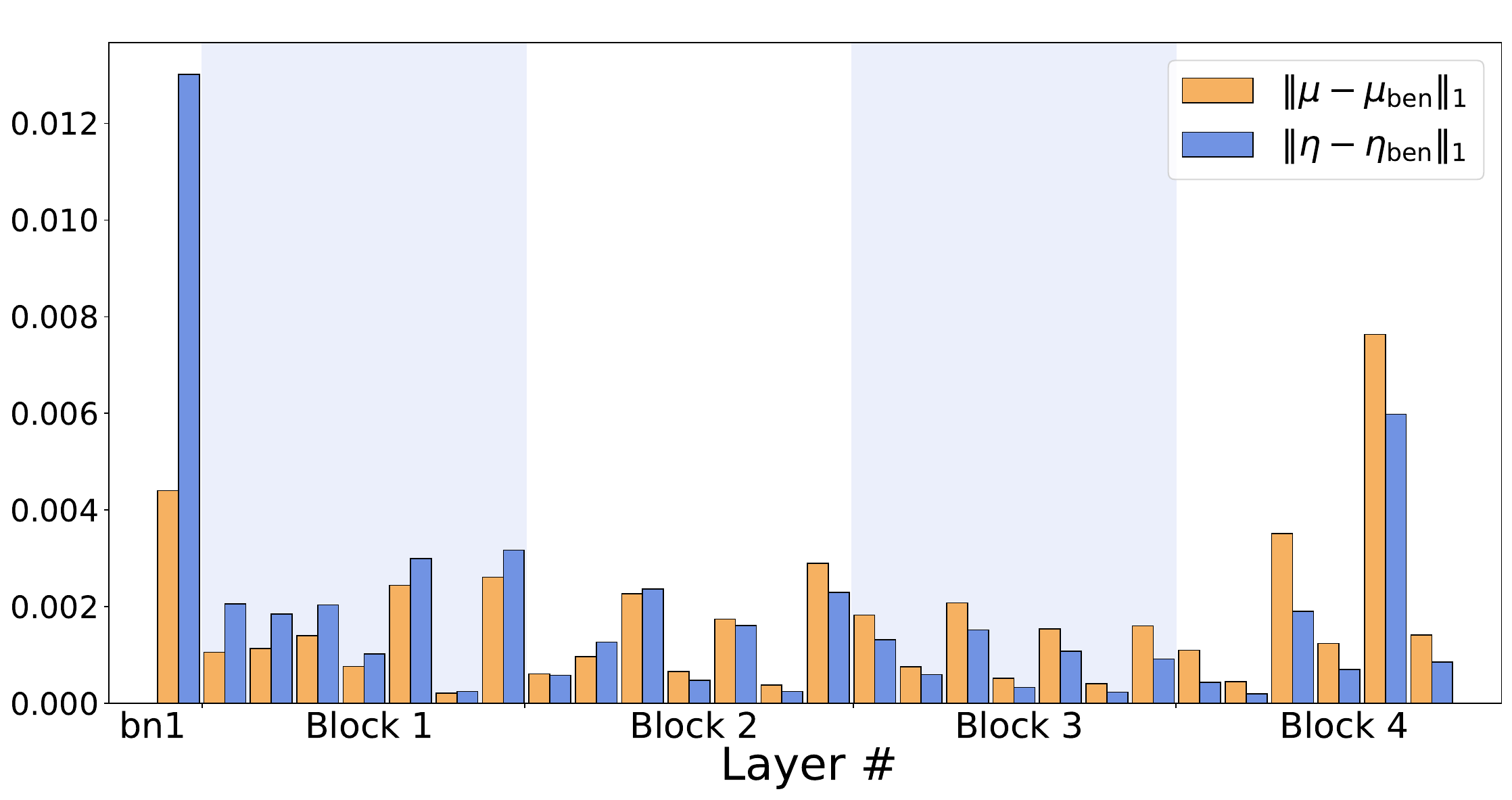}
         \caption{Comparison of $\| \mu - \mu_{\text{ben}}\|_1$ and $\| \eta -\eta_{\text{ben}} \|_1$ across different layers.}
         \label{fig:bn:mean}
     \end{subfigure}
    \hspace{1.0em}
    \begin{subfigure}[b]{0.45\textwidth}
         \centering
        \includegraphics[width=\textwidth]{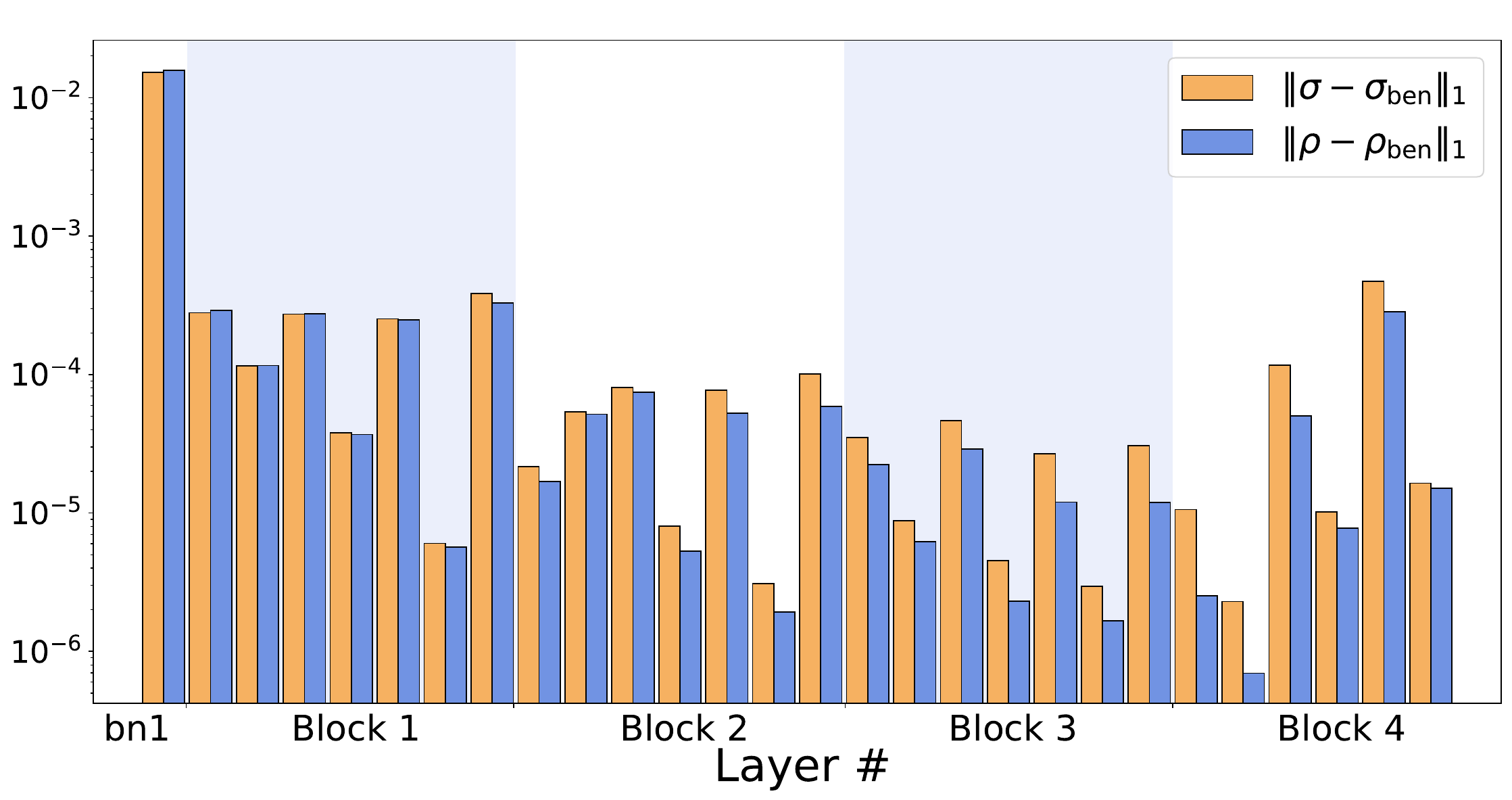}
         \caption{Comparison of $\|  \sigma -\sigma_{\text{ben}} \|_1$ and $\| \rho -\rho_{\text{ben}} \|_1$ across different layers.}
         \label{fig:bn:sigma}
     \end{subfigure}
     \vspace{-0.4em}
    \caption{L1 distance for measuring the amount of perturbation by malicious samples.
    }
    \vspace{-1.0em}
    \label{fig:bn_med_dist}
\end{figure*}

\smallskip
\noindent{\textbf{Evaluation metrics.}}
For the evaluation of targeted attacks, we utilize the metric of Attack Success Rate (ASR), i.e., 
$
\frac{1}{T} \sum_{t=1}^T \mathbbm{1}(f(x^t_\text{target}; \hat{\theta}_{t}) = y^t_\text{target}).
$
The performance of indiscriminate attacks is assessed through the Error Rate (ER) on benign samples after the attack, i.e., 
$
\frac{1}{T} \sum_{t=1}^T \frac{1}{|\mathcal{Z}^t_{\text{ben}}|} \sum_{(x, y) \in \mathcal{Z}^t_{\text{ben}}} \mathbbm{1}(f(x; \hat{\theta}_{t}) \neq y).
$
For each purpose of attack, 
$f(\cdot; \hat{\theta}_{t})$ is an adapted model after each attack using $\mathcal{Z}^t$.
Note that in the instant attack scenario at time $t$, the model $f(\cdot, \hat\theta_{t-1})$ is updated until $(t-1)$ via TTA without any attacks.  
Finally, 
to measure the model's performance under a normal TTA setup, 
we use the standard TTA metric, i.e., 
the ER on benign samples without attacks (i.e., $m=0$).

\subsection{Main results}\label{sec:exp:main}
We demonstrate the efficacy of our method used with seven different TTA algorithms and evaluate three TTA benchmarks under four different attack scenarios.

\smallskip
\noindent{\textbf{The instant attack scenario.}}
Table~\ref{tab:ins:target:avg} and
Table~\ref{tab:ins:indis:avg} demonstrate the effectiveness of MedBN for targeted attacks and indiscriminate attacks, respectively, under the instant attack scenario. 
By simply integrating MedBN into TTA methods with BN layers, it demonstrates significant robustness against malicious samples, i.e., the lower attack success rates under targeted attacks and lower error rates under indiscriminate attacks over all cases, 
but also achieves minimal performance degradation without attacks.

In Table~\ref{tab:ins:target:avg} for targeted attacks, the ASRs of all TTA methods are consistently less than 20\% for CIFAR10-C, 10\% for CIFAR100-C, and 1\% for ImageNet-C. 
While SoTTA and EMA inherently possess defensive capabilities with the use of batch statistics EMA, integrating MedBN further enhances the robustness, yielding the lowest ASRs compared to other methods assessed in this study.
In Table~\ref{tab:ins:indis:avg} for indiscriminate attacks,
the results across all TTA methods indicate that MedBN shows reduced error rates as high as approximately 9\% in CIFAR10-C, 11\% in CIFAR100-C, and 12\% in ImageNet-C. 
As in targeted attacks, it is noteworthy that while SoTTA and sEMA naturally provide some defense with standard BN layers, incorporating MedBN substantially enhances this protection, leading to the lowest error rates observed in all studied methods.

\smallskip
\noindent{\textbf{The cumulative attack scenario.}}
The efficiency of MedBN is indicated in Table~\ref{tab:cum:avg} under the cumulative attack scenario including targeted attacks and indiscriminate attacks on CIFAR10-C. The results on other datasets can be found in Appendix~\ref{appendix:cum:attack}.
Unlike an instant attack, which involves injecting malicious data into a single batch after adapting to previous benign batches, a cumulative attack spreads across all batches. 
Integrating malicious samples consistently throughout the entire dataset can significantly degrade the model, depending on the attacker's goals. Particularly, cumulative attacks have a more pronounced impact in indiscriminate scenarios, where the performance reductions from earlier attacks  can accumulate.
Even in the cumulative attack scenario, MedBN shows lower ASRs in the targeted attack scenario and lower ERs in the indiscriminate attack scenario.

\subsection{Experiments on semantic segmentation}\label{sec:exp:seg}
We expand our experiments to incorporate a semantic segmentation task, examining two instant attack objectives: 
a targeted attack on segmentation, which aims to manipulate the prediction for a targeted pixel within an image, and 
an indiscriminate attack on segmentation, intending to disturb predictions on all the benign samples. Each batch comprises one malicious image and the others benign image with a batch size of 3.
Table~\ref{tab:seg} shows that MedBN effectively defends against both attack scenarios while preserving the mean Intersection over Union (mIoU) on benign images.
Additional experimental details are provided in Appendix~\ref{appendix:exp:details}.

\begin{table}[htb!]
\centering
\vspace{-0.4em}
\caption{Attack Sucess Rate (\%) in instant targeted attack on segmentation and mIoU (\%) on
the benign images in instant indiscriminate attack on segmentation using TeBN to adapt the model trained on Cityscapes \cite{cordts2016cityscapes} for SYNTHIA \cite{ros2016synthia}.} \label{tab:seg}
\begin{adjustbox}{width=0.48\textwidth}
\begin{tabular}{c|c|c|c}
\toprule
Objective & Normalization & TeBN & mIoU (\%) ($m=0$) \\ 
\midrule
\textit{Targeted Attack} & BatchNorm & 69.17 & 25.43 \\
(ASR $\downarrow$)  & \cellcolor[HTML]{EFEFEF}MedBN (Ours) & \cellcolor[HTML]{EFEFEF}\textbf{0.00} & \cellcolor[HTML]{EFEFEF}24.24  \\
\midrule
\textit{Indiscriminate Attack} & BatchNorm & 17.11 & 25.43 \\
(mIoU $\uparrow$) & \cellcolor[HTML]{EFEFEF}MedBN (Ours) & \cellcolor[HTML]{EFEFEF}\textbf{21.55} & \cellcolor[HTML]{EFEFEF}24.24  \\
\bottomrule
\end{tabular}
\end{adjustbox}
\vspace{-0.7em}
\end{table}

\subsection{Why is MedBN robust against attacks?}\label{sec:exp:why}
We investigate how MedBN counteracts the effects of malicious samples. First, we plot the t-SNE of features for each block before going through BN layers
during the adaptation using TeBN on Gaussian corruptions in CIFAR10-C. The t-SNE for all BN layers can be found in Appendix~\ref{appendix:t-sne}.
In Figure~\ref{fig:tsne:bn}, except for the early layers, malicious samples become outliers compared to benign ones. Therefore, as demonstrated in Theorem~\ref{thm:1}, the mean is exposed to be contaminated by these malicious samples and results in the misbehavior of the model. However, when we plot the same t-SNE for MedBN layers, we observe that the malicious samples are closed from the benign samples as shown in Figure~\ref{fig:tsne:med}, i.e., the effect of malicious samples is significantly mitigated.
For the early layers that capture low-level features \cite{cadena2018diverse, olah2017feature, zeiler2014visualizing}, the features of malicious samples are close to those of benign samples, since the malicious samples are generated by adding the imperceptible noise,
making them similar to the benign samples.
However, for deeper layers, the malicious samples tend to go distant from the benign samples to mislead the model.
Secondly, to verify the robustness of MedBN, we measure the $L_1$ distance $\| \mu - \mu_{\text{ben}}\|_1$ and $\| \eta -\eta_{\text{ben}} \|_1$, $\|  \sigma -\sigma_{\text{ben}} \|_1$ and $\| \rho -\rho_{\text{ben}} \|_1$. As shown in Figure~\ref{fig:bn_med_dist}, as the layer gets deeper, the influence of perturbation by malicious samples is smaller for MedBN statistics than BN statistics. These results align with Theorem~\ref{thm:1} and the results of t-SNE for BN and MedBN.

\subsection{Ablation studies} \label{sec:exp:ab}


We perform ablation studies on four distint cases, using CIFAR10-C and CIFAR100-C datasets in targeted and indiscriminate attack scenarios, varying malicious samples and test batch size.
Our focus is on evaluating the TeBN method, as it addresses the vulnerabilities related to robustly estimating BN statistics while excluding learnable parameters.

\smallskip
\noindent{\textbf{The number of malicious samples.}}
We investigate the robustness of the MedBN against various ratios of malicious samples with batch size of 200.
Across various malicious ratios, MedBN is consistently robust 
under targeted attacks in CIFAR10-C (Table~\ref{tab:ab:mal:tar:v2}).
The remaining three cases with results are provided in Appendix~\ref{appendix:ab}.

\begin{table}[htb!]
\vspace{-0.5em}
\centering
\caption{Attack Success Rate (\%) of targeted and instant attack for different numbers of malicious samples with batch size of 200.} 
\vspace{-0.4em}
\label{tab:ab:mal:tar:v2}
\begin{adjustbox}{width=0.48\textwidth}
\begin{tabular}{c|c|ccccc}
\toprule
&  & \multicolumn{5}{c}{\# of Malicious Samples ($m$)} \\ \cmidrule{3-7}
\multicolumn{1}{c|}{\multirow{-2}{*}{Dataset}} & \multicolumn{1}{c|}{\multirow{-2}{*}{Normalization}} & 10 & 20 & 40 & 60 & 80 \\ 
\midrule
\multirow{2}{*}{CIFAR10-C} & BatchNorm & 21.60 & 42.00 & 84.00 & 96.67 & 99.47 \\
&\cellcolor[HTML]{EFEFEF}MedBN (Ours) & \cellcolor[HTML]{EFEFEF}\textbf{7.07} & \cellcolor[HTML]{EFEFEF}\textbf{10.27} & \cellcolor[HTML]{EFEFEF}\textbf{19.20} & \cellcolor[HTML]{EFEFEF}\textbf{26.80}  & \cellcolor[HTML]{EFEFEF}\textbf{38.27} \\
\bottomrule
\end{tabular}
\end{adjustbox}
\vspace{-.3em}
\end{table}


\noindent{\textbf{Test batch size.}}
We explore the effect of different batch sizes.
In all cases, the ratio of malicious samples is about 20\%.
For targeted attacks in CIFAR10-C (Table~\ref{tab:ab:batch:tar:v2}), MedBN achieves significantly lower ASR than BN at all batch sizes.
Note that as the batch size gets smaller, a successful attack gets more difficult because there is less malicious data.
The results for the remaining three cases are included in Appendix~\ref{appendix:ab}.

\begin{table}[htb!]
\centering
\vspace{-0.5em}
\caption{Attack Success Rate (\%) of targeted and instant attack for different batch size $B$ with a consistent 20\% of malicious samples.} \label{tab:ab:batch:tar:v2}
\vspace{-0.4em}
\begin{adjustbox}{width=0.48\textwidth}
\begin{tabular}{c|c|ccccc}
\toprule
&  & \multicolumn{5}{c}{Batch-size ($B$)} \\ \cmidrule{3-7}
\multicolumn{1}{c|}{\multirow{-2}{*}{Dataset}} & \multicolumn{1}{c|}{\multirow{-2}{*}{Normalization}} & 200 & 128 & 64 & 32 & 16 \\ 
\midrule
\multirow{2}{*}{CIFAR10-C} & BatchNorm & 83.91 & 87.76 & 84.84 & 83.87 & 84.60 \\
&\cellcolor[HTML]{EFEFEF}MedBN (Ours) & \cellcolor[HTML]{EFEFEF}\textbf{19.16} & \cellcolor[HTML]{EFEFEF}\textbf{20.51} & \cellcolor[HTML]{EFEFEF}\textbf{17.83} & \cellcolor[HTML]{EFEFEF}\textbf{20.19} & \cellcolor[HTML]{EFEFEF}\textbf{29.14} \\
\bottomrule
\end{tabular}
\end{adjustbox}
\vspace{-.71em}
\end{table}


\section{Conclusion}

We provide a comprehensive study disclosing potential threats of existing TTA methods mainly due to their vulnerable estimation of BN statistics despite the remarkable advances in TTA.
Hence, we propose MedBN, an simple yet effective robust batch normalization method against malicious samples, which  can be effortlessly combined with most of the existing TTA methods if BN layers are being adapted.
Our comprehensive experiments demonstrate the robustness and general applicability of MedBN.
In particular, we show that applying MedBN to other methods results in significant performance improvements, implying that MedBN helps attain outstanding robustness.
For example, applying MedBN to SoTTA (one of the state-of-the-art) shows the best robustness across all benchmarks.
We believe that our work can provide a general robust batch normalization for future work. 


\vspace{.3em}
\smallskip
\noindent{\textbf{Acknowledgements.}}
This work was supported by Institute of Information \& communications Technology Planning \& Evaluation (IITP) grant funded by the Korea government (MSIT); (No. 2019-0-01906, Artificial Intelligence Graduate School Program (POSTECH)), (No. 2021-0-00739, Development of Distributed/Cooperative AI based 5G+ Network Data Analytics Functions and Control Technology), and (No. 2021-0-02068, Artificial Intelligence Innovation Hub).

\bibliographystyle{ieee_fullname}
\bibliography{egbib}

\begin{thebibliography}{10}\itemsep=-1pt

\bibitem{baruch2019little}
Gilad Baruch, Moran Baruch, and Yoav Goldberg.
\newblock A little is enough: Circumventing defenses for distributed learning.
\newblock {\em Advances in Neural Information Processing Systems}, 32:8632--8642, 2019.

\bibitem{biggio2012poisoning}
Battista Biggio, Blaine Nelson, and Pavel Laskov.
\newblock Poisoning attacks against support vector machines.
\newblock In {\em International Conference on Machine Learning}, pages 1467--1474. PMLR, 2012.

\bibitem{cadena2018diverse}
Santiago~A Cadena, Marissa~A Weis, Leon~A Gatys, Matthias Bethge, and Alexander~S Ecker.
\newblock Diverse feature visualizations reveal invariances in early layers of deep neural networks.
\newblock In {\em Proceedings of the European Conference on Computer Vision (ECCV)}, pages 217--232, 2018.

\bibitem{cai2020tinytl}
Han Cai, Chuang Gan, Ligeng Zhu, and Song Han.
\newblock Tinytl: Reduce memory, not parameters for efficient on-device learning.
\newblock {\em Advances in Neural Information Processing Systems}, 33:11285--11297, 2020.

\bibitem{carlini2021poisoning}
Nicholas Carlini and Andreas Terzis.
\newblock Poisoning and backdooring contrastive learning.
\newblock In {\em International Conference on Learning Representations}, 2021.

\bibitem{chen2022contrastive}
Dian Chen, Dequan Wang, Trevor Darrell, and Sayna Ebrahimi.
\newblock Contrastive test-time adaptation.
\newblock In {\em Proceedings of the IEEE/CVF Conference on Computer Vision and Pattern Recognition}, pages 295--305, 2022.

\bibitem{chen2017deeplab}
Liang-Chieh Chen, George Papandreou, Iasonas Kokkinos, Kevin Murphy, and Alan~L Yuille.
\newblock Deeplab: Semantic image segmentation with deep convolutional nets, atrous convolution, and fully connected crfs.
\newblock {\em IEEE Transactions on Pattern Analysis and Machine Intelligence}, 40(4):834--848, 2017.

\bibitem{chen2017distributed}
Yudong Chen, Lili Su, and Jiaming Xu.
\newblock Distributed statistical machine learning in adversarial settings: Byzantine gradient descent.
\newblock {\em Proceedings of the ACM on Measurement and Analysis of Computing Systems}, 1(2):1--25, 2017.

\bibitem{choi2021robustnet}
Sungha Choi, Sanghun Jung, Huiwon Yun, Joanne~T Kim, Seungryong Kim, and Jaegul Choo.
\newblock Robustnet: Improving domain generalization in urban-scene segmentation via instance selective whitening.
\newblock In {\em Proceedings of the IEEE/CVF Conference on Computer Vision and Pattern Recognition}, pages 11580--11590, 2021.

\bibitem{choi2022improving}
Sungha Choi, Seunghan Yang, Seokeon Choi, and Sungrack Yun.
\newblock Improving test-time adaptation via shift-agnostic weight regularization and nearest source prototypes.
\newblock In {\em Proceedings of the European Conference on Computer Vision (ECCV)}, pages 440--458. Springer, 2022.

\bibitem{cong2023test}
Tianshuo Cong, Xinlei He, Yun Shen, and Yang Zhang.
\newblock Test-time poisoning attacks against test-time adaptation models.
\newblock In {\em 2024 IEEE Symposium on Security and Privacy (SP)}, pages 72--72. IEEE Computer Society, 2023.

\bibitem{cordts2016cityscapes}
Marius Cordts, Mohamed Omran, Sebastian Ramos, Timo Rehfeld, Markus Enzweiler, Rodrigo Benenson, Uwe Franke, Stefan Roth, and Bernt Schiele.
\newblock The cityscapes dataset for semantic urban scene understanding.
\newblock In {\em Proceedings of the IEEE/CVF Conference on Computer Vision and Pattern Recognition}, pages 3213--3223, 2016.

\bibitem{croce2021robustbench}
Francesco Croce, Maksym Andriushchenko, Vikash Sehwag, Edoardo Debenedetti, Nicolas Flammarion, Mung Chiang, Prateek Mittal, and Matthias Hein.
\newblock {RobustBench: a standardized adversarial robustness benchmark}.
\newblock In {\em Thirty-fifth Conference on Neural Information Processing Systems Datasets and Benchmarks Track}, 2021.

\bibitem{dobler2023robust}
Mario D{\"o}bler, Robert~A Marsden, and Bin Yang.
\newblock Robust mean teacher for continual and gradual test-time adaptation.
\newblock In {\em Proceedings of the IEEE/CVF Conference on Computer Vision and Pattern Recognition}, pages 7704--7714, 2023.

\bibitem{el2021collaborative}
El~Mahdi El-Mhamdi, Sadegh Farhadkhani, Rachid Guerraoui, Arsany Guirguis, L{\^e}-Nguy{\^e}n Hoang, and S{\'e}bastien Rouault.
\newblock Collaborative learning in the jungle (decentralized, byzantine, heterogeneous, asynchronous and nonconvex learning).
\newblock {\em Advances in Neural Information Processing Systems}, 34:25044--25057, 2021.

\bibitem{farhadkhani2022byzantine}
Sadegh Farhadkhani, Rachid Guerraoui, Nirupam Gupta, Rafael Pinot, and John Stephan.
\newblock Byzantine machine learning made easy by resilient averaging of momentums.
\newblock In {\em International Conference on Machine Learning}, pages 6246--6283. PMLR, 2022.

\bibitem{foret2020sharpness}
Pierre Foret, Ariel Kleiner, Hossein Mobahi, and Behnam Neyshabur.
\newblock Sharpness-aware minimization for efficiently improving generalization.
\newblock In {\em International Conference on Learning Representations}, 2020.

\bibitem{geiping2021doesn}
J Geiping, L Fowl, G Somepalli, M Goldblum, M Moeller, and T Goldstein.
\newblock What doesn’t kill you makes you robust (er): Adversarial training against poisons and backdoors. corr, 2021.

\bibitem{gong2022note}
Taesik Gong, Jongheon Jeong, Taewon Kim, Yewon Kim, Jinwoo Shin, and Sung-Ju Lee.
\newblock Note: Robust continual test-time adaptation against temporal correlation.
\newblock {\em Advances in Neural Information Processing Systems}, 35:27253--27266, 2022.

\bibitem{gong2023sotta}
Taesik Gong, Yewon Kim, Taeckyung Lee, Sorn Chottananurak, and Sung-Ju Lee.
\newblock Sotta: Robust test-time adaptation on noisy data streams.
\newblock {\em Advances in Neural Information Processing Systems}, 36, 2023.

\bibitem{goyal2022test}
Sachin Goyal, Mingjie Sun, Aditi Raghunathan, and J~Zico Kolter.
\newblock Test time adaptation via conjugate pseudo-labels.
\newblock {\em Advances in Neural Information Processing Systems}, 35:6204--6218, 2022.

\bibitem{guerraoui2018hidden}
Rachid Guerraoui, S{\'e}bastien Rouault, et~al.
\newblock The hidden vulnerability of distributed learning in byzantium.
\newblock In {\em International Conference on Machine Learning}, pages 3521--3530. PMLR, 2018.

\bibitem{gupta2021byzantine}
Nirupam Gupta, Shuo Liu, and Nitin Vaidya.
\newblock Byzantine fault-tolerant distributed machine learning with norm-based comparative gradient elimination.
\newblock In {\em 2021 51st Annual IEEE/IFIP International Conference on Dependable Systems and Networks Workshops (DSN-W)}, pages 175--181. IEEE, 2021.

\bibitem{he2016deep}
Kaiming He, Xiangyu Zhang, Shaoqing Ren, and Jian Sun.
\newblock Deep residual learning for image recognition.
\newblock In {\em Proceedings of the IEEE/CVF Conference on Computer Vision and Pattern Recognition}, pages 770--778, 2016.

\bibitem{hendrycks2018benchmarking}
Dan Hendrycks and Thomas Dietterich.
\newblock Benchmarking neural network robustness to common corruptions and perturbations.
\newblock In {\em International Conference on Learning Representations}, 2018.

\bibitem{hendrycks2019augmix}
Dan Hendrycks, Norman Mu, Ekin~Dogus Cubuk, Barret Zoph, Justin Gilmer, and Balaji Lakshminarayanan.
\newblock Augmix: A simple data processing method to improve robustness and uncertainty.
\newblock In {\em International Conference on Learning Representations}, 2019.

\bibitem{hong2023mecta}
Junyuan Hong, Lingjuan Lyu, Jiayu Zhou, and Michael Spranger.
\newblock Mecta: Memory-economic continual test-time model adaptation.
\newblock In {\em International Conference on Learning Representations}, 2023.

\bibitem{ioffe2015batch}
Sergey Ioffe and Christian Szegedy.
\newblock Batch normalization: Accelerating deep network training by reducing internal covariate shift.
\newblock In {\em International Conference on Machine Learning}, pages 448--456. PMLR, 2015.

\bibitem{kang2023leveraging}
Juwon Kang, Nayeong Kim, Donghyeon Kwon, Jungseul Ok, and Suha Kwak.
\newblock Leveraging proxy of training data for test-time adaptation.
\newblock In {\em International Conference on Machine Learning}, pages 15737--15752. PMLR, 2023.

\bibitem{khurana2021sita}
Ansh Khurana, Sujoy Paul, Piyush Rai, Soma Biswas, and Gaurav Aggarwal.
\newblock Sita: Single image test-time adaptation.
\newblock {\em arXiv preprint arXiv:2112.02355}, 2021.

\bibitem{koh2021wilds}
Pang~Wei Koh, Shiori Sagawa, Henrik Marklund, Sang~Michael Xie, Marvin Zhang, Akshay Balsubramani, Weihua Hu, Michihiro Yasunaga, Richard~Lanas Phillips, Irena Gao, et~al.
\newblock Wilds: A benchmark of in-the-wild distribution shifts.
\newblock In {\em International Conference on Machine Learning}, pages 5637--5664. PMLR, 2021.

\bibitem{lamport2019byzantine}
Leslie Lamport, Robert Shostak, and Marshall Pease.
\newblock The byzantine generals problem.
\newblock In {\em Concurrency: the works of leslie lamport}, pages 203--226. 2019.

\bibitem{lee2013pseudo}
Dong-Hyun Lee et~al.
\newblock Pseudo-label: The simple and efficient semi-supervised learning method for deep neural networks.
\newblock In {\em Workshop on challenges in representation learning, ICML}, volume~3, page 896. Atlanta, 2013.

\bibitem{lim2022ttn}
Hyesu Lim, Byeonggeun Kim, Jaegul Choo, and Sungha Choi.
\newblock Ttn: A domain-shift aware batch normalization in test-time adaptation.
\newblock In {\em International Conference on Learning Representations}, 2022.

\bibitem{liu2021ttt++}
Yuejiang Liu, Parth Kothari, Bastien Van~Delft, Baptiste Bellot-Gurlet, Taylor Mordan, and Alexandre Alahi.
\newblock Ttt++: When does self-supervised test-time training fail or thrive?
\newblock {\em Advances in Neural Information Processing Systems}, 34:21808--21820, 2021.

\bibitem{madry2018towards}
Aleksander Madry, Aleksandar Makelov, Ludwig Schmidt, Dimitris Tsipras, and Adrian Vladu.
\newblock Towards deep learning models resistant to adversarial attacks.
\newblock In {\em International Conference on Learning Representations}, 2018.

\bibitem{nado2020evaluating}
Zachary Nado, Shreyas Padhy, D Sculley, Alexander D'Amour, Balaji Lakshminarayanan, and Jasper Snoek.
\newblock Evaluating prediction-time batch normalization for robustness under covariate shift.
\newblock {\em arXiv preprint arXiv:2006.10963}, 2020.

\bibitem{nelson2008exploiting}
Blaine Nelson, Marco Barreno, Fuching~Jack Chi, Anthony~D Joseph, Benjamin~IP Rubinstein, Udam Saini, Charles Sutton, JD Tygar, and Kai Xia.
\newblock Exploiting machine learning to subvert your spam filter.
\newblock In {\em Proceedings of the 1st Usenix Workshop on Large-Scale Exploits and Emergent Threats}, pages 1--9, 2008.

\bibitem{niu2022efficient}
Shuaicheng Niu, Jiaxiang Wu, Yifan Zhang, Yaofo Chen, Shijian Zheng, Peilin Zhao, and Mingkui Tan.
\newblock Efficient test-time model adaptation without forgetting.
\newblock In {\em International Conference on Machine Learning}, pages 16888--16905. PMLR, 2022.

\bibitem{niu2022towards}
Shuaicheng Niu, Jiaxiang Wu, Yifan Zhang, Zhiquan Wen, Yaofo Chen, Peilin Zhao, and Mingkui Tan.
\newblock Towards stable test-time adaptation in dynamic wild world.
\newblock In {\em International Conference on Learning Representations}, 2022.

\bibitem{olah2017feature}
Chris Olah, Alexander Mordvintsev, and Ludwig Schubert.
\newblock Feature visualization.
\newblock {\em Distill}, 2(11):e7, 2017.

\bibitem{ros2016synthia}
German Ros, Laura Sellart, Joanna Materzynska, David Vazquez, and Antonio~M Lopez.
\newblock The synthia dataset: A large collection of synthetic images for semantic segmentation of urban scenes.
\newblock In {\em Proceedings of the IEEE/CVF Conference on Computer Vision and Pattern Recognition}, pages 3234--3243, 2016.

\bibitem{rusak2021if}
Evgenia Rusak, Steffen Schneider, George Pachitariu, Luisa Eck, Peter Gehler, Oliver Bringmann, Wieland Brendel, and Matthias Bethge.
\newblock If your data distribution shifts, use self-learning.
\newblock {\em arXiv preprint arXiv:2104.12928}, 2021.

\bibitem{schneider2020improving}
Steffen Schneider, Evgenia Rusak, Luisa Eck, Oliver Bringmann, Wieland Brendel, and Matthias Bethge.
\newblock Improving robustness against common corruptions by covariate shift adaptation.
\newblock {\em Advances in Neural Information Processing Systems}, 33:11539--11551, 2020.

\bibitem{shafahi2018poison}
Ali Shafahi, W~Ronny Huang, Mahyar Najibi, Octavian Suciu, Christoph Studer, Tudor Dumitras, and Tom Goldstein.
\newblock Poison frogs! targeted clean-label poisoning attacks on neural networks.
\newblock {\em Advances in Neural Information Processing Systems}, 31:6106--6116, 2018.

\bibitem{singh2019evalnorm}
Saurabh Singh and Abhinav Shrivastava.
\newblock Evalnorm: Estimating batch normalization statistics for evaluation.
\newblock In {\em Proceedings of the IEEE/CVF International Conference on Computer Vision}, pages 3633--3641, 2019.

\bibitem{song2023ecotta}
Junha Song, Jungsoo Lee, In~So Kweon, and Sungha Choi.
\newblock Ecotta: Memory-efficient continual test-time adaptation via self-distilled regularization.
\newblock In {\em Proceedings of the IEEE/CVF Conference on Computer Vision and Pattern Recognition}, pages 11920--11929, 2023.

\bibitem{steinhardt2017certified}
Jacob Steinhardt, Pang Wei~W Koh, and Percy~S Liang.
\newblock Certified defenses for data poisoning attacks.
\newblock {\em Advances in Neural Information Processing Systems}, 30:3520--3532, 2017.

\bibitem{su2016fault}
Lili Su and Nitin~H Vaidya.
\newblock Fault-tolerant multi-agent optimization: optimal iterative distributed algorithms.
\newblock In {\em Proceedings of the 2016 ACM Symposium on Principles of Distributed Computing}, pages 425--434, 2016.

\bibitem{summers2019four}
Cecilia Summers and Michael~J Dinneen.
\newblock Four things everyone should know to improve batch normalization.
\newblock In {\em International Conference on Learning Representations}, 2019.

\bibitem{sun2020test}
Yu Sun, Xiaolong Wang, Zhuang Liu, John Miller, Alexei Efros, and Moritz Hardt.
\newblock Test-time training with self-supervision for generalization under distribution shifts.
\newblock In {\em International Conference on Machine Learning}, pages 9229--9248. PMLR, 2020.

\bibitem{wang2020tent}
Dequan Wang, Evan Shelhamer, Shaoteng Liu, Bruno Olshausen, and Trevor Darrell.
\newblock Tent: Fully test-time adaptation by entropy minimization.
\newblock In {\em International Conference on Learning Representations}, 2020.

\bibitem{wang2022continual}
Qin Wang, Olga Fink, Luc Van~Gool, and Dengxin Dai.
\newblock Continual test-time domain adaptation.
\newblock In {\em Proceedings of the IEEE/CVF Conference on Computer Vision and Pattern Recognition}, pages 7201--7211, 2022.

\bibitem{wu2023uncovering}
Tong Wu, Feiran Jia, Xiangyu Qi, Jiachen~T Wang, Vikash Sehwag, Saeed Mahloujifar, and Prateek Mittal.
\newblock Uncovering adversarial risks of test-time adaptation.
\newblock In {\em International Conference on Machine Learning}, pages 37456--37495. PMLR, 2023.

\bibitem{xie2018generalized}
Cong Xie, Oluwasanmi Koyejo, and Indranil Gupta.
\newblock Generalized byzantine-tolerant sgd.
\newblock {\em arXiv preprint arXiv:1802.10116}, 2018.

\bibitem{xie2020fall}
Cong Xie, Oluwasanmi Koyejo, and Indranil Gupta.
\newblock Fall of empires: Breaking byzantine-tolerant sgd by inner product manipulation.
\newblock In {\em Uncertainty in Artificial Intelligence}, pages 261--270. PMLR, 2020.

\bibitem{yang2022rep}
Li Yang, Adnan~Siraj Rakin, and Deliang Fan.
\newblock Rep-net: Efficient on-device learning via feature reprogramming.
\newblock In {\em Proceedings of the IEEE/CVF Conference on Computer Vision and Pattern Recognition}, pages 12277--12286, 2022.

\bibitem{yin2018byzantine}
Dong Yin, Yudong Chen, Ramchandran Kannan, and Peter Bartlett.
\newblock Byzantine-robust distributed learning: Towards optimal statistical rates.
\newblock In {\em International Conference on Machine Learning}, pages 5650--5659. PMLR, 2018.

\bibitem{yuan2023robust}
Longhui Yuan, Binhui Xie, and Shuang Li.
\newblock Robust test-time adaptation in dynamic scenarios.
\newblock In {\em Proceedings of the IEEE/CVF Conference on Computer Vision and Pattern Recognition}, pages 15922--15932, 2023.

\bibitem{zeiler2014visualizing}
Matthew~D Zeiler and Rob Fergus.
\newblock Visualizing and understanding convolutional networks.
\newblock In {\em Proceedings of the European Conference on Computer Vision (ECCV)}, pages 818--833. Springer, 2014.

\bibitem{zhang2022memo}
Marvin Zhang, Sergey Levine, and Chelsea Finn.
\newblock Memo: Test time robustness via adaptation and augmentation.
\newblock {\em Advances in Neural Information Processing Systems}, 35:38629--38642, 2022.

\end{thebibliography}

\clearpage
\appendix
\onecolumn

\clearpage
\section{Implementation Details of Distribution Invading Attack}\label{appendix:attack:algo}

In our study, we use the Distribution Invading Attack (DIA) in \cite{wu2023uncovering}, with a detailed description found in Algorithm~\ref{alg:attack}.
Specifically, the test batch $\batch^{t}$  
undergoes an update process as outlined in Line~5.
Notice that, unlike the general method of using mean, our approach utilizes median calculations as per \eqref{eq:med} for these BN statistics.
For models with BN layers,
executing Line~6 is optional.
TTA methods typically perform a single-step
update using TTA loss $\mathcal{L}_{\textnormal{TTA}}$ on $\theta$ for each $\batch^{t}$, allowing us to estimate $\hat{\theta}$ to be approximately equal to $\theta$.
In Line~7, the perturbation $\delta_{i-1}$ is updated through projected gradient descent (PGD) \cite{madry2018towards}, where the projection $\Pi_{\varepsilon}$ is used to clip $\delta_i$ within the constraint $\varepsilon$. This process ensures that the images remain valid within the [0, 1] range. $\mathcal{L}_{\textnormal{attack}}$ is replaced by adversary's objectives: targeted attack or indiscriminate attack in Section~\ref{sec:pf:attack_obj}.
After $N$-steps PGD, we get the optimal malicious samples $\hat{\batch}^{t}_{\textrm{mal}} = \batch^{t}_{\textrm{mal}} + \delta_N$.


\vspace{-0.8em}
\begin{algorithm}[htb!]
   \caption{Distribution Invading Attack \cite{wu2023uncovering}}\label{alg:attack}
    \begin{algorithmic}[1]
    \STATE \textbf{Input:} 
    Model $f(\cdot; \theta)$ of parameters $\theta$ which include BN statistics $(\hat{\mu}_{c}, \hat{\sigma}^2_{c})$, 
    test batch $\batch^t = \batch^{t}_{\textrm{mal}} \cup \batch^{t}_{\textrm{ben}}$ at time $t$, a targeted label $y_{\textnormal{target}}^t$ on a targeted sample $x_{\textnormal{target}}^t \in \batch^{t}_{\textrm{ben}}$, 
    learning rate $\eta$ of TTA update, 
    learning rate $\alpha$ of attack, 
    the number of attack steps $N$,
    constraint $\varepsilon$,
    and perturbation $\delta_0$.
    \STATE \textbf{Output:}  
    Perturbed malicious samples $\hat{\batch}^{t}_{\textrm{mal}} = \batch^{t}_{\textrm{mal}} + \delta_N$ 
    \STATE  \textbf{for}  $i=1, 2, \dots,  N$ \textbf{do}:
    \STATE \quad  $\batch^t \leftarrow (\batch^{t}_{\textrm{mal}} + \delta_{i-1}) \cup \batch^{t}_{\textrm{ben}}$
    \STATE \quad $(\hat{\mu}_{c}, \hat{\sigma}^2_{c}) \leftarrow (\hat{\mu}_{c}(\batch^{t}), \hat{\sigma}^2_{c}(\batch^{t}))$ \label{alg1:update}
    \STATE \quad (Optional) $\hat{\theta} \leftarrow  \theta - \eta \cdot \partial \mathcal{L}_{\textrm{TTA}}(\batch^t)/ \partial \theta$ 
    \STATE \quad $\delta_i \leftarrow \Pi_{\varepsilon} (\delta_{i-1} - \alpha \cdot \mathrm{sign}(\nabla_{\delta_{i-1}} \mathcal{L}_{\textrm{attack}}(f(\cdot \,; \hat{\theta}(\batch^t))))$
   \STATE   \textbf{end for}
   \STATE \textbf{return} $\hat{\batch}^{t}_{\textrm{mal}} = \batch^{t}_{\textrm{mal}} + \delta_N$
\end{algorithmic}
\end{algorithm}
\vspace{-0.8em}

As we mentioned in Section~\ref{sec:pf:attack_obj}, we discuss the attack objectives $\mathcal{L}_\textnormal{attack}$ for two types of attacks: targeted attack and indiscriminate.
The targeted attack involves an adversarial input $\hat{\batch}_\text{mal}^t$ to make the model misclassify a specific sample $x^t_{\textrm{target}}$ to incorrect label $y_{\textrm{target}}$, formulated as:
$\hat{\batch}^{t}_{\textrm{mal}} = \argmax_{\batch^{t}_{\textrm{mal}}} - \mathcal{L}_{\text{CE}}(f(x^t_{\textrm{target}}; \hat{\theta}(\batch^t)), y^t_{\textrm{target}})$.
On the other hand, the objective of indiscriminate attack is to reduce the model's accuracy on all benign data by manipulating the adversarial input $\hat{\batch}_\text{mal}^t$, given by: $\hat{\batch}^{t}_{\textrm{mal}} = \argmax_{\batch^{t}_{\textrm{mal}}} \sum_{(x, y) \in \mathcal{Z}^{t}_{\textrm{ben}}}\mathcal{L}_{\text{CE}}(f(x; \hat{\theta}(\batch^t)), y)$.

\section{Extended Attack Scenarios}\label{appendix:extended:attack}
The required knowledge of the white-box attack is excessive but not unattainable. Nevertheless, it is crucial to explore more feasible attacks with constrained knowledge of the adversary. 
Therefore, we consider two additional attack scenarios: the semi-white box attack scenario, in which the adversary has constrained knowledge, and the adaptive attack, in which the adversary adapts its adversarial objective to obfuscate defense mechanisms.

\smallskip
\noindent\textbf{Semi-white-box attack.}
We construct a semi-white-box attack that generates malicious samples using only the initial model parameters, while the system continues to adapt its parameters. This approach is more feasible but weaker than the white-box attack. As indicated in Table~\ref{tab:contrained}, the malicious samples generated by the semi-white-box attacker are comparably toxic to those from the white-box attacker in an instant attack scenario, and our method demonstrates robustness against such attacks.

\smallskip

\noindent\textbf{Adaptive attack.}
Since the adversary is aware of defense mechanisms, it can adapt its adversarial objective to obfuscate them. To verify the robustness of our method against such an adaptive attack, we implement it with an additional regularization term, $|\textnormal{med}(\mathcal{B}_{\textnormal{mal}}) - \textnormal{med}(\mathcal{B}_{\textnormal{ben}})|$, which ensures alignment between the median of malicious samples and that of benign samples. However, as shown in Table~\ref{tab:contrained}, the adaptive attack is weaker than the white-box attack, and our method (MedBN) is still robust against such adaptive attacks.

\begin{table}[htb!]
\caption{Attack Success Rate (\%) of targeted attack with TENT.} \label{tab:contrained}
\centering
\begin{adjustbox}{width=0.5\textwidth}
\begin{tabular}{c|c|c|c}
\toprule
Attack Method & White-box & Semi-white-box  & Adaptive white-box  \\
\midrule
 BatchNorm & 72.36 & 53.73 & 31.87\\
\cellcolor[HTML]{EFEFEF}MedBN (Ours) & \cellcolor[HTML]{EFEFEF}\textbf{18.36} & \cellcolor[HTML]{EFEFEF}\textbf{11.20} & \cellcolor[HTML]{EFEFEF}\textbf{7.47} \\
\bottomrule
\end{tabular}
\end{adjustbox}
\end{table}


\section{Experiment Details}\label{appendix:exp:details}

\smallskip
\noindent{\textbf{Datasets.}}
Three major benchmarks for TTA \cite{hendrycks2018benchmarking} CIFAR10-C, CIFAR100-C, and ImageNet-C.
These benchmarks are designed to measure the robustness of networks in classification tasks. 
Each dataset includes 15 types of corruption and 5 levels of severity. Our evaluation concentrates on the most severe level 5 of corruption. 
The CIFAR10-C and CIFAR100-C datasets contain 10,000 test images with 10 and 100 classes, respectively, and the ImageNet-C dataset contains 5,000 test images with 1,000 classes for each type of corruption. 

\smallskip
\noindent{\textbf{Implementation details.}}
In all experiments, we adapt the Adam optimizer with a learning rate of 0.001 and no weight decay.
For SAR and SoTTA, we use the SAM optimizer with the Adam optimizer.
We follow the baseline papers or official codes to set the hyper-parameters for each TTA method.
For data poisoning attacks, we follow the experimental setting of unconstrained attack in \cite{wu2023uncovering}, which is the most threatening attack.
Specifically, we use attack steps of 100 with an attacking optimization rate $\alpha$ of 1/255, the initial perturbation $\delta_0$ of $0.5$,
and the perturbation constraint $\varepsilon$ of 1.0.

\smallskip
\noindent{\textbf{Details on semantic segmentation task.}} Our experimental setup aligns with prior works \cite{choi2021robustnet, lim2022ttn} on semantic segmentation.
We utilize DeepLabv3+ \cite{chen2017deeplab} with ResNet-101 backbone pre-trained on the Cityscapes training set \cite{cordts2016cityscapes} and evaluate its performance on the validation set of SYNTHIA \cite{ros2016synthia}.
In evaluating targeted attack within the segmentation task, we adopt the metric of Attack Success Rate (ASR), akin to image classification. The performance of indiscriminate attacks is evaluated through the mean Intersection over Union (mIoU) on benign samples after the attack.

\section{Extended Related Works}\label{appendix:related}
\noindent\textbf{Test-time adaptation (TTA).}
TTA has been studied to address the issue of distribution shift between source and target domains during the online testing phase, without altering training phase.
TTA methods can be broadly categorized into three groups on the specific parameters they update within a model: (i) fully-updated TTA; update all parameters of the model, (ii) BN-updated TTA; update only BN parameters of the model, and (iii) meta-updated TTA; update meta networks attached with frozen pre-trained model.
Several studies \cite{dobler2023robust, wang2022continual, chen2022contrastive, liu2021ttt++, choi2022improving} 
have improved performance by updating entire model parameters, which may be impractical when the available memory sizes are limited. 
The majority of fully-updated TTA methods adopt the mean-teacher framework, which largely relies on pseudo-labeling of a more reliable teacher model.
The stability of mean-teacher frameworks in changing environments is attributed to their use of an exponential moving average with various loss functions, such as symmetric cross-entropy.

Since fully-updated TTA methods encompass BN-updated TTA approaches, 
TTA typically involves adapting pre-trained models that include BN layers \cite{ioffe2015batch}, which often struggle with domain shifts at test time due to their reliance on training statistics optimized for the training distribution.
Prior methods in TTA \cite{schneider2020improving, nado2020evaluating} have indicated that adapting BN statistics can effectively mitigate distributional shifts.
Moreover, recent TTA approaches \cite{niu2022towards, gong2023sotta, lim2022ttn} have primarily focused on utilizing normalization statistics directly from the current test input, often in conjunction with self-training techniques, such as entropy minimization \cite{wang2020tent, niu2022efficient, zhang2022memo}.
Meanwhile, in addition to works \cite{hong2023mecta, yang2022rep, cai2020tinytl} focusing on memory efficiency, \cite{song2023ecotta} proposes an architecture that is efficient in terms of memory. This design combines frozen original networks with newly proposed meta networks, requiring an initial warm-up using source data.
To address the adversarial risks in TTA methods if BN layers are being adapted, we propose MedBN method that can be integrated into any existing TTA methods if BN layers are being adapted and demonstrate a theoretical analysis of our method. 
When MedBN is integrated into these methods, they consistently demonstrate robustness against malicious samples.



\smallskip
\noindent{\textbf{Data poisoning attacks and defenses.}}
Data poisoning attacks involve injecting poisoned samples into a dataset, causing the model trained with the poisoned dataset to produce inaccurate results at test time. These attacks pose threats to various machine learning algorithms \cite{nelson2008exploiting, biggio2012poisoning, shafahi2018poison, carlini2021poisoning}.
Furthermore, recently, \cite{wu2023uncovering, cong2023test} suggest the risks of data poisoning attacks in the test-time adaptation process, wherein TTA methods adapt the model at test time.

For defense against data poisoning attacks, \cite{steinhardt2017certified} removes outliers by approximating the upper bounds of loss. This method requires the assumption that the dataset is large enough to approximate the loss. However, for test-time adaptation, the number of test data is insufficient to concentrate statistics of loss, and there are no labels for the test data, which means that this approach is not suitable for adaptation during the test phase. \cite{geiping2021doesn} demonstrates that adversarial training is an effective defense method for data poisoning attacks, enhancing the robustness of models in the training phase. However, in test-time adaptation, access to the training process is restricted, primarily due to privacy concerns related to the training data and the substantial computational resources required for training. Additionally, adversarial training leads to performance degradation in test data.
Due to the aforementioned limitations, adversarial training is infeasible in this context. To address the above limitations of existing defenses, we propose a robust batch normalization method that is not only simple and effective but also universally applicable across any existing TTA methods if BN layers are being adapted.

\smallskip
\noindent{\textbf{Median aggregation for robust distributed learning.}}
The abundance of collected data has led to the emergence of distributed learning frameworks. In such systems, several data owners or workers collaborate to construct a global model, typically employing the widely used distributed stochastic gradient descent (SGD) algorithm with a central server. This server iteratively updates the model parameter estimated 
by aggregating the stochastic gradients calculated by the workers. 
However, this algorithm is susceptible to misbehaving workers, referred to as Byzantine in \cite{lamport2019byzantine}, that may send arbitrarily deceptive gradients to the server, potentially disrupting the learning process \cite{ baruch2019little, xie2020fall, su2016fault}.
To address these issues, extensive researches \cite{chen2017distributed, xie2018generalized, guerraoui2018hidden, yin2018byzantine, el2021collaborative, gupta2021byzantine} have been dedicated to robustly aggregating gradients regardless of Byzantine behavior. 
Among a wide range of aggregation methods, the median is widely used for robust aggregation and its effectiveness has been verified: \cite{chen2017distributed} employs the geometric median for robust aggregation, \cite{xie2018generalized} uses the mean around the median, and \cite{yin2018byzantine} utilizes coordinate-wise median. In terms of robust aggregation, the median can also be applied to robustly aggregate batch statistics against malicious samples. To the best of our knowledge, we are the first to use the median for robustly aggregating batch statistics to defend against malicious samples.

\section{Effectiveness of MedBN across Different Model Architectures} \label{appendix:arch}

In the main text, we have focused on ResNet-26. Beyond ResNet-26, our study includes two additional architectures, which are commonly used in TTA: WideResNet-28 (WRN-28) for CIFAR10-C, as referenced in the RobustBench benchmark \cite{croce2021robustbench}, and ResNext-29 for CIFAR100-C from \cite{hendrycks2019augmix}.
Table~\ref{tab:arch} demonstrates the efficacy of MedBN across various architectures over both attack instant scenarios, indicating that MedBN is independent of specific architectural designs, i.e., architecture-agnostic.

\begin{table*}[htb!]\centering
\caption{Effectiveness of MedBN across various model architectures. We use the batch size of 200 with 40 malicious samples.} \label{tab:arch}
\begin{adjustbox}{width=0.95\textwidth}
\begin{tabular}{c|c|c|c|ccccccc|c}
\toprule
\multicolumn{1}{l|}{} & \multicolumn{1}{l|}{} & \multicolumn{1}{l|}{} & \multicolumn{1}{l|}{} & \multicolumn{7}{c}{Method} & \multicolumn{1}{|c}{$m=0$} \\ \cmidrule{5-12} 
\multicolumn{1}{c|}{\multirow{-2}{*}{Objective}} &
\multicolumn{1}{c|}{\multirow{-2}{*}{Dataset}} & \multicolumn{1}{c|}{\multirow{-2}{*}{Architectures}} & \multicolumn{1}{c|}{\multirow{-2}{*}{Normalization}} & TeBN & TENT & ETA & SAR & SoTTA & sEMA & mDIA & TeBN (ER \%) \\

\midrule
 & & & BatchNorm & 86.67 & 80.53 & 82.00 & 82.00 & 27.47 & 20.53 & 25.87 & 20.43 \\ 
 & \multirow{-2}{*}{CIFAR10-C} & \multirow{-2}{*}{WRN-28} & \cellcolor[HTML]{EFEFEF}Ours (MedBN) & \cellcolor[HTML]{EFEFEF}\textbf{24.53} & \cellcolor[HTML]{EFEFEF}\textbf{23.33} & \cellcolor[HTML]{EFEFEF}\textbf{23.07} & \cellcolor[HTML]{EFEFEF}\textbf{22.27} & \cellcolor[HTML]{EFEFEF}\textbf{9.07} & \cellcolor[HTML]{EFEFEF}\textbf{8.53} & \cellcolor[HTML]{EFEFEF}\textbf{11.87} & \cellcolor[HTML]{EFEFEF}\textbf{21.49} \\  \cmidrule{2-12}
 & & & BatchNorm & 96.67 & 80.00 & 79.73 & 84.00 & 12.93 & 9.20 & 7.87 & 35.56 \\ 
\multicolumn{1}{c|}{\multirow{-4}{*}{\begin{tabular}[c]{@{}c@{}}\textit{Targeted}\\ \textit{Attack}\end{tabular}}} & \multirow{-2}{*}{CIFAR100-C} & \multirow{-2}{*}{ResNext-29} & \cellcolor[HTML]{EFEFEF}Ours (MedBN) & \cellcolor[HTML]{EFEFEF}\textbf{3.07} & \cellcolor[HTML]{EFEFEF}\textbf{2.13} & \cellcolor[HTML]{EFEFEF}\textbf{2.27} & \cellcolor[HTML]{EFEFEF}\textbf{2.13} & \cellcolor[HTML]{EFEFEF}\textbf{1.60} & \cellcolor[HTML]{EFEFEF}\textbf{2.00} & \cellcolor[HTML]{EFEFEF}\textbf{1.07} & \cellcolor[HTML]{EFEFEF}\textbf{37.62} \\ 

\midrule
& & & BatchNorm & 37.30 & 34.35 & 33.70 & 34.20 & 27.69 & 28.45 & 42.95 & 20.43 \\ 
 & \multirow{-2}{*}{CIFAR10-C} & \multirow{-2}{*}{WRN-28} & \cellcolor[HTML]{EFEFEF}Ours (MedBN) & \cellcolor[HTML]{EFEFEF}\textbf{29.63} & \cellcolor[HTML]{EFEFEF}\textbf{27.24} & \cellcolor[HTML]{EFEFEF}\textbf{26.69} & \cellcolor[HTML]{EFEFEF}\textbf{26.96} & \cellcolor[HTML]{EFEFEF}\textbf{23.98} & \cellcolor[HTML]{EFEFEF}\textbf{25.42} & \cellcolor[HTML]{EFEFEF}\textbf{35.21} & \cellcolor[HTML]{EFEFEF}\textbf{21.49} \\  \cmidrule{2-12}
 & & & BatchNorm & 62.35 & 52.02 & 51.14 & 52.75 & 44.44 & 45.93 & 46.90 & 35.56 \\ 
\multicolumn{1}{c|}{\multirow{-4}{*}{\begin{tabular}[c]{@{}c@{}}\textit{Indiscriminate}\\ \textit{Attack}\end{tabular}}} & \multirow{-2}{*}{CIFAR100-C} & \multirow{-2}{*}{ResNext-29} & \cellcolor[HTML]{EFEFEF}Ours (MedBN) & \cellcolor[HTML]{EFEFEF}\textbf{43.81} & \cellcolor[HTML]{EFEFEF}\textbf{39.15} & \cellcolor[HTML]{EFEFEF}\textbf{39.32} & \cellcolor[HTML]{EFEFEF}\textbf{40.20} & \cellcolor[HTML]{EFEFEF}\textbf{37.47} & \cellcolor[HTML]{EFEFEF}\textbf{40.51} & \cellcolor[HTML]{EFEFEF}\textbf{40.06} & \cellcolor[HTML]{EFEFEF}\textbf{37.62} \\ 
\bottomrule
\end{tabular}
\end{adjustbox}
\end{table*}

\section{Extended Ablation Study Cases}\label{appendix:ab}

In this section, we present detailed results of the three additional cases in our ablation studies, which were not included in Section~\ref{sec:exp:ab}. Each case explores different combinations of datasets and attack scenarios, providing further insights into the robustness of our method.

\noindent{\textbf{The number of malicious samples.}}
We investigate MedBN's robustness against different ratios of malicious
samples using a batch size of 200. 
In the instant attack scenario, MedBN demonstrates robustness across all malicious ratios, performing well under both targeted attacks (Table~\ref{tab:ab:mal:tar}) and indiscriminate attacks (Table~\ref{tab:ab:mal:indis}).

\begin{table}[htb!]
\centering
\caption{Attack Success Rate (\%) of targeted and instant attack for different numbers of malicious samples $m$ with batch size of 200.} \label{tab:ab:mal:tar}
\begin{adjustbox}{width=0.5\textwidth}
\begin{tabular}{c|c|ccccc}
\toprule
&  & \multicolumn{5}{c}{\# of Malicious Samples ($m$)} \\ \cmidrule{3-7}
\multicolumn{1}{c|}{\multirow{-2}{*}{Dataset}} & \multicolumn{1}{c|}{\multirow{-2}{*}{Normalization}} & 10 & 20 & 40 & 60 & 80 \\ 
\midrule
\multirow{2}{*}{CIFAR10-C} & BatchNorm & 21.60 & 42.00 & 84.00 & 96.67 & 99.47 \\
&\cellcolor[HTML]{EFEFEF}Ours (MedBN) & \cellcolor[HTML]{EFEFEF}\textbf{7.07} & \cellcolor[HTML]{EFEFEF}\textbf{10.27} & \cellcolor[HTML]{EFEFEF}\textbf{19.20} & \cellcolor[HTML]{EFEFEF}\textbf{26.80}  & \cellcolor[HTML]{EFEFEF}\textbf{38.27} \\
\midrule
\multirow{2}{*}{CIFAR100-C} & BatchNorm &  16.80 & 42.13 & 92.00 & 99.73 & 99.87  \\
&\cellcolor[HTML]{EFEFEF}Ours (MedBN) & \cellcolor[HTML]{EFEFEF}\textbf{1.73} & \cellcolor[HTML]{EFEFEF}\textbf{2.00} & \cellcolor[HTML]{EFEFEF}\textbf{2.93} & \cellcolor[HTML]{EFEFEF}\textbf{3.60} & \cellcolor[HTML]{EFEFEF}\textbf{4.27} \\
\bottomrule
\end{tabular}
\end{adjustbox}
\end{table}


\clearpage

\begin{table}[htb!]
\centering
\caption{Error Rate (\%) of indiscriminate and instant attack for different number of malicious samples $m$ with batch size of 200.} \label{tab:ab:mal:indis}
\begin{adjustbox}{width=0.5\textwidth}
\begin{tabular}{c|c|ccccc}
\toprule
&  & \multicolumn{5}{c}{\# of Malicious Samples ($m$)} \\ \cmidrule{3-7}
\multicolumn{1}{c|}{\multirow{-2}{*}{Dataset}} & \multicolumn{1}{c|}{\multirow{-2}{*}{Normalization}} & 10 & 20 & 40 & 60 & 80 \\ 
\midrule
\multirow{2}{*}{CIFAR10-C} & BatchNorm & 19.07 & 22.98	& 31.02	& 40.14	& 50.52 \\
&\cellcolor[HTML]{EFEFEF}Ours (MedBN) & \cellcolor[HTML]{EFEFEF}\textbf{16.42} & \cellcolor[HTML]{EFEFEF}\textbf{18.00} & \cellcolor[HTML]{EFEFEF}\textbf{22.34} & \cellcolor[HTML]{EFEFEF}\textbf{28.00}  & \cellcolor[HTML]{EFEFEF}\textbf{34.24} \\
\midrule
\multirow{2}{*}{CIFAR100-C} & BatchNorm & 45.35	& 50.03	& 59.84	& 69.21	& 78.99 \\
&\cellcolor[HTML]{EFEFEF}Ours (MedBN) & \cellcolor[HTML]{EFEFEF}\textbf{43.31}  & \cellcolor[HTML]{EFEFEF}\textbf{44.38} & \cellcolor[HTML]{EFEFEF}\textbf{48.58} & \cellcolor[HTML]{EFEFEF}\textbf{53.86} & \cellcolor[HTML]{EFEFEF}\textbf{61.44} \\
\bottomrule
\end{tabular}
\end{adjustbox}
\end{table}

\noindent{\textbf{Test batch size.}}
We assess the effect of varying batch sizes with a fixed ratio of malicious smaples around 20\%.
In the case of targeted attacks, MedBN consistently achieves significantly lower ASR compared to BN across all batch sizes (refer to Table~\ref{tab:ab:batch:tar}. Similarty, in the case of indiscriminate attacks, MedBN consistently outperforms BN with lower error rates across all tested batch sizes (refer to Table~\ref{tab:ab:batch:indis}).

\begin{table}[htb!]
\centering
\caption{Attack Success Rate (\%) of targeted and instant attack for different batch size $B$ with a consistent 20\% of malicious samples.} \label{tab:ab:batch:tar}
\begin{adjustbox}{width=0.5\textwidth}
\begin{tabular}{c|c|ccccc}
\toprule
&  & \multicolumn{5}{c}{Batch-size ($B$)} \\ \cmidrule{3-7}
\multicolumn{1}{c|}{\multirow{-2}{*}{Dataset}} & \multicolumn{1}{c|}{\multirow{-2}{*}{Normalization}} & 200 & 128 & 64 & 32 & 16 \\ 
\midrule
\multirow{2}{*}{CIFAR10-C} & BatchNorm & 83.91 & 87.76 & 84.84 & 83.87 & 84.60 \\
&\cellcolor[HTML]{EFEFEF}MedBN (Ours) & \cellcolor[HTML]{EFEFEF}\textbf{19.16} & \cellcolor[HTML]{EFEFEF}\textbf{20.51} & \cellcolor[HTML]{EFEFEF}\textbf{17.83} & \cellcolor[HTML]{EFEFEF}\textbf{20.19} & \cellcolor[HTML]{EFEFEF}\textbf{29.14} \\
\midrule
\multirow{2}{*}{CIFAR100-C} & BatchNorm & 91.78 & 88.44 & 89.43 & 90.46 & 90.47 \\
&\cellcolor[HTML]{EFEFEF}MedBN (Ours) & \cellcolor[HTML]{EFEFEF}\textbf{2.80} & \cellcolor[HTML]{EFEFEF}\textbf{4.72} & \cellcolor[HTML]{EFEFEF}\textbf{5.01} & \cellcolor[HTML]{EFEFEF}\textbf{8.20} & \cellcolor[HTML]{EFEFEF}\textbf{12.65} \\
\bottomrule
\end{tabular}
\end{adjustbox}
\end{table}

\begin{table}[htb!]
\centering
\caption{Error Rate (\%) of indiscriminate and instant attack for different batch size $B$ with a consistent 20\% of malicious samples.} \label{tab:ab:batch:indis}
\begin{adjustbox}{width=0.5\textwidth}
\begin{tabular}{c|c|ccccc}
\toprule
&  & \multicolumn{5}{c}{Batch-size ($B$)} \\ \cmidrule{3-7}
\multicolumn{1}{c|}{\multirow{-2}{*}{Dataset}} & \multicolumn{1}{c|}{\multirow{-2}{*}{Normalization}} & 200 & 128 & 64 & 32 & 16 \\ 
\midrule
\multirow{2}{*}{CIFAR10-C} & BatchNorm & 31.02 & 33.14 & 35.01 & 40.67 & 49.85 \\
&\cellcolor[HTML]{EFEFEF}MedBN (Ours) & \cellcolor[HTML]{EFEFEF}\textbf{22.34} & \cellcolor[HTML]{EFEFEF}\textbf{23.83} & \cellcolor[HTML]{EFEFEF}\textbf{24.78} & \cellcolor[HTML]{EFEFEF}\textbf{28.58} & \cellcolor[HTML]{EFEFEF}\textbf{34.81} \\
\midrule
\multirow{2}{*}{CIFAR100-C} & BatchNorm & 59.80 & 62.35 & 67.07 & 73.73 & 83.08 \\
&\cellcolor[HTML]{EFEFEF}MedBN (Ours) & \cellcolor[HTML]{EFEFEF}\textbf{48.55} & \cellcolor[HTML]{EFEFEF}\textbf{49.86} & \cellcolor[HTML]{EFEFEF}\textbf{52.88} & \cellcolor[HTML]{EFEFEF}\textbf{58.80} & \cellcolor[HTML]{EFEFEF}\textbf{67.63} \\
\bottomrule
\end{tabular}
\end{adjustbox}
\end{table}

\section{Extension of Theorem~\ref{thm:1}}\label{appendix:multi-dim-median} 
In this appendix, we extend Theorem~\ref{thm:1} for multi-dimensional vectors. For median of multi-dimensional vectors, we consider coordinate-wise median (\textnormal{cwmed}) and geometric median (\textnormal{geomed}). 
The coordinate-wise median is the median along each dimension. The geometric median is a vector that minimizes the sum of the distances to vectors in $\batch = \{x_i \in \mathbb{R}^{C}: i \in [n]\}$ with a set of $n$ numbers, which is defined as follows:
\begin{align}
    \textnormal{geomed}(\batch) =  \argmin_{z \in \realnum^d} \sum_{x_i \in \batch } \| z - x_i \|_2 \;.
\end{align}
Note that \textnormal{cwmed} is a solution of $\argmin_{z \in \realnum^d} \sum_{x_i \in \batch} \| z - x_i \|_1$.

\begin{theorem}[Extension of Theorem~\ref{thm:1}]
\label{thm:2}
Consider a set of $n$ numbers $\batch = \{x_i \in \mathbb{R}^{C}: i \in [n]\}$ and $1 \le m \le n$ where the first $m$ numbers are possibly manipulated by adversaries. Let $\batch_{\textnormal{mal}} = \{x_i : i \in [m]\}$, and $\batch_{\textnormal{ben}} = \batch \setminus \batch_{\textnormal{mal}}$.

\noindent(i) The mean can be arbitrarily manipulated by a single malicious sample, i.e.,
for any $1 \le m \le n$, 
\vspace{-2mm}
\begin{align}
\sup_{\batch_\textnormal{mal}} \|
\textnormal{mean}(\batch_\textnormal{mal} \cup \batch_\textnormal{ben})
- \textnormal{mean}(\batch_\textnormal{ben})
\|_2 &= \infty 
\label{eq:multi:meanmean}
\;.
\end{align}

\noindent(ii) The \textnormal{cwmed} or \textnormal{geomed} are robust against malicious samples unless they are not the majority, i.e.,
for any $1 \leq m < n/2$. For the simplicity, we denote the \textnormal{med} instead of \textnormal{cwmed} or \textnormal{geomed},
\vspace{-2mm}
\begin{align}
\sup_{\batch_\textnormal{mal}} \|
\textnormal{med}(\batch_\textnormal{mal}\cup \batch_\textnormal{ben})
-\textnormal{med}(\batch_\textnormal{ben})
\|_2 &<
\infty \;, \;\text{and}
\label{eq:multi:medmed}
\\
\sup_{\batch_\textnormal{mal}} \|
\textnormal{med}(\batch_\textnormal{mal}\cup \batch_\textnormal{ben})
-\textnormal{mean}(\batch_\textnormal{ben})
\|_2 &< \infty \;.
\label{eq:multi:medmean}
\end{align}
\end{theorem}

\noindent\textbf{Proof of Theorem~\ref{thm:2}.}
First, we prove the vulnerability of mean~\eqref{eq:multi:meanmean}.
The $k$-th coodrinate of $\|
\textnormal{mean}(\batch_\textnormal{mal}\cup \batch_\textnormal{ben})
-\textnormal{mean}(\batch_\textnormal{ben}) \| $  is $\textnormal{mean}(\batch_\textnormal{mal} \cup \batch_\textnormal{ben})_k - \textnormal{mean}(\batch_\textnormal{ben})_k$. Then, $\|
\textnormal{mean}(\batch_\textnormal{mal} \cup \batch_\textnormal{ben})
- \textnormal{mean}(\batch_\textnormal{ben})\|
= \sqrt{ \sum_{k=1}^{C} | \textnormal{mean}(\batch_\textnormal{mal} \cup \batch_\textnormal{ben})_k - \textnormal{mean}(\batch_\textnormal{ben})_k|^2 }
$. Consequently, by \eqref{eq:meanmean}, the equation~\eqref{eq:multi:meanmean} holds. 

For the second part on the robustness of the median, particularly for the \textnormal{cwmed}, we can demonstrate (\ref{eq:multi:medmed}) and (\ref{eq:multi:medmean}) by using (\ref{eq:medmean}) and (\ref{eq:medmed}), similar to the proof of (\ref{eq:multi:meanmean}). 
Regarding the \textnormal{geomed}, we can use Lemma~9 in \cite{farhadkhani2022byzantine}, 
\begin{align}\label{lemma9}
\|
\textnormal{med}(\batch_\textnormal{mal}\cup \batch_\textnormal{ben})
-\textnormal{med}(\batch_\textnormal{ben})
\|_2 = \frac{1}{\sqrt{1 - \frac{m^2}{(n-m)^2}}}\max_{x_j \in \batch_\textnormal{ben}}\| x_j - \textnormal{med}(\batch_\textnormal{ben}) \|_2 < \infty \;.
\end{align}
Therefore, by \eqref{lemma9}, the equation~\eqref{eq:multi:medmed} holds. Similarly, we can demonstrate the \eqref{eq:multi:medmean}.

\paragraph{Remarks.}
In contrast to \textnormal{cwmed}, calculating \textnormal{geomed} is computationally expensive as it necessitates an optimization procedure. Therefore, although \textnormal{geomed} can be considered for robust batch normalization, it is challenging to apply it to the neural network, which generally operates with high dimensional features.

\section{Discussion on Median Absolute Deviation (MAD)}\label{appendix:mad}

We further explore the feasibility of using Median Absolute Deviation (MAD) as an alternative to the mean of squared deviations $(z_{bchw} - \eta_{c})^2$, used in our MedBN. The MAD is calculated as the median of the absolute deviations from the median of data, formulated as: 
\begin{align}
    \textrm{med}\left( | z_{bchw} - \eta_{c} | \right)_{bhw} \;.
\end{align}
As discussed in Section~\ref{sec:method:medbn}, $\rho_c$ typically computes the mean of squared deviations $(z_{bchw} - \eta_{c})^2$
, opting for MAD presents an alternative method.
Our findings reveal that while adopting MAD enhances defense capabilities, specifically in the targeted attack, it also results in a notable decrease in performance, particularly over ImageNet-C, as detailed in Table~\ref{tab:mad}.

\begin{table*}[htb!]\centering
\caption{Comparison of BatchNorm, MedBN (our method), and MAD in terms of Attack Success Rate (\%) for the targeted and instant attack scenario and Error Rate (\%) for the indiscriminate and instant attack scenario using TeBN. This table also includes Error Rate (\%) on benign samples without attack as per TTA benchmarks. } \label{tab:mad}
\begin{adjustbox}{width=0.8\textwidth}
\begin{tabular}{c|c|cc|cc|cc}
\toprule
 & Dataset & \multicolumn{2}{c|}{CIFAR10-C} & \multicolumn{2}{c|}{CIFAR100-C} & \multicolumn{2}{c}{ImageNet-C} \\
\midrule
& $m$ / $B$ & \multicolumn{2}{c|}{40 / 200} & \multicolumn{2}{c|}{40 / 200} & \multicolumn{2}{c}{20 / 200} \\
\midrule\midrule
Objective & Normalization & \begin{tabular}[c]{@{}c@{}}ER (\%)\\ w/o Attack \end{tabular} & ASR (\%) & \begin{tabular}[c]{@{}c@{}}ER (\%)\\ w/o Attack \end{tabular} & ASR (\%) & \begin{tabular}[c]{@{}c@{}}ER (\%)\\ w/o Attack \end{tabular} & ASR (\%) \\
\midrule
 & BatchNorm &  14.92 & 83.90 & 40.08 & 91.78 & 66.62 & 97.78 \\
\multicolumn{1}{c|}{\multirow{-1.5}{*}{\begin{tabular}[c]{@{}c@{}}\textit{Targeted}\\ \textit{Attack}\end{tabular}}} & \cellcolor[HTML]{EFEFEF}Ours (MedBN) & \cellcolor[HTML]{EFEFEF}15.19 & \cellcolor[HTML]{EFEFEF}19.16 & \cellcolor[HTML]{EFEFEF}40.77 & \cellcolor[HTML]{EFEFEF}2.80 & \cellcolor[HTML]{EFEFEF}69.55 & \cellcolor[HTML]{EFEFEF}0.36 \\
 & MAD & 18.40  & 2.93 & 46.13 & 0.13 & 85.08 & 0.27 \\
\midrule\midrule
Objective & Normalization & \begin{tabular}[c]{@{}c@{}}ER (\%)\\ w/o Attack \end{tabular} & ER (\%) & \begin{tabular}[c]{@{}c@{}}ER (\%)\\ w/o Attack \end{tabular} & ER (\%) & \begin{tabular}[c]{@{}c@{}}ER (\%)\\ w/o Attack \end{tabular} & ER (\%) \\
\midrule
 & BatchNorm &  14.92 & 31.02 & 40.08 & 59.80 & 66.62 & 81.46 \\
\multicolumn{1}{c|}{\multirow{-1.5}{*}{\begin{tabular}[c]{@{}c@{}}\textit{Indiscriminate}\\ \textit{Attack}\end{tabular}}} & \cellcolor[HTML]{EFEFEF}Ours (MedBN) & \cellcolor[HTML]{EFEFEF}15.19 & \cellcolor[HTML]{EFEFEF}22.34 & \cellcolor[HTML]{EFEFEF}40.77 & \cellcolor[HTML]{EFEFEF}48.55 & \cellcolor[HTML]{EFEFEF}69.55 & \cellcolor[HTML]{EFEFEF}69.74 \\
 & MAD & 18.40 & 23.46 & 46.13 & 53.41 & 85.08 & 84.99 \\
\bottomrule
\end{tabular}
\end{adjustbox}
\end{table*}

\section{Comprehensive Analysis of Malicious Samples on Every BN Layers}\label{appendix:t-sne}
For analyzing the effect of MedBN, we plot the t-SNE of features before going through BN layers.
For evaluation, we use Gaussian corruptions in CIFAR10-C with ResNet26 and TeBN for the adaptation method. The attack is implemented for targeted and instant attack scenario and we use $\varepsilon=1$ for the attack. Figure~\ref{fig:tsne:bn:all} shows that for the deeper layer, the malicious samples tend to be clustered and distant from the benign samples to mislead the output of the model. 

\paragraph{Additional analysis with constrained $\varepsilon$.}

We conduct a comparative analysis of BN and MedBN under the same setup, except for using a constrained $\varepsilon$ value of $8/255$.
Table~\ref{result:small:eps} shows that the reduced $\varepsilon$ leads a lower ASR compared to $\varepsilon=1$,
indicating a weaker attack. 
Moreover, our methods outperforms in both cases, with $\varepsilon=1$ and $\varepsilon=8/255$.
We plot the $L_1$ distance as outlined in Section~\ref{sec:exp:why}. Figure~\ref{fig:bn_med_dist:v2} and Figure~\ref{fig:bn_med_dist:v3} show that MedBN statistics is less influenced by malicious samples than BN statistics.
Comparing the early layers (specifically, in bn1) between attack with $\varepsilon=1$ (the left of the Figure~\ref{fig:bn_med_dist:v2}) and attack with $\varepsilon=8/255$ (the right of the Figure~\ref{fig:bn_med_dist:v2}),
we can observe that a smaller $\varepsilon$ value leads to reduced perturbations in the early layers.
In other words, as the weaker attack, the perturbation for early layers is reduced. 

Additionally, we visualize t-SNE of all layers and $\varepsilon$'s in Figure~\ref{fig:tsne:1} and Figure~\ref{fig:tsne:8-255}.
In contrast to BN layers in $\varepsilon=1$ (Figure~\ref{fig:tsne:bn:all}), BN layers in $\varepsilon=8/255$ (Figure~\ref{fig:tsne:bn:small:all}) shows that malicious samples tend to be clustered and become more distant from benign samples at deeper layers than under $\varepsilon=1$, indicating a weakened attack. 
However, as shown in MedBN layers in $\varepsilon=1$ (Figure~\ref{fig:tsne:med:all}), 
MedBN layers in $\varepsilon=8/255$ (Figure~\ref{fig:tsne:med:small:all}) demonstrates that MedBN effectively mitigates the malicious samples to not be outlier against the benign samples, i.e., malicious samples are closed from the benign samples.


\begin{table}[htb!]\centering 
\caption{Attack Success Rate (\%) of targeted and instant attacks for different $\varepsilon$ by using TeBN.} \label{result:small:eps}
\begin{adjustbox}{width=0.35\textwidth}
\begin{tabular}{c|c|cc}
\toprule
&  & \multicolumn{2}{c}{value of $\varepsilon$} \\ \cmidrule{3-4}
\multicolumn{1}{c|}{\multirow{-2}{*}{Dataset}} & \multicolumn{1}{c|}{\multirow{-2}{*}{Normalization}} & 8/255 & 1 \\ 
\midrule
\multirow{2}{*}{CIFAR10-C} & BatchNorm & 58.00 & 83.91 \\
&\cellcolor[HTML]{EFEFEF}MedBN (Ours) & \cellcolor[HTML]{EFEFEF}\textbf{16.13} & \cellcolor[HTML]{EFEFEF}\textbf{19.16}\\
\bottomrule
\end{tabular}
\end{adjustbox}
\end{table}

\begin{figure}[htb!]
    \centering
    \begin{subfigure}[b]{0.4\textwidth}
         \centering
    \includegraphics[width=\textwidth]{figures/f5-mean_diff.pdf}
     \end{subfigure}
    \hspace{1.0em}
    \begin{subfigure}[b]{0.4\textwidth}
         \centering
         \includegraphics[width=\textwidth]{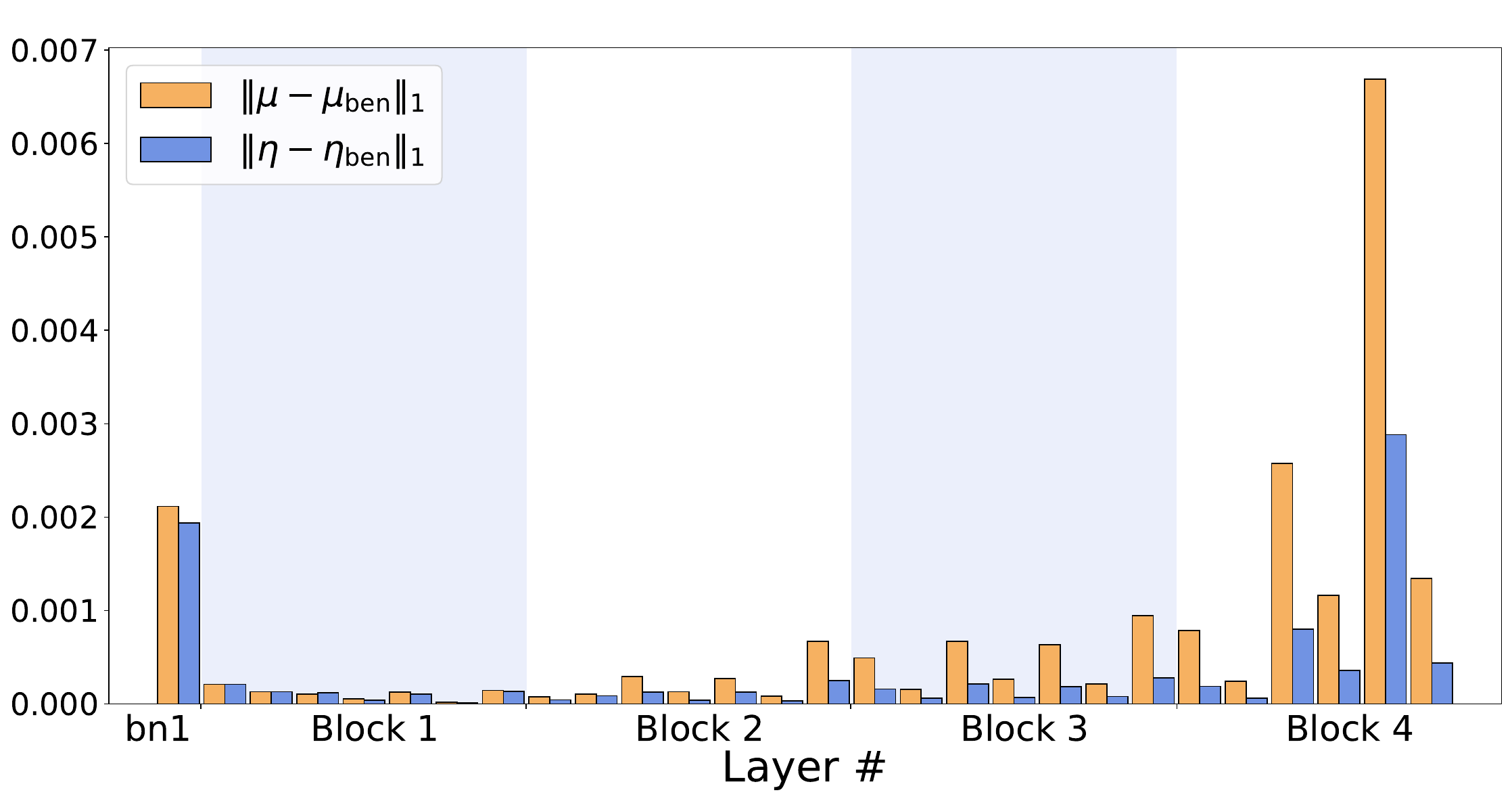}
     \end{subfigure}
    \caption{L1 distance for measuring the amount of perturbation $\| \mu - \mu_{\text{ben}}\|_1$ and $\| \eta -\eta_{\text{ben}} \|_1$ by malicious samples across various layers, with $\varepsilon=1$ on the left and $\varepsilon=8/255$ on the right.
    }
    \label{fig:bn_med_dist:v2}
    \vspace{-1em}
\end{figure}

\begin{figure}[htb!]
    \centering
    \begin{subfigure}[b]{0.4\textwidth}
         \centering
         \includegraphics[width=\textwidth]{figures/f5-sigma_diff.pdf}
     \end{subfigure}
    \hspace{1.0em}
    \begin{subfigure}[b]{0.4\textwidth}
         \centering
         \includegraphics[width=\textwidth]{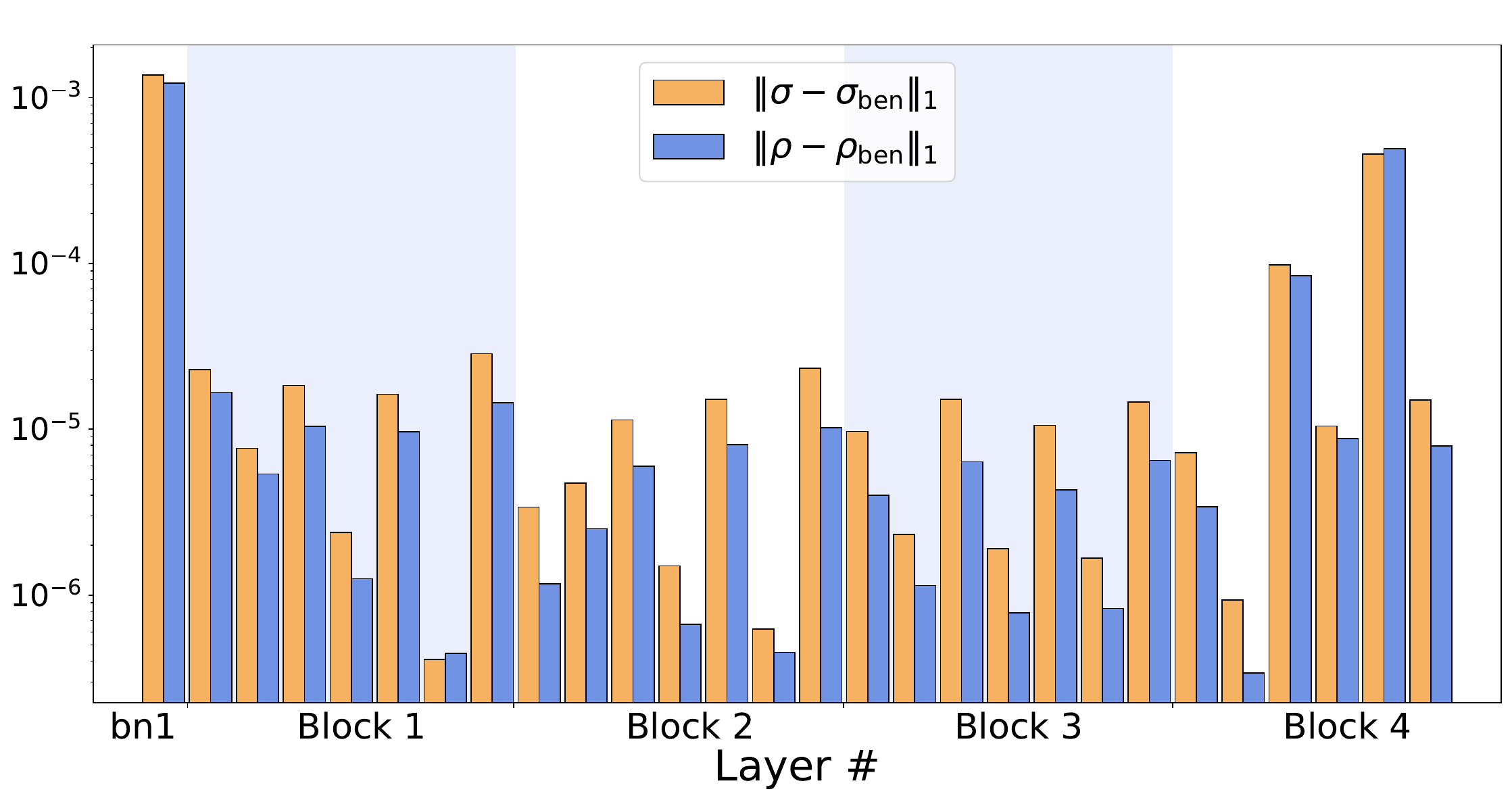}
     \end{subfigure}
    \caption{L1 distance for measuring the amount of perturbation $\| \sigma - \sigma_{\text{ben}}\|_1$ and $\| \rho -\rho_{\text{ben}} \|_1$ by malicious samples across various layers, with $\varepsilon=1$ on the left and $\varepsilon=8/255$ on the right.
    }
    \label{fig:bn_med_dist:v3}
\end{figure}

\begin{figure*}[htb]
    \centering
    \begin{subfigure}[b]{1.0\textwidth}
        \centering
        \includegraphics[width=\textwidth]{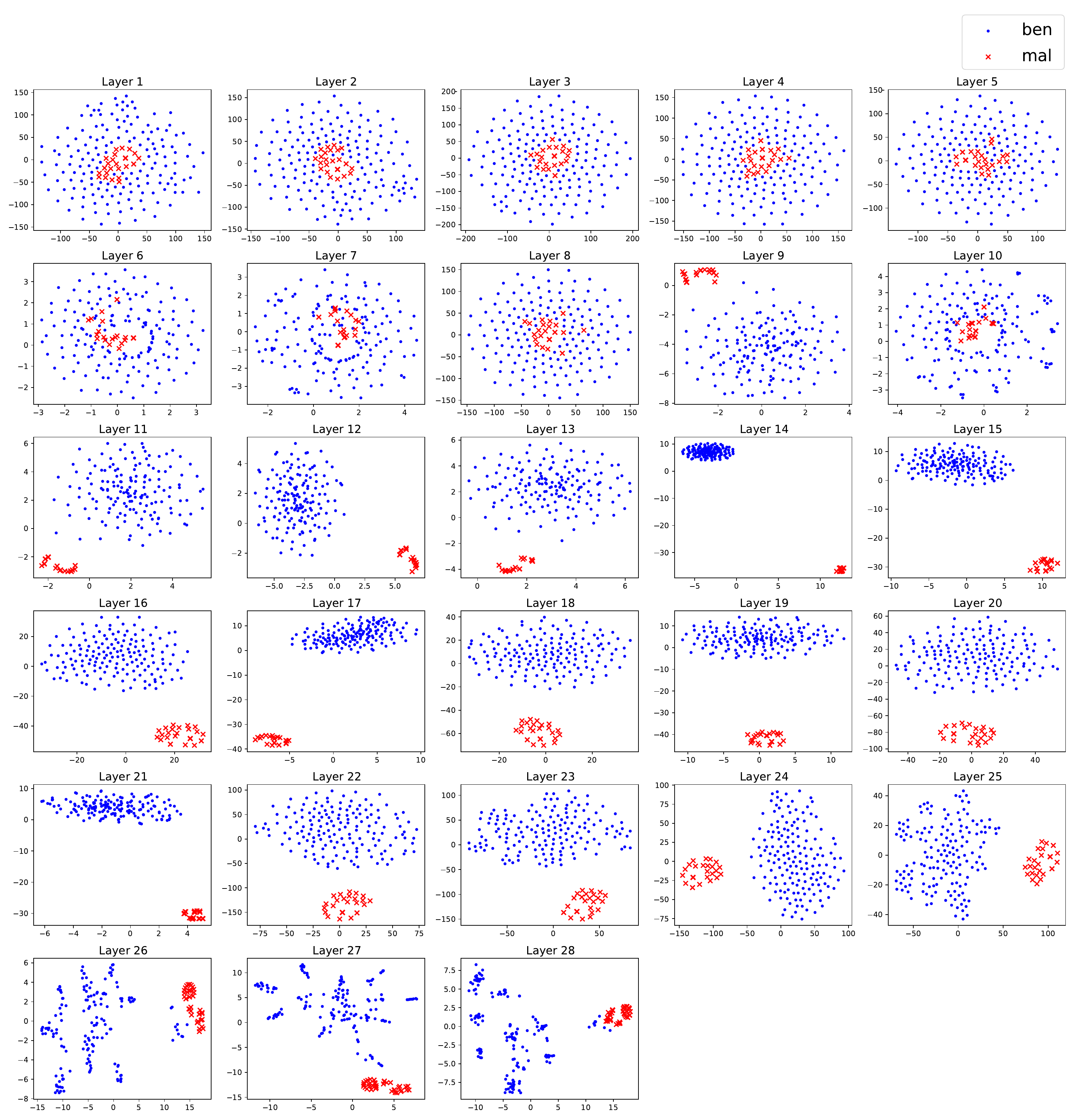} 
        \caption{t-SNE visualization of all BN layers ($\varepsilon=1$).} \label{fig:tsne:bn:all}
    \end{subfigure}
\end{figure*}
\begin{figure*}[htb]\ContinuedFloat
    \begin{subfigure}[b]{1.0\textwidth}
       \centering
       \includegraphics[width=\textwidth]{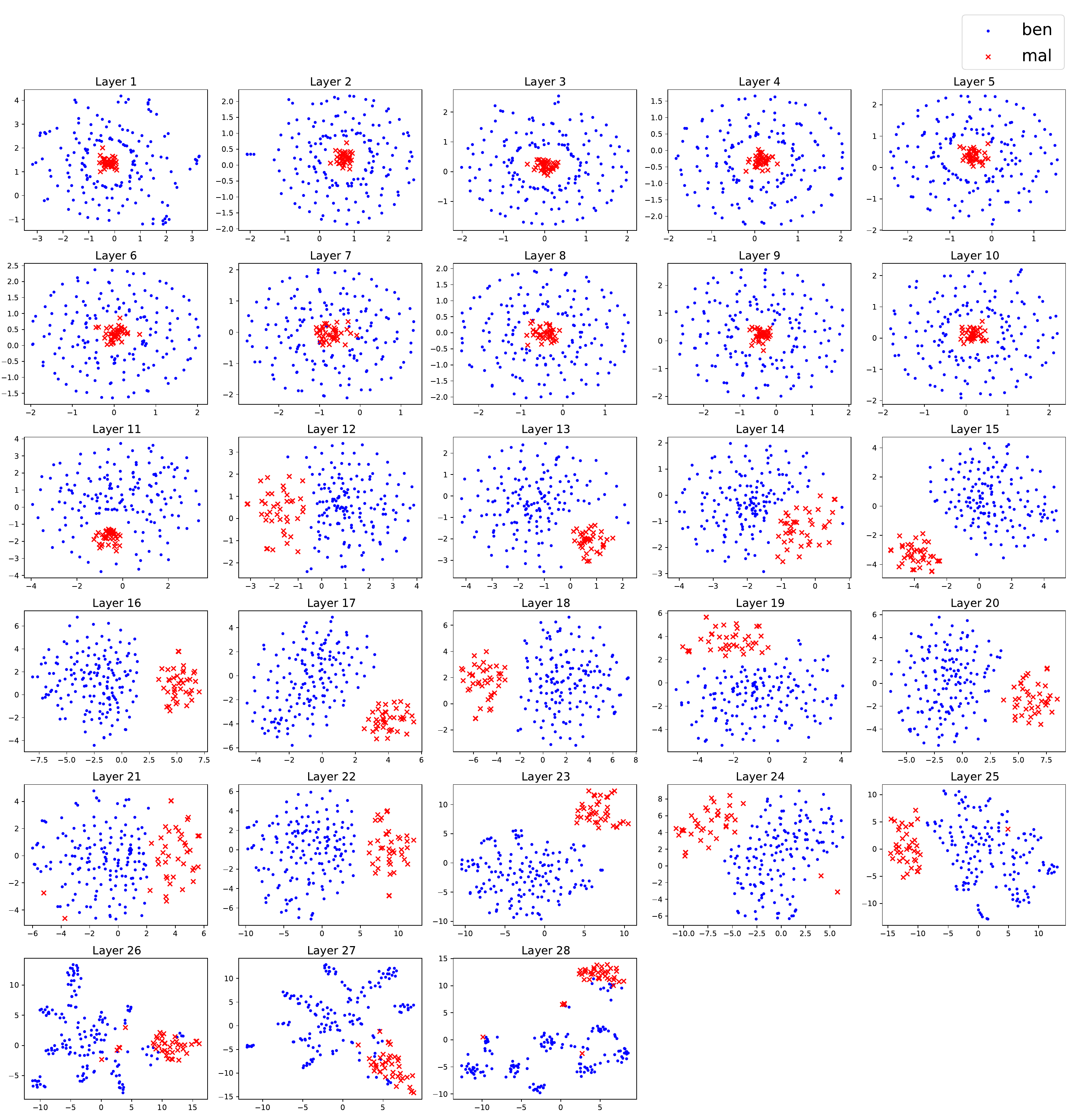} 
       \caption{t-SNE visualization of all MedBN layers ($\varepsilon=1$).} \label{fig:tsne:med:all}
    \end{subfigure}
    \caption{t-SNE visualization of all BN layers (Figure~\ref{fig:tsne:bn:all}) and all MedBN layers (Figure~\ref{fig:tsne:med:all}) with $\varepsilon=1$. 
    In Figure~\ref{fig:tsne:bn:all}, for deeper layers, the features of malicious samples tend to be distant from benign samples to mislead the outputs of model. However, when we apply MedBN,
    Figure~\ref{fig:tsne:med:all} demonstrates that malicious samples are closed to benign samples, i.e. the effect of malicious samples is significantly mitigated.
    }\label{fig:tsne:1}
\end{figure*}



\begin{figure*}[htb]
    \centering
    \begin{subfigure}[b]{1.0\textwidth}
        \centering
        \includegraphics[width=\textwidth]{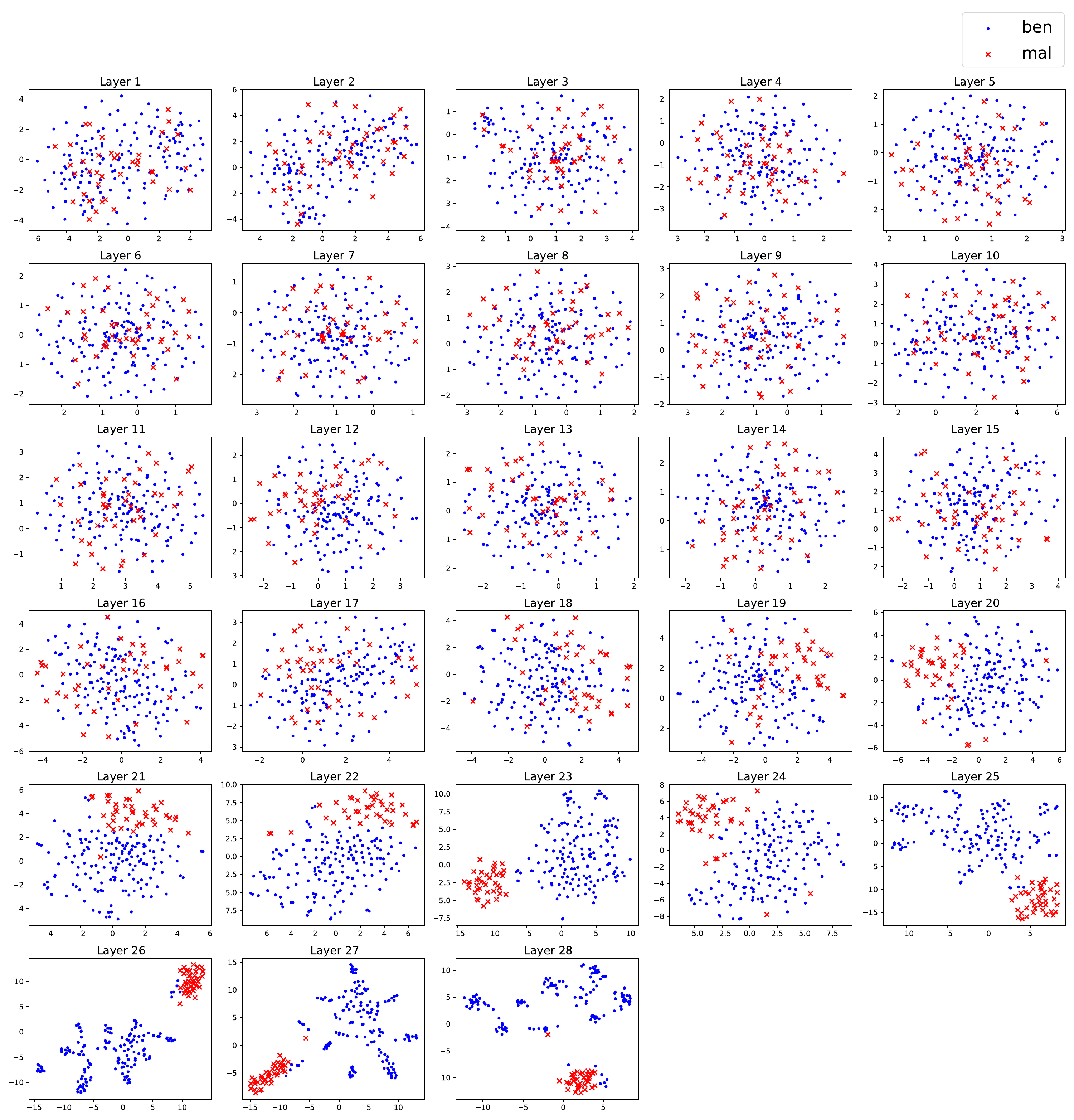} 
        \caption{t-SNE visualization of all BN layers ($\varepsilon=8/255$).} \label{fig:tsne:bn:small:all}
    \end{subfigure}
\end{figure*}
\begin{figure*}[htb]\ContinuedFloat
    \begin{subfigure}[b]{1.0\textwidth}
       \centering
       \includegraphics[width=\textwidth]{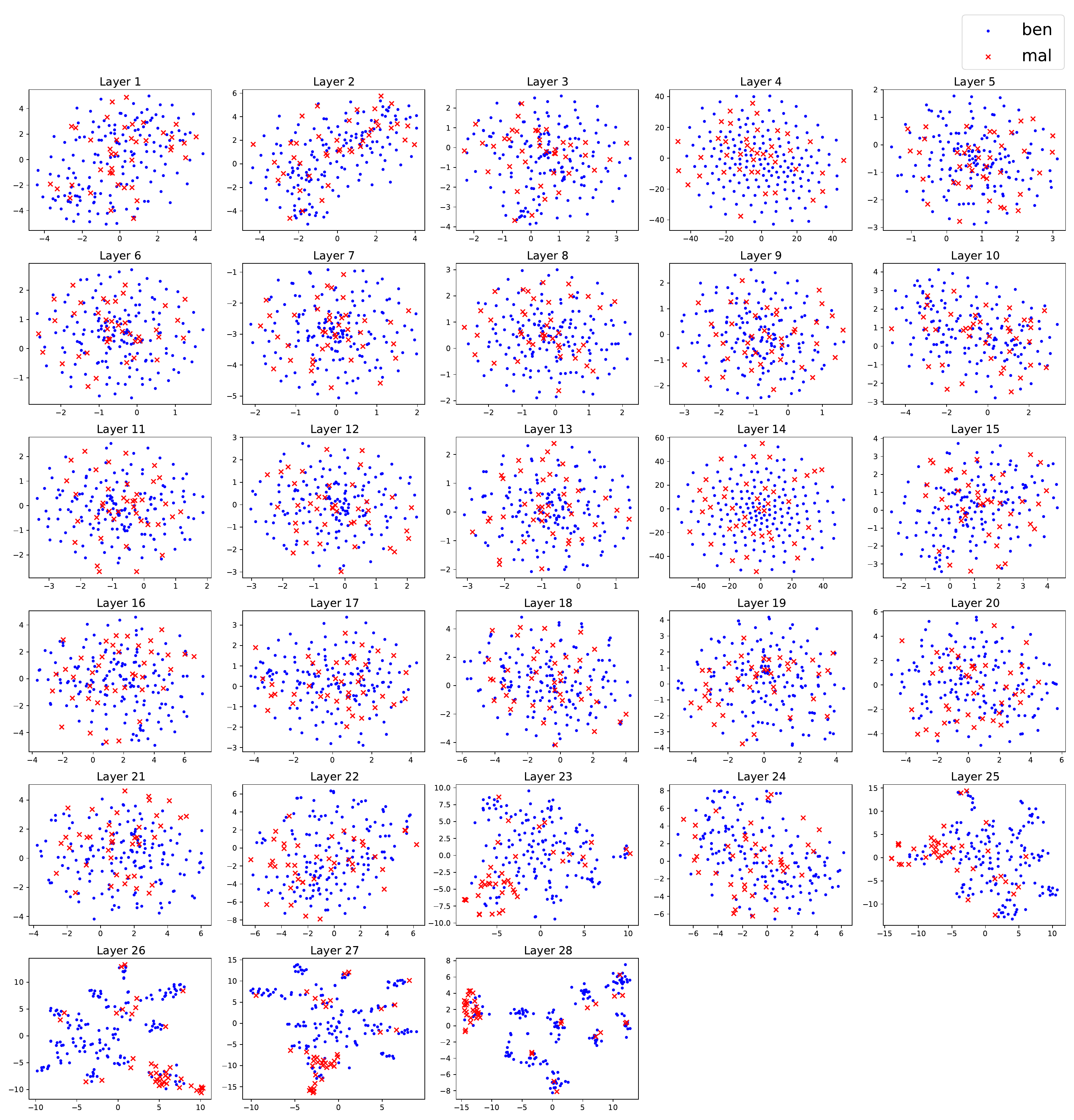} 
       \caption{t-SNE visualization of all MedBN layers ($\varepsilon=8/255$).} \label{fig:tsne:med:small:all}
    \end{subfigure}
    \caption{t-SNE visualization of all BN layers (Figure~\ref{fig:tsne:bn:small:all}) and all MedBN layers (Figure~\ref{fig:tsne:med:small:all}) with $\varepsilon=8/255$. 
    For the early layers, the features of malicious samples tend to be more close to those of benign samples as the $\varepsilon$ is reduced.
    For deeper layers, similar to Figure~\ref{fig:tsne:bn:all}, the malicious samples tend to move away from the benign samples to mislead the model. However, when we apply the MedBN, the impact of malicious samples is significantly mitigated as shown in Figure~\ref{fig:tsne:med:small:all}.
    }\label{fig:tsne:8-255}
\end{figure*}



\clearpage

\section{Comprehensive Results of Instant Attack Scenario}\label{appendix:ins}
We provide detailed results of instant and targeted attack scenario in Table~\ref{tab:ins:target} and instant and indiscriminate attack scenario in Table~\ref{tab:ins:indis} across all types of corruptions in the TTA benchmark datasets.

\begin{table*}[!ht]\centering
\caption{Extended analysis of Attack Success Rate (\%) for targeted and instant attack scenario over all types of corruption (detailed version of Table~\ref{tab:ins:target:avg}).} \label{tab:ins:target}
\begin{adjustbox}{width=1.0\textwidth}
\begin{tabular}
{l|l|cccccccccccccccc}
\toprule
 & & \multicolumn{3}{c}{Noise} & \multicolumn{4}{c}{Blur} & \multicolumn{4}{c}{Weather} & \multicolumn{4}{c}{Digital} &     
\\
\cmidrule(lr){3-5} \cmidrule(lr){6-9} \cmidrule(lr){10-13} \cmidrule(lr){14-17} & \multicolumn{1}{c|}{Method} & Gauss. & Shot & Impul. & Defoc. & Glass & Motion & Zoom & Snow & Frost & Fog & Brit. & Contr. & Elastic & Pixel. & JPEG & Avg. \\ 
\midrule

\multirow{14}{*}{\rotatebox[origin=c]{90}{CIFAR10-C}} & \multicolumn{1}{c|}{TeBN} & 82.00 & 90.00 & 91.33 & 76.67 & 94.00 & 82.67 & 80.00 & 82.00 & 84.00 & 89.33 & 73.33 & 92.00 & 86.00 & 77.33 & 78.00 & 83.91 \\
 & \cellcolor[HTML]{EFEFEF}+MedBN & \cellcolor[HTML]{EFEFEF}\textbf{26.00} & \cellcolor[HTML]{EFEFEF}\textbf{23.33} & \cellcolor[HTML]{EFEFEF}\textbf{22.00} & \cellcolor[HTML]{EFEFEF}\textbf{16.00} & \cellcolor[HTML]{EFEFEF}\textbf{28.67} & \cellcolor[HTML]{EFEFEF}\textbf{16.00} & \cellcolor[HTML]{EFEFEF}\textbf{14.67} & \cellcolor[HTML]{EFEFEF}\textbf{21.33} & \cellcolor[HTML]{EFEFEF}\textbf{16.00} & \cellcolor[HTML]{EFEFEF}\textbf{16.67} & \cellcolor[HTML]{EFEFEF}\textbf{12.00} & \cellcolor[HTML]{EFEFEF}\textbf{16.67} & \cellcolor[HTML]{EFEFEF}\textbf{21.33} & \cellcolor[HTML]{EFEFEF}\textbf{10.67} & \cellcolor[HTML]{EFEFEF}\textbf{26.00} & \cellcolor[HTML]{EFEFEF}\textbf{19.16} \\
 & \multicolumn{1}{c|}{TENT} & 73.33 & 78.00 & 77.33 & 57.33 & 76.00 & 67.33 & 68.67 & 72.67 & 71.33 & 72.67 & 67.33 & 82.67 & 74.67 & 70.00 & 76.00 & 72.36 \\
 & \cellcolor[HTML]{EFEFEF}+MedBN & \cellcolor[HTML]{EFEFEF}\textbf{23.33} & \cellcolor[HTML]{EFEFEF}\textbf{18.67} & \cellcolor[HTML]{EFEFEF}\textbf{19.33} & \cellcolor[HTML]{EFEFEF}\textbf{16.00} & \cellcolor[HTML]{EFEFEF}\textbf{30.67} & \cellcolor[HTML]{EFEFEF}\textbf{16.00} & \cellcolor[HTML]{EFEFEF}\textbf{14.00} & \cellcolor[HTML]{EFEFEF}\textbf{16.67} & \cellcolor[HTML]{EFEFEF}\textbf{12.67} & \cellcolor[HTML]{EFEFEF}\textbf{14.67} & \cellcolor[HTML]{EFEFEF}\textbf{14.00} & \cellcolor[HTML]{EFEFEF}\textbf{19.33} & \cellcolor[HTML]{EFEFEF}\textbf{21.33} & \cellcolor[HTML]{EFEFEF}\textbf{16.67} & \cellcolor[HTML]{EFEFEF}\textbf{22.00} & \cellcolor[HTML]{EFEFEF}\textbf{18.36} \\
 & \multicolumn{1}{c|}{ETA} & 70.67 & 80.67 & 82.67 & 71.33 & 80.00 & 74.67 & 75.33 & 71.33 & 76.00 & 81.33 & 56.67 & 88.00 & 76.67 & 66.67 & 74.00 & 75.007 \\
 & \cellcolor[HTML]{EFEFEF}+MedBN & \cellcolor[HTML]{EFEFEF}\textbf{24.67} & \cellcolor[HTML]{EFEFEF}\textbf{22.00} & \cellcolor[HTML]{EFEFEF}\textbf{21.33} & \cellcolor[HTML]{EFEFEF}\textbf{14.67} & \cellcolor[HTML]{EFEFEF}\textbf{26.00} & \cellcolor[HTML]{EFEFEF}\textbf{14.67} & \cellcolor[HTML]{EFEFEF}\textbf{12.00} & \cellcolor[HTML]{EFEFEF}\textbf{20.00} & \cellcolor[HTML]{EFEFEF}\textbf{12.00} & \cellcolor[HTML]{EFEFEF}\textbf{18.00} & \cellcolor[HTML]{EFEFEF}\textbf{10.67} & \cellcolor[HTML]{EFEFEF}\textbf{16.00} & \cellcolor[HTML]{EFEFEF}\textbf{20.00} & \cellcolor[HTML]{EFEFEF}\textbf{14.67} & \cellcolor[HTML]{EFEFEF}\textbf{23.33} & \cellcolor[HTML]{EFEFEF}\textbf{18.00} \\
 & \multicolumn{1}{c|}{SAR} & 74.00 & 78.67 & 82.67 & 69.33 & 86.00 & 80.67 & 77.33 & 75.33 & 76.00 & 77.33 & 66.67 & 86.67 & 83.33 & 72.67 & 74.67 & 77.42 \\
 & \cellcolor[HTML]{EFEFEF}+MedBN & \cellcolor[HTML]{EFEFEF}\textbf{24.00} & \cellcolor[HTML]{EFEFEF}\textbf{24.00} & \cellcolor[HTML]{EFEFEF}\textbf{17.33} & \cellcolor[HTML]{EFEFEF}\textbf{15.33} & \cellcolor[HTML]{EFEFEF}\textbf{28.00} & \cellcolor[HTML]{EFEFEF}\textbf{14.00} & \cellcolor[HTML]{EFEFEF}\textbf{11.33} & \cellcolor[HTML]{EFEFEF}\textbf{17.33} & \cellcolor[HTML]{EFEFEF}\textbf{14.00} & \cellcolor[HTML]{EFEFEF}\textbf{14.67} & \cellcolor[HTML]{EFEFEF}\textbf{14.67} & \cellcolor[HTML]{EFEFEF}\textbf{16.67} & \cellcolor[HTML]{EFEFEF}\textbf{19.33} & \cellcolor[HTML]{EFEFEF}\textbf{16.00} & \cellcolor[HTML]{EFEFEF}\textbf{24.00} & \cellcolor[HTML]{EFEFEF}\textbf{18.04} \\
 & \multicolumn{1}{c|}{SoTTA} & 25.33 & 20.67 & 30.00 & 20.00 & 24.00 & 22.00 & 15.33 & 21.33 & 17.33 & 18.67 & 22.67 & 18.00 & 22.67 & 21.33 & 22.67 & 21.47 \\
 & \cellcolor[HTML]{EFEFEF}+MedBN & \cellcolor[HTML]{EFEFEF}\textbf{7.33} & \cellcolor[HTML]{EFEFEF}\textbf{16.67} & \cellcolor[HTML]{EFEFEF}\textbf{10.67} & \cellcolor[HTML]{EFEFEF}\textbf{6.67} & \cellcolor[HTML]{EFEFEF}\textbf{6.67} & \cellcolor[HTML]{EFEFEF}\textbf{3.33} & \cellcolor[HTML]{EFEFEF}\textbf{6.00} & \cellcolor[HTML]{EFEFEF}\textbf{8.00} & \cellcolor[HTML]{EFEFEF}\textbf{5.33} & \cellcolor[HTML]{EFEFEF}\textbf{6.00} & \cellcolor[HTML]{EFEFEF}\textbf{4.00} & \cellcolor[HTML]{EFEFEF}\textbf{8.00} & \cellcolor[HTML]{EFEFEF}\textbf{12.00} & \cellcolor[HTML]{EFEFEF}\textbf{6.67} & \cellcolor[HTML]{EFEFEF}\textbf{10.00} & \cellcolor[HTML]{EFEFEF}\textbf{7.82} \\
 & \multicolumn{1}{c|}{sEMA} & 24.67 & 25.33 & 23.33 & 14.00 & 24.67 & 13.33 & 14.00 & 20.00 & 12.00 & 12.67 & 14.67 & 12.67 & 19.33 & 16.00 & 26.00 & 18.18 \\
 & \cellcolor[HTML]{EFEFEF}+MedBN & \cellcolor[HTML]{EFEFEF}\textbf{14.00} & \cellcolor[HTML]{EFEFEF}\textbf{14.00} & \cellcolor[HTML]{EFEFEF}\textbf{14.67} & \cellcolor[HTML]{EFEFEF}\textbf{2.00} & \cellcolor[HTML]{EFEFEF}\textbf{12.00} & \cellcolor[HTML]{EFEFEF}\textbf{6.00} & \cellcolor[HTML]{EFEFEF}\textbf{4.00} & \cellcolor[HTML]{EFEFEF}\textbf{8.00} & \cellcolor[HTML]{EFEFEF}\textbf{6.00} & \cellcolor[HTML]{EFEFEF}\textbf{10.00} & \cellcolor[HTML]{EFEFEF}\textbf{4.00} & \cellcolor[HTML]{EFEFEF}\textbf{6.00} & \cellcolor[HTML]{EFEFEF}\textbf{9.33} & \cellcolor[HTML]{EFEFEF}\textbf{2.00} & \cellcolor[HTML]{EFEFEF}\textbf{18.00} & \cellcolor[HTML]{EFEFEF}\textbf{8.67} \\
 & \multicolumn{1}{c|}{mDIA} & 44.00 & 34.00 & 52.67 & 24.00 & 52.00 & 28.67 & 26.00 & 25.33 & 20.00 & 34.00 & 22.00 & 36.00 & 42.00 & 34.00 & 34.00 & 33.91 \\
 & \cellcolor[HTML]{EFEFEF}+MedBN & \cellcolor[HTML]{EFEFEF}\textbf{12.00} & \cellcolor[HTML]{EFEFEF}\textbf{14.00} & \cellcolor[HTML]{EFEFEF}\textbf{16.00} & \cellcolor[HTML]{EFEFEF}\textbf{2.00} & \cellcolor[HTML]{EFEFEF}\textbf{10.00} & \cellcolor[HTML]{EFEFEF}\textbf{6.00} & \cellcolor[HTML]{EFEFEF}\textbf{4.00} & \cellcolor[HTML]{EFEFEF}\textbf{8.00} & \cellcolor[HTML]{EFEFEF}\textbf{4.00} & \cellcolor[HTML]{EFEFEF}\textbf{10.00} & \cellcolor[HTML]{EFEFEF}\textbf{4.00} & \cellcolor[HTML]{EFEFEF}\textbf{8.00} & \cellcolor[HTML]{EFEFEF}\textbf{12.00} & \cellcolor[HTML]{EFEFEF}\textbf{3.33} & \cellcolor[HTML]{EFEFEF}\textbf{18.00} & \cellcolor[HTML]{EFEFEF}\textbf{8.76} \\

\midrule
\multirow{14}{*}{\rotatebox[origin=c]{90}{CIFAR100-C}} & \multicolumn{1}{c|}{TeBN} & 96.00 & 96.00 & 98.00 & 77.33 & 91.33 & 88.67 & 86.67 & 98.00 & 98.67 & 99.33 & 94.00 & 98.00 & 88.00 & 83.33 & 83.33 & 91.78 \\
 & \cellcolor[HTML]{EFEFEF}+MedBN & \cellcolor[HTML]{EFEFEF}\textbf{2.67} & \cellcolor[HTML]{EFEFEF}\textbf{2.00} & \cellcolor[HTML]{EFEFEF}\textbf{2.00} & \cellcolor[HTML]{EFEFEF}\textbf{2.00} & \cellcolor[HTML]{EFEFEF}\textbf{5.33} & \cellcolor[HTML]{EFEFEF}\textbf{2.00} & \cellcolor[HTML]{EFEFEF}\textbf{2.00} & \cellcolor[HTML]{EFEFEF}\textbf{2.67} & \cellcolor[HTML]{EFEFEF}\textbf{2.00} & \cellcolor[HTML]{EFEFEF}\textbf{4.00} & \cellcolor[HTML]{EFEFEF}\textbf{4.00} & \cellcolor[HTML]{EFEFEF}\textbf{2.67} & \cellcolor[HTML]{EFEFEF}\textbf{4.00} & \cellcolor[HTML]{EFEFEF}\textbf{2.67} & \cellcolor[HTML]{EFEFEF}\textbf{2.00} & \cellcolor[HTML]{EFEFEF}\textbf{2.80} \\
 & \multicolumn{1}{c|}{TENT} & 78.67 & 84.67 & 73.33 & 81.33 & 73.33 & 81.33 & 65.33 & 74.67 & 78.67 & 88.67 & 80.67 & 92.00 & 84.00 & 84.00 & 68.67 & 79.29 \\
 & \cellcolor[HTML]{EFEFEF}+MedBN & \cellcolor[HTML]{EFEFEF}\textbf{3.33} & \cellcolor[HTML]{EFEFEF}\textbf{4.00} & \cellcolor[HTML]{EFEFEF}\textbf{4.00} & \cellcolor[HTML]{EFEFEF}\textbf{4.00} & \cellcolor[HTML]{EFEFEF}\textbf{6.67} & \cellcolor[HTML]{EFEFEF}\textbf{3.33} & \cellcolor[HTML]{EFEFEF}\textbf{4.67} & \cellcolor[HTML]{EFEFEF}\textbf{4.67} & \cellcolor[HTML]{EFEFEF}\textbf{2.00} & \cellcolor[HTML]{EFEFEF}\textbf{6.67} & \cellcolor[HTML]{EFEFEF}\textbf{4.67} & \cellcolor[HTML]{EFEFEF}\textbf{4.67} & \cellcolor[HTML]{EFEFEF}\textbf{4.67} & \cellcolor[HTML]{EFEFEF}\textbf{3.33} & \cellcolor[HTML]{EFEFEF}\textbf{2.00} & \cellcolor[HTML]{EFEFEF}\textbf{4.18} \\
 & \multicolumn{1}{c|}{ETA} & 78.00 & 80.67 & 81.33 & 84.67 & 74.00 & 78.00 & 71.33 & 77.33 & 84.00 & 92.00 & 72.67 & 90.67 & 84.67 & 76.67 & 73.33 & 79.96 \\
 & \cellcolor[HTML]{EFEFEF}+MedBN & \cellcolor[HTML]{EFEFEF}\textbf{2.00} & \cellcolor[HTML]{EFEFEF}\textbf{3.33} & \cellcolor[HTML]{EFEFEF}\textbf{2.00} & \cellcolor[HTML]{EFEFEF}\textbf{4.00} & \cellcolor[HTML]{EFEFEF}\textbf{5.33} & \cellcolor[HTML]{EFEFEF}\textbf{3.33} & \cellcolor[HTML]{EFEFEF}\textbf{3.33} & \cellcolor[HTML]{EFEFEF}\textbf{2.00} & \cellcolor[HTML]{EFEFEF}\textbf{2.00} & \cellcolor[HTML]{EFEFEF}\textbf{6.00} & \cellcolor[HTML]{EFEFEF}\textbf{3.33} & \cellcolor[HTML]{EFEFEF}\textbf{2.00} & \cellcolor[HTML]{EFEFEF}\textbf{2.67} & \cellcolor[HTML]{EFEFEF}\textbf{2.00} & \cellcolor[HTML]{EFEFEF}\textbf{2.00} & \cellcolor[HTML]{EFEFEF}\textbf{3.02} \\
 & \multicolumn{1}{c|}{SAR} & 86.00 & 83.33 & 86.00 & 74.67 & 78.67 & 76.00 & 70.67 & 81.33 & 85.33 & 93.33 & 76.00 & 96.67 & 79.33 & 86.00 & 71.33 & 81.64 \\
 & \cellcolor[HTML]{EFEFEF}+MedBN & \cellcolor[HTML]{EFEFEF}\textbf{2.00} & \cellcolor[HTML]{EFEFEF}\textbf{2.00} & \cellcolor[HTML]{EFEFEF}\textbf{2.00} & \cellcolor[HTML]{EFEFEF}\textbf{4.00} & \cellcolor[HTML]{EFEFEF}\textbf{7.33} & \cellcolor[HTML]{EFEFEF}\textbf{1.33} & \cellcolor[HTML]{EFEFEF}\textbf{4.00} & \cellcolor[HTML]{EFEFEF}\textbf{2.00} & \cellcolor[HTML]{EFEFEF}\textbf{1.33} & \cellcolor[HTML]{EFEFEF}\textbf{6.00} & \cellcolor[HTML]{EFEFEF}\textbf{4.00} & \cellcolor[HTML]{EFEFEF}\textbf{2.00} & \cellcolor[HTML]{EFEFEF}\textbf{3.33} & \cellcolor[HTML]{EFEFEF}\textbf{2.67} & \cellcolor[HTML]{EFEFEF}\textbf{1.33} & \cellcolor[HTML]{EFEFEF}\textbf{3.02} \\
 & \multicolumn{1}{c|}{SoTTA} & 6.67 & 10.00 & 7.33 & 7.33 & 12.67 & 8.00 & 5.33 & 8.67 & 6.67 & 8.67 & 4.00 & 7.33 & 10.00 & 3.33 & 8.00 & 7.60 \\
 & \cellcolor[HTML]{EFEFEF}+MedBN & \cellcolor[HTML]{EFEFEF}\textbf{2.00} & \cellcolor[HTML]{EFEFEF}\textbf{2.00} & \cellcolor[HTML]{EFEFEF}\textbf{3.33} & \cellcolor[HTML]{EFEFEF}\textbf{2.00} & \cellcolor[HTML]{EFEFEF}\textbf{4.67} & \cellcolor[HTML]{EFEFEF}\textbf{2.00} & \cellcolor[HTML]{EFEFEF}\textbf{2.00} & \cellcolor[HTML]{EFEFEF}\textbf{2.00} & \cellcolor[HTML]{EFEFEF}\textbf{1.33} & \cellcolor[HTML]{EFEFEF}\textbf{1.33} & \cellcolor[HTML]{EFEFEF}\textbf{4.00} & \cellcolor[HTML]{EFEFEF}\textbf{3.33} & \cellcolor[HTML]{EFEFEF}\textbf{4.00} & \cellcolor[HTML]{EFEFEF}\textbf{2.67} & \cellcolor[HTML]{EFEFEF}\textbf{2.00} & \cellcolor[HTML]{EFEFEF}\textbf{2.58} \\
 & \multicolumn{1}{c|}{sEMA} & 10.00 & 14.67 & 10.00 & 8.00 & 9.33 & 8.67 & 4.00 & 5.33 & 11.33 & 10.00 & 6.67 & 8.00 & 9.33 & 6.00 & 9.33 & 8.71 \\
 & \cellcolor[HTML]{EFEFEF}+MedBN & \cellcolor[HTML]{EFEFEF}\textbf{2.00} & \cellcolor[HTML]{EFEFEF}\textbf{2.00} & \cellcolor[HTML]{EFEFEF}\textbf{2.00} & \cellcolor[HTML]{EFEFEF}\textbf{2.00} & \cellcolor[HTML]{EFEFEF}\textbf{4.00} & \cellcolor[HTML]{EFEFEF}\textbf{2.00} & \cellcolor[HTML]{EFEFEF}\textbf{0.00} & \cellcolor[HTML]{EFEFEF}\textbf{0.00} & \cellcolor[HTML]{EFEFEF}\textbf{0.00} & \cellcolor[HTML]{EFEFEF}\textbf{0.00} & \cellcolor[HTML]{EFEFEF}\textbf{4.00} & \cellcolor[HTML]{EFEFEF}\textbf{2.00} & \cellcolor[HTML]{EFEFEF}\textbf{2.00} & \cellcolor[HTML]{EFEFEF}\textbf{2.00} & \cellcolor[HTML]{EFEFEF}\textbf{0.00} & \cellcolor[HTML]{EFEFEF}\textbf{1.60} \\
 & \multicolumn{1}{c|}{mDIA} & 15.33 & 18.00 & 22.00 & 14.00 & 14.00 & 16.00 & 12.00 & 16.00 & 18.00 & 24.00 & 10.00 & 24.00 & 16.00 & 12.00 & 18.00 & 16.62 \\
 & \cellcolor[HTML]{EFEFEF}+MedBN & \cellcolor[HTML]{EFEFEF}\textbf{2.00} & \cellcolor[HTML]{EFEFEF}\textbf{4.00} & \cellcolor[HTML]{EFEFEF}\textbf{6.00} & \cellcolor[HTML]{EFEFEF}\textbf{2.00} & \cellcolor[HTML]{EFEFEF}\textbf{4.00} & \cellcolor[HTML]{EFEFEF}\textbf{0.00} & \cellcolor[HTML]{EFEFEF}\textbf{2.00} & \cellcolor[HTML]{EFEFEF}\textbf{2.00} & \cellcolor[HTML]{EFEFEF}\textbf{0.00} & \cellcolor[HTML]{EFEFEF}\textbf{0.00} & \cellcolor[HTML]{EFEFEF}\textbf{4.00} & \cellcolor[HTML]{EFEFEF}\textbf{2.00} & \cellcolor[HTML]{EFEFEF}\textbf{2.00} & \cellcolor[HTML]{EFEFEF}\textbf{0.00} & \cellcolor[HTML]{EFEFEF}\textbf{0.00} & \cellcolor[HTML]{EFEFEF}\textbf{2.00} \\

\midrule
\multirow{14}{*}{\rotatebox[origin=c]{90}{ImageNet-C}} & \multicolumn{1}{c|}{TeBN} & 100.00 & 100.00 & 100.00 & 100.00 & 100.00 & 100.00 & 96.00 & 97.33 & 94.67 & 98.67 & 98.67 & 100.00 & 100.00 & 89.33 & 92.00 & 97.78 \\
 & \cellcolor[HTML]{EFEFEF}+MedBN & \cellcolor[HTML]{EFEFEF}\textbf{0.00} & \cellcolor[HTML]{EFEFEF}\textbf{0.00} & \cellcolor[HTML]{EFEFEF}\textbf{0.00} & \cellcolor[HTML]{EFEFEF}\textbf{0.00} & \cellcolor[HTML]{EFEFEF}\textbf{0.00} & \cellcolor[HTML]{EFEFEF}\textbf{0.00} & \cellcolor[HTML]{EFEFEF}\textbf{0.00} & \cellcolor[HTML]{EFEFEF}\textbf{0.00} & \cellcolor[HTML]{EFEFEF}\textbf{1.33} & \cellcolor[HTML]{EFEFEF}\textbf{0.00} & \cellcolor[HTML]{EFEFEF}\textbf{0.00} & \cellcolor[HTML]{EFEFEF}\textbf{4.00} & \cellcolor[HTML]{EFEFEF}\textbf{0.00} & \cellcolor[HTML]{EFEFEF}\textbf{0.00} & \cellcolor[HTML]{EFEFEF}\textbf{0.00} & \cellcolor[HTML]{EFEFEF}\textbf{0.36} \\
 & \multicolumn{1}{c|}{TENT} & 89.33 & 84.00 & 88.00 & 96.00 & 94.67 & 96.00 & 92.00 & 96.00 & 94.67 & 96.00 & 90.67 & 100.00 & 94.67 & 80.00 & 80.00 & 91.47 \\
 & \cellcolor[HTML]{EFEFEF}+MedBN & \cellcolor[HTML]{EFEFEF}\textbf{0.00} & \cellcolor[HTML]{EFEFEF}\textbf{1.33} & \cellcolor[HTML]{EFEFEF}\textbf{0.00} & \cellcolor[HTML]{EFEFEF}\textbf{0.00} & \cellcolor[HTML]{EFEFEF}\textbf{0.00} & \cellcolor[HTML]{EFEFEF}\textbf{0.00} & \cellcolor[HTML]{EFEFEF}\textbf{1.33} & \cellcolor[HTML]{EFEFEF}\textbf{0.00} & \cellcolor[HTML]{EFEFEF}\textbf{0.00} & \cellcolor[HTML]{EFEFEF}\textbf{0.00} & \cellcolor[HTML]{EFEFEF}\textbf{0.00} & \cellcolor[HTML]{EFEFEF}\textbf{4.00} & \cellcolor[HTML]{EFEFEF}\textbf{0.00} & \cellcolor[HTML]{EFEFEF}\textbf{0.00} & \cellcolor[HTML]{EFEFEF}\textbf{0.00} & \cellcolor[HTML]{EFEFEF}\textbf{0.44} \\
 & \multicolumn{1}{c|}{ETA} & 100.00 & 100.00 & 100.00 & 96.00 & 100.00 & 96.00 & 90.67 & 92.00 & 88.00 & 98.67 & 96.00 & 100.00 & 97.33 & 77.33 & 85.33 & 94.49 \\
 & \cellcolor[HTML]{EFEFEF}+MedBN & \cellcolor[HTML]{EFEFEF}\textbf{0.00} & \cellcolor[HTML]{EFEFEF}\textbf{0.00} & \cellcolor[HTML]{EFEFEF}\textbf{0.00} & \cellcolor[HTML]{EFEFEF}\textbf{0.00} & \cellcolor[HTML]{EFEFEF}\textbf{0.00} & \cellcolor[HTML]{EFEFEF}\textbf{0.00} & \cellcolor[HTML]{EFEFEF}\textbf{2.67} & \cellcolor[HTML]{EFEFEF}\textbf{0.00} & \cellcolor[HTML]{EFEFEF}\textbf{0.00} & \cellcolor[HTML]{EFEFEF}\textbf{0.00} & \cellcolor[HTML]{EFEFEF}\textbf{0.00} & \cellcolor[HTML]{EFEFEF}\textbf{4.00} & \cellcolor[HTML]{EFEFEF}\textbf{0.00} & \cellcolor[HTML]{EFEFEF}\textbf{0.00} & \cellcolor[HTML]{EFEFEF}\textbf{0.00} & \cellcolor[HTML]{EFEFEF}\textbf{0.44} \\
 & \multicolumn{1}{c|}{SAR} & 66.67 & 66.67 & 66.67 & 66.67 & 66.67 & 66.67 & 65.33 & 64.00 & 60.00 & 66.67 & 64.00 & 68.00 & 66.67 & 53.33 & 60.00 & 64.53 \\
 & \cellcolor[HTML]{EFEFEF}+MedBN & \cellcolor[HTML]{EFEFEF}\textbf{0.00} & \cellcolor[HTML]{EFEFEF}\textbf{0.00} & \cellcolor[HTML]{EFEFEF}\textbf{0.00} & \cellcolor[HTML]{EFEFEF}\textbf{1.33} & \cellcolor[HTML]{EFEFEF}\textbf{0.00} & \cellcolor[HTML]{EFEFEF}\textbf{0.00} & \cellcolor[HTML]{EFEFEF}\textbf{0.00} & \cellcolor[HTML]{EFEFEF}\textbf{0.00} & \cellcolor[HTML]{EFEFEF}\textbf{1.33} & \cellcolor[HTML]{EFEFEF}\textbf{0.00} & \cellcolor[HTML]{EFEFEF}\textbf{0.00} & \cellcolor[HTML]{EFEFEF}\textbf{2.67} & \cellcolor[HTML]{EFEFEF}\textbf{0.00} & \cellcolor[HTML]{EFEFEF}\textbf{1.33} & \cellcolor[HTML]{EFEFEF}\textbf{0.00} & \cellcolor[HTML]{EFEFEF}\textbf{0.44} \\
 & \multicolumn{1}{c|}{SoTTA} & 4.00 & 5.33 & 8.00 & 26.67 & 14.67 & 20.00 & 18.67 & 14.67 & 30.67 & 16.00 & 12.00 & 18.67 & 13.33 & 16.00 & 10.67 & 15.29 \\
 & \cellcolor[HTML]{EFEFEF}+MedBN & \cellcolor[HTML]{EFEFEF}\textbf{0.00} & \cellcolor[HTML]{EFEFEF}\textbf{0.00} & \cellcolor[HTML]{EFEFEF}\textbf{0.00} & \cellcolor[HTML]{EFEFEF}\textbf{0.00} & \cellcolor[HTML]{EFEFEF}\textbf{0.00} & \cellcolor[HTML]{EFEFEF}\textbf{0.00} & \cellcolor[HTML]{EFEFEF}\textbf{1.33} & \cellcolor[HTML]{EFEFEF}\textbf{2.67} & \cellcolor[HTML]{EFEFEF}\textbf{4.00} & \cellcolor[HTML]{EFEFEF}\textbf{0.00} & \cellcolor[HTML]{EFEFEF}\textbf{0.00} & \cellcolor[HTML]{EFEFEF}\textbf{4.00} & \cellcolor[HTML]{EFEFEF}\textbf{0.00} & \cellcolor[HTML]{EFEFEF}\textbf{0.00} & \cellcolor[HTML]{EFEFEF}\textbf{0.00} & \cellcolor[HTML]{EFEFEF}\textbf{0.80} \\
 & \multicolumn{1}{c|}{sEMA} & 0.00 & 8.00 & 0.00 & 16.00 & 8.00 & 20.00 & 16.00 & 8.00 & 17.33 & 24.00 & 4.00 & 17.33 & 8.00 & 12.00 & 6.67 & 11.02 \\
 & \cellcolor[HTML]{EFEFEF}+MedBN & \cellcolor[HTML]{EFEFEF}\textbf{0.00} & \cellcolor[HTML]{EFEFEF}\textbf{0.00} & \cellcolor[HTML]{EFEFEF}\textbf{0.00} & \cellcolor[HTML]{EFEFEF}\textbf{4.00} & \cellcolor[HTML]{EFEFEF}\textbf{0.00} & \cellcolor[HTML]{EFEFEF}\textbf{0.00} & \cellcolor[HTML]{EFEFEF}\textbf{0.00} & \cellcolor[HTML]{EFEFEF}\textbf{0.00} & \cellcolor[HTML]{EFEFEF}\textbf{0.00} & \cellcolor[HTML]{EFEFEF}\textbf{0.00} & \cellcolor[HTML]{EFEFEF}\textbf{0.00} & \cellcolor[HTML]{EFEFEF}\textbf{0.00} & \cellcolor[HTML]{EFEFEF}\textbf{0.00} & \cellcolor[HTML]{EFEFEF}\textbf{0.00} & \cellcolor[HTML]{EFEFEF}\textbf{0.00} & \cellcolor[HTML]{EFEFEF}\textbf{0.27} \\
 & \multicolumn{1}{c|}{mDIA} & 24.00 & 22.67 & 32.00 & 37.33 & 26.67 & 36.00 & 36.00 & 25.33 & 45.33 & 40.00 & 26.67 & 57.33 & 28.00 & 21.33 & 24.00 & 32.18 \\
 & \cellcolor[HTML]{EFEFEF}+MedBN & \cellcolor[HTML]{EFEFEF}\textbf{8.00} & \cellcolor[HTML]{EFEFEF}\textbf{0.00} & \cellcolor[HTML]{EFEFEF}\textbf{0.00} & \cellcolor[HTML]{EFEFEF}\textbf{0.00} & \cellcolor[HTML]{EFEFEF}\textbf{0.00} & \cellcolor[HTML]{EFEFEF}\textbf{0.00} & \cellcolor[HTML]{EFEFEF}\textbf{0.00} & \cellcolor[HTML]{EFEFEF}\textbf{0.00} & \cellcolor[HTML]{EFEFEF}\textbf{4.00} & \cellcolor[HTML]{EFEFEF}\textbf{0.00} & \cellcolor[HTML]{EFEFEF}\textbf{0.00} & \cellcolor[HTML]{EFEFEF}\textbf{0.00} & \cellcolor[HTML]{EFEFEF}\textbf{0.00} & \cellcolor[HTML]{EFEFEF}\textbf{4.00} & \cellcolor[HTML]{EFEFEF}\textbf{0.00} & \cellcolor[HTML]{EFEFEF}\textbf{1.07} \\
\bottomrule
\end{tabular}
\end{adjustbox}
\end{table*}

\begin{table*}[!ht]\centering
\caption{Extended analysis of Error Rate (\%) of indiscriminate and instant attack scenario over all types of corruption (detailed version
of Table~\ref{tab:ins:indis:avg}).
} \label{tab:ins:indis}
\begin{adjustbox}{width=1.0\textwidth}
\begin{tabular}
{l|l|cccccccccccccccc}
\toprule
& & \multicolumn{3}{c}{Noise} & \multicolumn{4}{c}{Blur} & \multicolumn{4}{c}{Weather} & \multicolumn{4}{c}{Digital} &     
\\
\cmidrule(lr){3-5} \cmidrule(lr){6-9} \cmidrule(lr){10-13} \cmidrule(lr){14-17} & \multicolumn{1}{c|}{Method} & Gauss. & Shot & Impul. & Defoc. & Glass & Motion & Zoom & Snow & Frost & Fog & Brit. & Contr. & Elastic & Pixel. & JPEG & Avg. \\ 
\midrule
\multirow{14}{*}{\rotatebox[origin=c]{90}{CIFAR10-C}} & 
\multicolumn{1}{c|}{TeBN} & 35.74 & 34.37 & 45.60 & 23.88 & 37.95 & 27.86 & 20.62 & 29.76 & 26.83 & 40.62 & 20.82 & 30.97 & 32.95 & 23.65 & 33.70 & 31.02 \\
& \cellcolor[HTML]{EFEFEF}+MedBN & \cellcolor[HTML]{EFEFEF}\textbf{26.60} & \cellcolor[HTML]{EFEFEF}\textbf{25.15} & \cellcolor[HTML]{EFEFEF}\textbf{35.01} & \cellcolor[HTML]{EFEFEF}\textbf{17.23} & \cellcolor[HTML]{EFEFEF}\textbf{27.51} & \cellcolor[HTML]{EFEFEF}\textbf{20.37} & \cellcolor[HTML]{EFEFEF}\textbf{14.75} & \cellcolor[HTML]{EFEFEF}\textbf{21.00} & \cellcolor[HTML]{EFEFEF}\textbf{18.36} & \cellcolor[HTML]{EFEFEF}\textbf{28.19} & \cellcolor[HTML]{EFEFEF}\textbf{14.24} & \cellcolor[HTML]{EFEFEF}\textbf{20.22} & \cellcolor[HTML]{EFEFEF}\textbf{24.93} & \cellcolor[HTML]{EFEFEF}\textbf{16.82} & \cellcolor[HTML]{EFEFEF}\textbf{24.74} & \cellcolor[HTML]{EFEFEF}\textbf{22.34} \\
 
& \multicolumn{1}{c|}{TENT} & 32.20 & 30.72 & 40.39 & 23.05 & 34.92 & 25.91 & 20.62 & 26.91 & 25.64 & 28.47 & 20.66 & 26.90 & 31.42 & 23.93 & 30.14 & 28.13 \\
& \cellcolor[HTML]{EFEFEF}+MedBN & \cellcolor[HTML]{EFEFEF}\textbf{24.12} & \cellcolor[HTML]{EFEFEF}\textbf{22.96} & \cellcolor[HTML]{EFEFEF}\textbf{32.31} & \cellcolor[HTML]{EFEFEF}\textbf{15.78} & \cellcolor[HTML]{EFEFEF}\textbf{27.08} & \cellcolor[HTML]{EFEFEF}\textbf{17.99} & \cellcolor[HTML]{EFEFEF}\textbf{14.59} & \cellcolor[HTML]{EFEFEF}\textbf{19.07} & \cellcolor[HTML]{EFEFEF}\textbf{17.05} & \cellcolor[HTML]{EFEFEF}\textbf{20.22} & \cellcolor[HTML]{EFEFEF}\textbf{13.87} & \cellcolor[HTML]{EFEFEF}\textbf{18.04} & \cellcolor[HTML]{EFEFEF}\textbf{23.88} & \cellcolor[HTML]{EFEFEF}\textbf{15.23} & \cellcolor[HTML]{EFEFEF}\textbf{22.30} & \cellcolor[HTML]{EFEFEF}\textbf{20.30} \\
& \multicolumn{1}{c|}{ETA} & 30.67 & 30.05 & 38.84 & 22.44 & 34.10 & 25.15 & 20.76 & 26.46 & 25.42 & 27.99 & 20.50 & 26.09 & 30.65 & 22.67 & 29.46 & 27.42 \\
& \cellcolor[HTML]{EFEFEF}+MedBN & \cellcolor[HTML]{EFEFEF}\textbf{23.42} & \cellcolor[HTML]{EFEFEF}\textbf{21.93} & \cellcolor[HTML]{EFEFEF}\textbf{31.02} & \cellcolor[HTML]{EFEFEF}\textbf{15.04} & \cellcolor[HTML]{EFEFEF}\textbf{26.37} & \cellcolor[HTML]{EFEFEF}\textbf{17.62} & \cellcolor[HTML]{EFEFEF}\textbf{14.20} & \cellcolor[HTML]{EFEFEF}\textbf{18.37} & \cellcolor[HTML]{EFEFEF}\textbf{16.93} & \cellcolor[HTML]{EFEFEF}\textbf{20.13} & \cellcolor[HTML]{EFEFEF}\textbf{13.60} & \cellcolor[HTML]{EFEFEF}\textbf{17.78} & \cellcolor[HTML]{EFEFEF}\textbf{23.12} & \cellcolor[HTML]{EFEFEF}\textbf{14.89} & \cellcolor[HTML]{EFEFEF}\textbf{22.70} & \cellcolor[HTML]{EFEFEF}\textbf{19.81} \\
& \multicolumn{1}{c|}{SAR} & 32.24 & 30.30 & 39.30 & 22.18 & 33.52 & 25.44 & 20.08 & 25.71 & 25.18 & 29.38 & 20.16 & 27.57 & 30.20 & 22.33 & 29.85 & 27.56 \\
& \cellcolor[HTML]{EFEFEF}+MedBN & \cellcolor[HTML]{EFEFEF}\textbf{23.36} & \cellcolor[HTML]{EFEFEF}\textbf{21.73} & \cellcolor[HTML]{EFEFEF}\textbf{30.87} & \cellcolor[HTML]{EFEFEF}\textbf{14.96} & \cellcolor[HTML]{EFEFEF}\textbf{25.14} & \cellcolor[HTML]{EFEFEF}\textbf{17.45} & \cellcolor[HTML]{EFEFEF}\textbf{13.84} & \cellcolor[HTML]{EFEFEF}\textbf{18.20} & \cellcolor[HTML]{EFEFEF}\textbf{16.61} & \cellcolor[HTML]{EFEFEF}\textbf{20.24} & \cellcolor[HTML]{EFEFEF}\textbf{13.57} & \cellcolor[HTML]{EFEFEF}\textbf{18.25} & \cellcolor[HTML]{EFEFEF}\textbf{22.86} & \cellcolor[HTML]{EFEFEF}\textbf{15.01} & \cellcolor[HTML]{EFEFEF}\textbf{21.90} & \cellcolor[HTML]{EFEFEF}\textbf{19.60} \\
& \multicolumn{1}{c|}{SoTTA} & 25.49 & 23.51 & 33.12 & 15.12 & 26.87 & 17.30 & 13.53 & 19.09 & 17.34 & 20.23 & 14.27 & 17.56 & 23.70 & 15.23 & 23.62 & 20.40 \\
& \cellcolor[HTML]{EFEFEF}+MedBN & \cellcolor[HTML]{EFEFEF}\textbf{21.37} & \cellcolor[HTML]{EFEFEF}\textbf{19.78} & \cellcolor[HTML]{EFEFEF}\textbf{29.50} & \cellcolor[HTML]{EFEFEF}\textbf{11.15} & \cellcolor[HTML]{EFEFEF}\textbf{23.23} & \cellcolor[HTML]{EFEFEF}\textbf{13.67} & \cellcolor[HTML]{EFEFEF}\textbf{10.01} & \cellcolor[HTML]{EFEFEF}\textbf{15.00} & \cellcolor[HTML]{EFEFEF}\textbf{14.15} & \cellcolor[HTML]{EFEFEF}\textbf{14.90} & \cellcolor[HTML]{EFEFEF}\textbf{10.66} & \cellcolor[HTML]{EFEFEF}\textbf{13.06} & \cellcolor[HTML]{EFEFEF}\textbf{19.17} & \cellcolor[HTML]{EFEFEF}\textbf{12.14} & \cellcolor[HTML]{EFEFEF}\textbf{19.52} & \cellcolor[HTML]{EFEFEF}\textbf{16.49} \\
& \multicolumn{1}{c|}{sEMA} & 27.22 & 25.65 & 37.32 & 14.40 & 27.90 & 18.43 & 12.09 & 20.28 & 17.76 & 26.50 & 13.11 & 19.35 & 23.60 & 15.08 & 26.01 & 21.65 \\
& \cellcolor[HTML]{EFEFEF}+MedBN & \cellcolor[HTML]{EFEFEF}\textbf{23.68} & \cellcolor[HTML]{EFEFEF}\textbf{22.09} & \cellcolor[HTML]{EFEFEF}\textbf{33.07} & \cellcolor[HTML]{EFEFEF}\textbf{11.00} & \cellcolor[HTML]{EFEFEF}\textbf{23.16} & \cellcolor[HTML]{EFEFEF}\textbf{14.42} & \cellcolor[HTML]{EFEFEF}\textbf{9.18} & \cellcolor[HTML]{EFEFEF}\textbf{16.32} & \cellcolor[HTML]{EFEFEF}\textbf{14.10} & \cellcolor[HTML]{EFEFEF}\textbf{20.85} & \cellcolor[HTML]{EFEFEF}\textbf{10.58} & \cellcolor[HTML]{EFEFEF}\textbf{14.73} & \cellcolor[HTML]{EFEFEF}\textbf{19.32} & \cellcolor[HTML]{EFEFEF}\textbf{12.35} & \cellcolor[HTML]{EFEFEF}\textbf{21.76} & \cellcolor[HTML]{EFEFEF}\textbf{17.77} \\
& \multicolumn{1}{c|}{mDIA} & 38.78 & 37.39 & 50.60 & 16.82 & 34.20 & 23.57 & 14.23 & 22.99 & 19.96 & 38.50 & 13.61 & 36.23 & 27.04 & 18.00 & 27.43 & 27.96 \\
& \cellcolor[HTML]{EFEFEF}+MedBN & \cellcolor[HTML]{EFEFEF}\textbf{26.26} & \cellcolor[HTML]{EFEFEF}\textbf{25.26} & \cellcolor[HTML]{EFEFEF}\textbf{34.70} & \cellcolor[HTML]{EFEFEF}\textbf{12.03} & \cellcolor[HTML]{EFEFEF}\textbf{24.90} & \cellcolor[HTML]{EFEFEF}\textbf{15.97} & \cellcolor[HTML]{EFEFEF}\textbf{9.60} & \cellcolor[HTML]{EFEFEF}\textbf{16.60} & \cellcolor[HTML]{EFEFEF}\textbf{13.69} & \cellcolor[HTML]{EFEFEF}\textbf{22.20} & \cellcolor[HTML]{EFEFEF}\textbf{9.86} & \cellcolor[HTML]{EFEFEF}\textbf{20.53} & \cellcolor[HTML]{EFEFEF}\textbf{19.72} & \cellcolor[HTML]{EFEFEF}\textbf{13.25} & \cellcolor[HTML]{EFEFEF}\textbf{21.37} & \cellcolor[HTML]{EFEFEF}\textbf{19.06} \\

\midrule
\multirow{14}{*}{\rotatebox[origin=c]{90}{CIFAR100-C}}& 
\multicolumn{1}{c|}{TeBN} & 66.56 & 65.76 & 73.97 & 49.61 & 64.98 & 55.07 & 45.74 & 59.85 & 57.30 & 71.08 & 50.01 & 64.44 & 60.07 & 49.70 & 62.91 & 59.80 \\
& \cellcolor[HTML]{EFEFEF}+MedBN & \cellcolor[HTML]{EFEFEF}\textbf{54.39} & \cellcolor[HTML]{EFEFEF}\textbf{54.82} & \cellcolor[HTML]{EFEFEF}\textbf{63.47} & \cellcolor[HTML]{EFEFEF}\textbf{39.26} & \cellcolor[HTML]{EFEFEF}\textbf{53.64} & \cellcolor[HTML]{EFEFEF}\textbf{43.44} & \cellcolor[HTML]{EFEFEF}\textbf{36.08} & \cellcolor[HTML]{EFEFEF}\textbf{48.36} & \cellcolor[HTML]{EFEFEF}\textbf{45.71} & \cellcolor[HTML]{EFEFEF}\textbf{58.51} & \cellcolor[HTML]{EFEFEF}\textbf{39.71} & \cellcolor[HTML]{EFEFEF}\textbf{50.15} & \cellcolor[HTML]{EFEFEF}\textbf{49.33} & \cellcolor[HTML]{EFEFEF}\textbf{40.22} & \cellcolor[HTML]{EFEFEF}\textbf{51.15} & \cellcolor[HTML]{EFEFEF}\textbf{48.55} \\
& \multicolumn{1}{c|}{TENT} & 60.80 & 59.45 & 66.79 & 47.84 & 60.47 & 51.82 & 45.31 & 55.78 & 55.21 & 58.65 & 47.08 & 54.01 & 57.22 & 48.24 & 57.75 & 55.10 \\
& \cellcolor[HTML]{EFEFEF}+MedBN & \cellcolor[HTML]{EFEFEF}\textbf{53.77} & \cellcolor[HTML]{EFEFEF}\textbf{52.26} & \cellcolor[HTML]{EFEFEF}\textbf{60.89} & \cellcolor[HTML]{EFEFEF}\textbf{39.68} & \cellcolor[HTML]{EFEFEF}\textbf{53.47} & \cellcolor[HTML]{EFEFEF}\textbf{42.24} & \cellcolor[HTML]{EFEFEF}\textbf{37.10} & \cellcolor[HTML]{EFEFEF}\textbf{46.99} & \cellcolor[HTML]{EFEFEF}\textbf{45.95} & \cellcolor[HTML]{EFEFEF}\textbf{49.69} & \cellcolor[HTML]{EFEFEF}\textbf{38.82} & \cellcolor[HTML]{EFEFEF}\textbf{45.33} & \cellcolor[HTML]{EFEFEF}\textbf{48.19} & \cellcolor[HTML]{EFEFEF}\textbf{40.08} & \cellcolor[HTML]{EFEFEF}\textbf{49.99} & \cellcolor[HTML]{EFEFEF}\textbf{46.96} \\
& \multicolumn{1}{c|}{ETA} & 60.50 & 58.90 & 67.04 & 47.27 & 60.60 & 50.97 & 44.07 & 54.27 & 54.83 & 57.19 & 46.33 & 54.64 & 56.07 & 48.23 & 55.80 & 54.45 \\
& \cellcolor[HTML]{EFEFEF}+MedBN & \cellcolor[HTML]{EFEFEF}\textbf{53.38} & \cellcolor[HTML]{EFEFEF}\textbf{52.40} & \cellcolor[HTML]{EFEFEF}\textbf{61.82} & \cellcolor[HTML]{EFEFEF}\textbf{39.34} & \cellcolor[HTML]{EFEFEF}\textbf{52.52} & \cellcolor[HTML]{EFEFEF}\textbf{42.05} & \cellcolor[HTML]{EFEFEF}\textbf{36.35} & \cellcolor[HTML]{EFEFEF}\textbf{45.77} & \cellcolor[HTML]{EFEFEF}\textbf{45.05} & \cellcolor[HTML]{EFEFEF}\textbf{49.99} & \cellcolor[HTML]{EFEFEF}\textbf{38.04} & \cellcolor[HTML]{EFEFEF}\textbf{44.60} & \cellcolor[HTML]{EFEFEF}\textbf{48.07} & \cellcolor[HTML]{EFEFEF}\textbf{39.76} & \cellcolor[HTML]{EFEFEF}\textbf{49.78} & \cellcolor[HTML]{EFEFEF}\textbf{46.59} \\
& \multicolumn{1}{c|}{SAR} & 63.51 & 61.82 & 69.44 & 48.43 & 61.60 & 52.69 & 46.33 & 55.21 & 55.38 & 59.57 & 48.31 & 57.58 & 57.68 & 48.99 & 59.49 & 56.40 \\
& \cellcolor[HTML]{EFEFEF}+MedBN & \cellcolor[HTML]{EFEFEF}\textbf{55.16} & \cellcolor[HTML]{EFEFEF}\textbf{53.88} & \cellcolor[HTML]{EFEFEF}\textbf{63.38} & \cellcolor[HTML]{EFEFEF}\textbf{39.98} & \cellcolor[HTML]{EFEFEF}\textbf{53.59} & \cellcolor[HTML]{EFEFEF}\textbf{42.99} & \cellcolor[HTML]{EFEFEF}\textbf{36.49} & \cellcolor[HTML]{EFEFEF}\textbf{47.42} & \cellcolor[HTML]{EFEFEF}\textbf{46.59} & \cellcolor[HTML]{EFEFEF}\textbf{52.34} & \cellcolor[HTML]{EFEFEF}\textbf{39.22} & \cellcolor[HTML]{EFEFEF}\textbf{46.52} & \cellcolor[HTML]{EFEFEF}\textbf{49.95} & \cellcolor[HTML]{EFEFEF}\textbf{41.02} & \cellcolor[HTML]{EFEFEF}\textbf{51.51} & \cellcolor[HTML]{EFEFEF}\textbf{48.00} \\
& \multicolumn{1}{c|}{SoTTA} & 54.94 & 54.19 & 63.48 & 40.10 & 54.22 & 44.14 & 38.49 & 48.03 & 48.08 & 50.45 & 39.36 & 46.16 & 50.12 & 41.85 & 51.31 & 48.33 \\
& \cellcolor[HTML]{EFEFEF}+MedBN & \cellcolor[HTML]{EFEFEF}\textbf{53.52} & \cellcolor[HTML]{EFEFEF}\textbf{51.94} & \cellcolor[HTML]{EFEFEF}\textbf{61.14} & \cellcolor[HTML]{EFEFEF}\textbf{37.24} & \cellcolor[HTML]{EFEFEF}\textbf{51.32} & \cellcolor[HTML]{EFEFEF}\textbf{40.71} & \cellcolor[HTML]{EFEFEF}\textbf{35.23} & \cellcolor[HTML]{EFEFEF}\textbf{45.16} & \cellcolor[HTML]{EFEFEF}\textbf{44.68} & \cellcolor[HTML]{EFEFEF}\textbf{46.67} & \cellcolor[HTML]{EFEFEF}\textbf{37.09} & \cellcolor[HTML]{EFEFEF}\textbf{40.82} & \cellcolor[HTML]{EFEFEF}\textbf{47.57} & \cellcolor[HTML]{EFEFEF}\textbf{38.51} & \cellcolor[HTML]{EFEFEF}\textbf{49.05} & \cellcolor[HTML]{EFEFEF}\textbf{45.38} \\
& \multicolumn{1}{c|}{sEMA} & 53.50 & 52.80 & 63.03 & 37.82 & 51.65 & 41.65 & 34.07 & 46.85 & 43.91 & 56.24 & 37.94 & 47.58 & 47.00 & 37.32 & 52.04 & 46.89 \\
& \cellcolor[HTML]{EFEFEF}+MedBN & \cellcolor[HTML]{EFEFEF}\textbf{50.87} & \cellcolor[HTML]{EFEFEF}\textbf{51.05} & \cellcolor[HTML]{EFEFEF}\textbf{59.57} & \cellcolor[HTML]{EFEFEF}\textbf{33.60} & \cellcolor[HTML]{EFEFEF}\textbf{48.07} & \cellcolor[HTML]{EFEFEF}\textbf{37.50} & \cellcolor[HTML]{EFEFEF}\textbf{30.90} & \cellcolor[HTML]{EFEFEF}\textbf{42.95} & \cellcolor[HTML]{EFEFEF}\textbf{40.52} & \cellcolor[HTML]{EFEFEF}\textbf{52.24} & \cellcolor[HTML]{EFEFEF}\textbf{34.01} & \cellcolor[HTML]{EFEFEF}\textbf{42.28} & \cellcolor[HTML]{EFEFEF}\textbf{43.89} & \cellcolor[HTML]{EFEFEF}\textbf{34.51} & \cellcolor[HTML]{EFEFEF}\textbf{48.30} & \cellcolor[HTML]{EFEFEF}\textbf{43.35} \\
& \multicolumn{1}{c|}{mDIA} & 64.90 & 63.06 & 72.89 & 43.07 & 59.61 & 49.40 & 38.27 & 50.80 & 48.32 & 72.11 & 40.45 & 77.57 & 52.47 & 44.45 & 54.01 & 55.43 \\
& \cellcolor[HTML]{EFEFEF}+MedBN & \cellcolor[HTML]{EFEFEF}\textbf{59.93} & \cellcolor[HTML]{EFEFEF}\textbf{57.65} & \cellcolor[HTML]{EFEFEF}\textbf{67.90} & \cellcolor[HTML]{EFEFEF}\textbf{36.14} & \cellcolor[HTML]{EFEFEF}\textbf{52.96} & \cellcolor[HTML]{EFEFEF}\textbf{40.41} & \cellcolor[HTML]{EFEFEF}\textbf{32.92} & \cellcolor[HTML]{EFEFEF}\textbf{44.47} & \cellcolor[HTML]{EFEFEF}\textbf{41.89} & \cellcolor[HTML]{EFEFEF}\textbf{56.29} & \cellcolor[HTML]{EFEFEF}\textbf{34.97} & \cellcolor[HTML]{EFEFEF}\textbf{57.40} & \cellcolor[HTML]{EFEFEF}\textbf{45.85} & \cellcolor[HTML]{EFEFEF}\textbf{39.03} & \cellcolor[HTML]{EFEFEF}\textbf{49.77} & \cellcolor[HTML]{EFEFEF}\textbf{47.84} 
\\
 
\midrule
\multirow{14}{*}{\rotatebox[origin=c]{90}{ImageNet-C}}& \multicolumn{1}{c|}{TeBN} & 96.96 & 94.46 & 96.52 & 93.47 & 93.50 & 85.97 & 75.93 & 80.93 & 83.13 & 69.50 & 49.99 & 96.58 & 70.21 & 63.28 & 71.45 & 81.46 \\
& \cellcolor[HTML]{EFEFEF}+MedBN & \cellcolor[HTML]{EFEFEF}\textbf{86.31} & \cellcolor[HTML]{EFEFEF}\textbf{85.25} & \cellcolor[HTML]{EFEFEF}\textbf{85.36} & \cellcolor[HTML]{EFEFEF}\textbf{83.69} & \cellcolor[HTML]{EFEFEF}\textbf{84.72} & \cellcolor[HTML]{EFEFEF}\textbf{75.56} & \cellcolor[HTML]{EFEFEF}\textbf{64.33} & \cellcolor[HTML]{EFEFEF}\textbf{68.46} & \cellcolor[HTML]{EFEFEF}\textbf{69.99} & \cellcolor[HTML]{EFEFEF}\textbf{52.80} & \cellcolor[HTML]{EFEFEF}\textbf{37.25} & \cellcolor[HTML]{EFEFEF}\textbf{77.99} & \cellcolor[HTML]{EFEFEF}\textbf{60.00} & \cellcolor[HTML]{EFEFEF}\textbf{52.70} & \cellcolor[HTML]{EFEFEF}\textbf{61.63} & \cellcolor[HTML]{EFEFEF}\textbf{69.74} \\
& \multicolumn{1}{c|}{TENT} & 86.54 & 84.05 & 86.24 & 86.71 & 86.08 & 77.19 & 66.77 & 69.92 & 74.90 & 60.79 & 46.15 & 89.35 & 60.77 & 55.25 & 61.64 & 72.82 \\
& \cellcolor[HTML]{EFEFEF}+MedBN & \cellcolor[HTML]{EFEFEF}\textbf{84.87} & \cellcolor[HTML]{EFEFEF}\textbf{83.07} & \cellcolor[HTML]{EFEFEF}\textbf{84.07} & \cellcolor[HTML]{EFEFEF}\textbf{82.48} & \cellcolor[HTML]{EFEFEF}\textbf{83.39} & \cellcolor[HTML]{EFEFEF}\textbf{73.50} & \cellcolor[HTML]{EFEFEF}\textbf{62.07} & \cellcolor[HTML]{EFEFEF}\textbf{66.89} & \cellcolor[HTML]{EFEFEF}\textbf{68.89} & \cellcolor[HTML]{EFEFEF}\textbf{51.07} & \cellcolor[HTML]{EFEFEF}\textbf{35.86} & \cellcolor[HTML]{EFEFEF}\textbf{77.04} & \cellcolor[HTML]{EFEFEF}\textbf{57.18} & \cellcolor[HTML]{EFEFEF}\textbf{50.36} & \cellcolor[HTML]{EFEFEF}\textbf{59.44} & \cellcolor[HTML]{EFEFEF}\textbf{68.01} \\
& \multicolumn{1}{c|}{ETA} & 90.81 & 88.00 & 90.76 & 88.37 & 87.61 & 77.50 & 67.86 & 71.07 & 75.13 & 61.63 & 46.11 & 87.50 & 61.45 & 55.63 & 62.79 & 74.15 \\
& \cellcolor[HTML]{EFEFEF}+MedBN & \cellcolor[HTML]{EFEFEF}\textbf{85.90} & \cellcolor[HTML]{EFEFEF}\textbf{84.41} & \cellcolor[HTML]{EFEFEF}\textbf{84.86} & \cellcolor[HTML]{EFEFEF}\textbf{82.99} & \cellcolor[HTML]{EFEFEF}\textbf{83.92} & \cellcolor[HTML]{EFEFEF}\textbf{74.37} & \cellcolor[HTML]{EFEFEF}\textbf{62.18} & \cellcolor[HTML]{EFEFEF}\textbf{66.47} & \cellcolor[HTML]{EFEFEF}\textbf{68.69} & \cellcolor[HTML]{EFEFEF}\textbf{51.73} & \cellcolor[HTML]{EFEFEF}\textbf{35.98} & \cellcolor[HTML]{EFEFEF}\textbf{77.03} & \cellcolor[HTML]{EFEFEF}\textbf{57.88} & \cellcolor[HTML]{EFEFEF}\textbf{51.08} & \cellcolor[HTML]{EFEFEF}\textbf{59.63} & \cellcolor[HTML]{EFEFEF}\textbf{68.47} \\
& \multicolumn{1}{c|}{SAR} & 95.42 & 92.74 & 94.83 & 92.15 & 91.16 & 82.32 & 72.15 & 75.73 & 77.79 & 63.24 & 46.52 & 92.92 & 64.96 & 58.24 & 65.93 & 77.74 \\
& \cellcolor[HTML]{EFEFEF}+MedBN & \cellcolor[HTML]{EFEFEF}\textbf{86.00} & \cellcolor[HTML]{EFEFEF}\textbf{85.09} & \cellcolor[HTML]{EFEFEF}\textbf{85.73} & \cellcolor[HTML]{EFEFEF}\textbf{84.02} & \cellcolor[HTML]{EFEFEF}\textbf{84.79} & \cellcolor[HTML]{EFEFEF}\textbf{74.92} & \cellcolor[HTML]{EFEFEF}\textbf{63.99} & \cellcolor[HTML]{EFEFEF}\textbf{68.24} & \cellcolor[HTML]{EFEFEF}\textbf{69.81} & \cellcolor[HTML]{EFEFEF}\textbf{52.65} & \cellcolor[HTML]{EFEFEF}\textbf{36.76} & \cellcolor[HTML]{EFEFEF}\textbf{78.36} & \cellcolor[HTML]{EFEFEF}\textbf{59.10} & \cellcolor[HTML]{EFEFEF}\textbf{52.29} & \cellcolor[HTML]{EFEFEF}\textbf{61.29} & \cellcolor[HTML]{EFEFEF}\textbf{69.54}\\

& \multicolumn{1}{c|}{SoTTA} & 78.69 & 78.14 & 79.30 & 80.96 & 80.99 & 70.53 & 60.11 & 63.37 & 68.14 & 52.59 & 39.32 & 77.96 & 55.30 & 49.95 & 55.41 & 66.05 \\
& \cellcolor[HTML]{EFEFEF}+MedBN & \cellcolor[HTML]{EFEFEF}\textbf{80.41} & \cellcolor[HTML]{EFEFEF}\textbf{81.56} & \cellcolor[HTML]{EFEFEF}\textbf{80.16} & \cellcolor[HTML]{EFEFEF}\textbf{78.65} & \cellcolor[HTML]{EFEFEF}\textbf{79.72} & \cellcolor[HTML]{EFEFEF}\textbf{69.13} & \cellcolor[HTML]{EFEFEF}\textbf{57.18} & \cellcolor[HTML]{EFEFEF}\textbf{61.41} & \cellcolor[HTML]{EFEFEF}\textbf{65.41} & \cellcolor[HTML]{EFEFEF}\textbf{48.27} & \cellcolor[HTML]{EFEFEF}\textbf{34.53} & \cellcolor[HTML]{EFEFEF}\textbf{72.69} & \cellcolor[HTML]{EFEFEF}\textbf{53.19} & \cellcolor[HTML]{EFEFEF}\textbf{47.10} & \cellcolor[HTML]{EFEFEF}\textbf{53.84} & \cellcolor[HTML]{EFEFEF}\textbf{64.22} \\
& \multicolumn{1}{c|}{sEMA} & 86.24 & 86.50 & 85.62 & 87.26 & 87.36 & 78.84 & 68.13 & 72.04 & 74.37 & 59.49 & 41.89 & 86.65 & 62.88 & 55.81 & 65.12 & 73.21 \\
& \cellcolor[HTML]{EFEFEF}+MedBN & \cellcolor[HTML]{EFEFEF}\textbf{86.44} & \cellcolor[HTML]{EFEFEF}\textbf{87.33} & \cellcolor[HTML]{EFEFEF}\textbf{85.73} & \cellcolor[HTML]{EFEFEF}\textbf{84.87} & \cellcolor[HTML]{EFEFEF}\textbf{84.66} & \cellcolor[HTML]{EFEFEF}\textbf{75.85} & \cellcolor[HTML]{EFEFEF}\textbf{64.19} & \cellcolor[HTML]{EFEFEF}\textbf{68.71} & \cellcolor[HTML]{EFEFEF}\textbf{69.62} & \cellcolor[HTML]{EFEFEF}\textbf{53.84} & \cellcolor[HTML]{EFEFEF}\textbf{36.90} & \cellcolor[HTML]{EFEFEF}\textbf{81.81} & \cellcolor[HTML]{EFEFEF}\textbf{58.58} & \cellcolor[HTML]{EFEFEF}\textbf{53.39} & \cellcolor[HTML]{EFEFEF}\textbf{61.42} & \cellcolor[HTML]{EFEFEF}\textbf{70.22} \\
& \multicolumn{1}{c|}{mDIA} & 93.77 & 92.77 & 93.22 & 87.22 & 87.50 & 80.57 & 73.80 & 77.39 & 76.65 & 67.41 & 43.81 & 90.10 & 65.74 & 61.76 & 67.51 & 77.28 \\
& \cellcolor[HTML]{EFEFEF}+MedBN & \cellcolor[HTML]{EFEFEF}\textbf{87.92} & \cellcolor[HTML]{EFEFEF}\textbf{86.90} & \cellcolor[HTML]{EFEFEF}\textbf{86.56} & \cellcolor[HTML]{EFEFEF}\textbf{79.84} & \cellcolor[HTML]{EFEFEF}\textbf{81.54} & \cellcolor[HTML]{EFEFEF}\textbf{73.44} & \cellcolor[HTML]{EFEFEF}\textbf{64.99} & \cellcolor[HTML]{EFEFEF}\textbf{68.14} & \cellcolor[HTML]{EFEFEF}\textbf{67.81} & \cellcolor[HTML]{EFEFEF}\textbf{52.44} & \cellcolor[HTML]{EFEFEF}\textbf{36.58} & \cellcolor[HTML]{EFEFEF}\textbf{78.04} & \cellcolor[HTML]{EFEFEF}\textbf{58.99} & \cellcolor[HTML]{EFEFEF}\textbf{55.19} & \cellcolor[HTML]{EFEFEF}\textbf{60.15} & \cellcolor[HTML]{EFEFEF}\textbf{69.24} \\
\bottomrule
\end{tabular}
\end{adjustbox}
\end{table*}


\clearpage

\section{Comprehensive Results of Cumulative Attack Scenario}\label{appendix:cum:attack}

We provide detailed results of cumulative and targeted attack scenarios in Table~\ref{tab:cum:target} and cumulative and indiscriminate attack scenarios in Table~\ref{tab:cum:indis} across all types of corruptions in the TTA benchmark datasets. The averaged results across all trials are presented in Table~\ref{tab:cum:avg}.
Within the scope of the cumulative attack scenario, we use EATA instead of ETA. EATA includes a Fisher regularizer that limits substantial change to important parameters, offering benefits in the cumulative scenario.

\begin{table*}[!ht]
\centering
\caption{Attack Success Rate (\%) of the targeted and cumulative attack scenario over all types of corruptions (full version of Table~\ref{tab:cum:avg}).} \label{tab:cum:target}
\begin{adjustbox}{width=1.0\textwidth}
\begin{tabular}
{l|l|ccccccccccccccccc}
\toprule
& & \multicolumn{3}{c}{Noise} & \multicolumn{4}{c}{Blur} & \multicolumn{4}{c}{Weather} & \multicolumn{4}{c}{Digital} &     
\\
\cmidrule(lr){3-5} \cmidrule(lr){6-9} \cmidrule(lr){10-13} \cmidrule(lr){14-17}
& \multicolumn{1}{c|}{Method} & Gauss. & Shot & Impul. & Defoc. & Glass & Motion & Zoom & Snow & Frost & Fog & Brit. & Contr. & Elastic & Pixel. & JPEG & Avg. \\ 
\midrule

\multirow{14}{*}{\rotatebox[origin=c]{90}{CIFAR10-C}} 
& \multicolumn{1}{c|}{TeBN} & 82.67 & 90.00 & 90.67 & 76.00 & 94.67 & 82.67 & 78.67 & 82.67 & 84.67 & 88.67 & 74.67 & 92.00 & 87.33 & 76.67 & 78.67 & 84.04 \\
& \multicolumn{1}{c|}{\cellcolor[HTML]{EFEFEF}+MedBN} & \cellcolor[HTML]{EFEFEF}\textbf{26.00} & \cellcolor[HTML]{EFEFEF}\textbf{24.00} & \cellcolor[HTML]{EFEFEF}\textbf{22.00} & \cellcolor[HTML]{EFEFEF}\textbf{16.00} & \cellcolor[HTML]{EFEFEF}\textbf{28.67} & \cellcolor[HTML]{EFEFEF}\textbf{16.00} & \cellcolor[HTML]{EFEFEF}\textbf{15.33} & \cellcolor[HTML]{EFEFEF}\textbf{20.67} & \cellcolor[HTML]{EFEFEF}\textbf{16.00} & \cellcolor[HTML]{EFEFEF}\textbf{16.67} & \cellcolor[HTML]{EFEFEF}\textbf{12.00} & \cellcolor[HTML]{EFEFEF}\textbf{17.33} & \cellcolor[HTML]{EFEFEF}\textbf{20.67} & \cellcolor[HTML]{EFEFEF}\textbf{10.67} & \cellcolor[HTML]{EFEFEF}\textbf{26.00} & \cellcolor[HTML]{EFEFEF}\textbf{19.20} \\
& \multicolumn{1}{c|}{TENT} & 70.67 & 76.67 & 76.67 & 66.67 & 84.67 & 72.00 & 70.67 & 70.00 & 72.00 & 76.00 & 62.00 & 84.67 & 81.33 & 74.00 & 74.67 & 74.18 \\
& \multicolumn{1}{c|}{\cellcolor[HTML]{EFEFEF}+MedBN} & \cellcolor[HTML]{EFEFEF}\textbf{22.67} & \cellcolor[HTML]{EFEFEF}\textbf{24.67} & \cellcolor[HTML]{EFEFEF}\textbf{19.33} & \cellcolor[HTML]{EFEFEF}\textbf{15.33} & \cellcolor[HTML]{EFEFEF}\textbf{27.33} & \cellcolor[HTML]{EFEFEF}\textbf{15.33} & \cellcolor[HTML]{EFEFEF}\textbf{16.67} & \cellcolor[HTML]{EFEFEF}\textbf{20.00} & \cellcolor[HTML]{EFEFEF}\textbf{14.67} & \cellcolor[HTML]{EFEFEF}\textbf{18.00} & \cellcolor[HTML]{EFEFEF}\textbf{12.67} & \cellcolor[HTML]{EFEFEF}\textbf{19.33} & \cellcolor[HTML]{EFEFEF}\textbf{20.00} & \cellcolor[HTML]{EFEFEF}\textbf{14.00} & \cellcolor[HTML]{EFEFEF}\textbf{22.00} & \cellcolor[HTML]{EFEFEF}\textbf{18.80} \\
& \multicolumn{1}{c|}{EATA} & 70.67 & 80.00 & 81.33 & 74.00 & 85.33 & 74.67 & 74.67 & 72.00 & 70.67 & 77.33 & 67.33 & 81.33 & 80.00 & 68.00 & 78.67 & 75.73 \\
& \multicolumn{1}{c|}{\cellcolor[HTML]{EFEFEF}+MedBN} & \cellcolor[HTML]{EFEFEF}\textbf{26.67} & \cellcolor[HTML]{EFEFEF}\textbf{23.33} & \cellcolor[HTML]{EFEFEF}\textbf{20.67} & \cellcolor[HTML]{EFEFEF}\textbf{18.00} & \cellcolor[HTML]{EFEFEF}\textbf{30.00} & \cellcolor[HTML]{EFEFEF}\textbf{22.00} & \cellcolor[HTML]{EFEFEF}\textbf{19.33} & \cellcolor[HTML]{EFEFEF}\textbf{18.00} & \cellcolor[HTML]{EFEFEF}\textbf{15.33} & \cellcolor[HTML]{EFEFEF}\textbf{20.00} & \cellcolor[HTML]{EFEFEF}\textbf{16.67} & \cellcolor[HTML]{EFEFEF}\textbf{20.00} & \cellcolor[HTML]{EFEFEF}\textbf{24.67} & \cellcolor[HTML]{EFEFEF}\textbf{18.00} & \cellcolor[HTML]{EFEFEF}\textbf{22.67} & \cellcolor[HTML]{EFEFEF}\textbf{21.02} \\
& \multicolumn{1}{c|}{SAR} & 72.67 & 78.00 & 80.00 & 70.00 & 82.00 & 74.00 & 78.00 & 72.67 & 76.00 & 83.33 & 68.67 & 86.00 & 84.67 & 73.33 & 72.67 & 76.80 \\
& \multicolumn{1}{c|}{\cellcolor[HTML]{EFEFEF}+MedBN} & \cellcolor[HTML]{EFEFEF}\textbf{25.33} & \cellcolor[HTML]{EFEFEF}\textbf{23.33} & \cellcolor[HTML]{EFEFEF}\textbf{20.00} & \cellcolor[HTML]{EFEFEF}\textbf{15.33} & \cellcolor[HTML]{EFEFEF}\textbf{26.00} & \cellcolor[HTML]{EFEFEF}\textbf{16.67} & \cellcolor[HTML]{EFEFEF}\textbf{10.00} & \cellcolor[HTML]{EFEFEF}\textbf{20.00} & \cellcolor[HTML]{EFEFEF}\textbf{14.67} & \cellcolor[HTML]{EFEFEF}\textbf{14.67} & \cellcolor[HTML]{EFEFEF}\textbf{15.33} & \cellcolor[HTML]{EFEFEF}\textbf{18.00} & \cellcolor[HTML]{EFEFEF}\textbf{22.67} & \cellcolor[HTML]{EFEFEF}\textbf{15.33} & \cellcolor[HTML]{EFEFEF}\textbf{24.67} & \cellcolor[HTML]{EFEFEF}\textbf{18.80} \\
& \multicolumn{1}{c|}{SoTTA} & 24.67 & 22.67 & 26.00 & 16.00 & 24.67 & 18.67 & 18.00 & 23.33 & 16.00 & 20.67 & 16.67 & 18.67 & 24.67 & 22.67 & 24.00 & 21.16 \\
& \multicolumn{1}{c|}{\cellcolor[HTML]{EFEFEF}+MedBN} & \cellcolor[HTML]{EFEFEF}\textbf{10.67} & \cellcolor[HTML]{EFEFEF}\textbf{16.00} & \cellcolor[HTML]{EFEFEF}\textbf{12.00} & \cellcolor[HTML]{EFEFEF}\textbf{6.67} & \cellcolor[HTML]{EFEFEF}\textbf{8.00} & \cellcolor[HTML]{EFEFEF}\textbf{3.33} & \cellcolor[HTML]{EFEFEF}\textbf{6.00} & \cellcolor[HTML]{EFEFEF}\textbf{11.33} & \cellcolor[HTML]{EFEFEF}\textbf{6.00} & \cellcolor[HTML]{EFEFEF}\textbf{5.33} & \cellcolor[HTML]{EFEFEF}\textbf{6.00} & \cellcolor[HTML]{EFEFEF}\textbf{8.00} & \cellcolor[HTML]{EFEFEF}\textbf{10.67} & \cellcolor[HTML]{EFEFEF}\textbf{7.33} & \cellcolor[HTML]{EFEFEF}\textbf{14.00} & \cellcolor[HTML]{EFEFEF}\textbf{8.76} \\
& \multicolumn{1}{c|}{sEMA} & 26.67 & 23.33 & 18.67 & 12.00 & 26.67 & 14.00 & 12.67 & 12.67 & 10.00 & 12.00 & 8.67 & 14.67 & 18.00 & 10.67 & 21.33 & 16.13 \\
& \multicolumn{1}{c|}{\cellcolor[HTML]{EFEFEF}+MedBN} & \cellcolor[HTML]{EFEFEF}\textbf{11.33} & \cellcolor[HTML]{EFEFEF}\textbf{14.00} & \cellcolor[HTML]{EFEFEF}\textbf{14.00} & \cellcolor[HTML]{EFEFEF}\textbf{2.00} & \cellcolor[HTML]{EFEFEF}\textbf{13.33} & \cellcolor[HTML]{EFEFEF}\textbf{5.33} & \cellcolor[HTML]{EFEFEF}\textbf{4.00} & \cellcolor[HTML]{EFEFEF}\textbf{8.00} & \cellcolor[HTML]{EFEFEF}\textbf{6.00} & \cellcolor[HTML]{EFEFEF}\textbf{8.00} & \cellcolor[HTML]{EFEFEF}\textbf{4.00} & \cellcolor[HTML]{EFEFEF}\textbf{8.00} & \cellcolor[HTML]{EFEFEF}\textbf{8.00} & \cellcolor[HTML]{EFEFEF}\textbf{2.00} & \cellcolor[HTML]{EFEFEF}\textbf{14.00} & \cellcolor[HTML]{EFEFEF}\textbf{8.13} \\
& \multicolumn{1}{c|}{mDIA} & 44.00 & 34.67 & 52.67 & 24.00 & 52.00 & 30.00 & 26.00 & 24.67 & 20.00 & 34.67 & 22.67 & 36.00 & 42.00 & 34.00 & 34.00 & 34.09 \\
& \multicolumn{1}{c|}{\cellcolor[HTML]{EFEFEF}+MedBN} & \cellcolor[HTML]{EFEFEF}\textbf{12.00} & \cellcolor[HTML]{EFEFEF}\textbf{14.00} & \cellcolor[HTML]{EFEFEF}\textbf{16.00} & \cellcolor[HTML]{EFEFEF}\textbf{2.00} & \cellcolor[HTML]{EFEFEF}\textbf{10.00} & \cellcolor[HTML]{EFEFEF}\textbf{6.00} & \cellcolor[HTML]{EFEFEF}\textbf{4.00} & \cellcolor[HTML]{EFEFEF}\textbf{8.00} & \cellcolor[HTML]{EFEFEF}\textbf{4.00} & \cellcolor[HTML]{EFEFEF}\textbf{10.00} & \cellcolor[HTML]{EFEFEF}\textbf{4.00} & \cellcolor[HTML]{EFEFEF}\textbf{8.00} & \cellcolor[HTML]{EFEFEF}\textbf{12.00} & \cellcolor[HTML]{EFEFEF}\textbf{4.00} & \cellcolor[HTML]{EFEFEF}\textbf{19.33} & \cellcolor[HTML]{EFEFEF}\textbf{8.89} \\

\midrule
\multirow{14}{*}{\rotatebox[origin=c]{90}{CIFAR100-C}} 
& \multicolumn{1}{c|}{TeBN} & 96.00 & 96.00 & 98.00 & 76.67 & 90.00 & 87.33 & 84.67 & 98.67 & 97.33 & 99.33 & 93.33 & 98.00 & 88.00 & 84.00 & 83.33 & 91.38 \\
& \multicolumn{1}{c|}{\cellcolor[HTML]{EFEFEF}+MedBN} & \cellcolor[HTML]{EFEFEF}\textbf{2.00} & \cellcolor[HTML]{EFEFEF}\textbf{2.00} & \cellcolor[HTML]{EFEFEF}\textbf{2.00} & \cellcolor[HTML]{EFEFEF}\textbf{2.00} & \cellcolor[HTML]{EFEFEF}\textbf{6.00} & \cellcolor[HTML]{EFEFEF}\textbf{2.00} & \cellcolor[HTML]{EFEFEF}\textbf{2.00} & \cellcolor[HTML]{EFEFEF}\textbf{2.00} & \cellcolor[HTML]{EFEFEF}\textbf{2.00} & \cellcolor[HTML]{EFEFEF}\textbf{4.00} & \cellcolor[HTML]{EFEFEF}\textbf{4.00} & \cellcolor[HTML]{EFEFEF}\textbf{2.67} & \cellcolor[HTML]{EFEFEF}\textbf{4.00} & \cellcolor[HTML]{EFEFEF}\textbf{2.00} & \cellcolor[HTML]{EFEFEF}\textbf{2.00} & \cellcolor[HTML]{EFEFEF}\textbf{2.71} \\
& \multicolumn{1}{c|}{TENT}
 & 71.33 & 75.33 & 68.67 & 74.00 & 72.00 & 79.33 & 68.00 & 77.33 & 77.33 & 78.67 & 74.00 & 90.00 & 74.00 & 72.67 & 63.33 & 74.40 \\
& \multicolumn{1}{c|}{\cellcolor[HTML]{EFEFEF}+MedBN} & \cellcolor[HTML]{EFEFEF}\textbf{2.00} & \cellcolor[HTML]{EFEFEF}\textbf{2.00} & \cellcolor[HTML]{EFEFEF}\textbf{2.67} & \cellcolor[HTML]{EFEFEF}\textbf{4.67} & \cellcolor[HTML]{EFEFEF}\textbf{4.00} & \cellcolor[HTML]{EFEFEF}\textbf{4.00} & \cellcolor[HTML]{EFEFEF}\textbf{4.67} & \cellcolor[HTML]{EFEFEF}\textbf{3.33} & \cellcolor[HTML]{EFEFEF}\textbf{2.00} & \cellcolor[HTML]{EFEFEF}\textbf{4.67} & \cellcolor[HTML]{EFEFEF}\textbf{4.67} & \cellcolor[HTML]{EFEFEF}\textbf{4.00} & \cellcolor[HTML]{EFEFEF}\textbf{4.67} & \cellcolor[HTML]{EFEFEF}\textbf{2.67} & \cellcolor[HTML]{EFEFEF}\textbf{2.00} & \cellcolor[HTML]{EFEFEF}\textbf{3.47} \\
& \multicolumn{1}{c|}{EATA} & 81.33 & 78.00 & 80.00 & 72.67 & 75.33 & 73.33 & 66.00 & 78.67 & 82.00 & 83.33 & 68.67 & 89.33 & 70.00 & 68.67 & 63.33 & 75.38 \\
& \multicolumn{1}{c|}{\cellcolor[HTML]{EFEFEF}+MedBN} & \cellcolor[HTML]{EFEFEF}\textbf{2.67} & \cellcolor[HTML]{EFEFEF}\textbf{2.00} & \cellcolor[HTML]{EFEFEF}\textbf{0.00} & \cellcolor[HTML]{EFEFEF}\textbf{3.33} & \cellcolor[HTML]{EFEFEF}\textbf{4.67} & \cellcolor[HTML]{EFEFEF}\textbf{2.00} & \cellcolor[HTML]{EFEFEF}\textbf{4.00} & \cellcolor[HTML]{EFEFEF}\textbf{4.00} & \cellcolor[HTML]{EFEFEF}\textbf{1.33} & \cellcolor[HTML]{EFEFEF}\textbf{2.00} & \cellcolor[HTML]{EFEFEF}\textbf{3.33} & \cellcolor[HTML]{EFEFEF}\textbf{3.33} & \cellcolor[HTML]{EFEFEF}\textbf{3.33} & \cellcolor[HTML]{EFEFEF}\textbf{4.00} & \cellcolor[HTML]{EFEFEF}\textbf{1.33} & \cellcolor[HTML]{EFEFEF}\textbf{2.76} \\
& \multicolumn{1}{c|}{SAR} & 84.00 & 86.00 & 86.00 & 82.67 & 75.33 & 78.00 & 73.33 & 84.67 & 86.00 & 92.67 & 75.33 & 92.67 & 77.33 & 81.33 & 72.67 & 81.87 \\
& \multicolumn{1}{c|}{\cellcolor[HTML]{EFEFEF}+MedBN} & \cellcolor[HTML]{EFEFEF}\textbf{2.67} & \cellcolor[HTML]{EFEFEF}\textbf{3.33} & \cellcolor[HTML]{EFEFEF}\textbf{2.00} & \cellcolor[HTML]{EFEFEF}\textbf{5.33} & \cellcolor[HTML]{EFEFEF}\textbf{5.33} & \cellcolor[HTML]{EFEFEF}\textbf{2.67} & \cellcolor[HTML]{EFEFEF}\textbf{3.33} & \cellcolor[HTML]{EFEFEF}\textbf{1.33} & \cellcolor[HTML]{EFEFEF}\textbf{2.00} & \cellcolor[HTML]{EFEFEF}\textbf{6.00} & \cellcolor[HTML]{EFEFEF}\textbf{3.33} & \cellcolor[HTML]{EFEFEF}\textbf{2.67} & \cellcolor[HTML]{EFEFEF}\textbf{3.33} & \cellcolor[HTML]{EFEFEF}\textbf{3.33} & \cellcolor[HTML]{EFEFEF}\textbf{2.00} & \cellcolor[HTML]{EFEFEF}\textbf{3.24} \\
& \multicolumn{1}{c|}{SoTTA} & 6.67 & 8.67 & 10.00 & 6.67 & 10.67 & 6.00 & 6.67 & 6.67 & 9.33 & 11.33 & 4.00 & 9.33 & 8.00 & 4.67 & 6.00 & 7.64 \\
& \multicolumn{1}{c|}{\cellcolor[HTML]{EFEFEF}+MedBN} & \cellcolor[HTML]{EFEFEF}\textbf{2.00} & \cellcolor[HTML]{EFEFEF}\textbf{2.67} & \cellcolor[HTML]{EFEFEF}\textbf{2.00} & \cellcolor[HTML]{EFEFEF}\textbf{2.67} & \cellcolor[HTML]{EFEFEF}\textbf{4.00} & \cellcolor[HTML]{EFEFEF}\textbf{4.00} & \cellcolor[HTML]{EFEFEF}\textbf{2.00} & \cellcolor[HTML]{EFEFEF}\textbf{2.67} & \cellcolor[HTML]{EFEFEF}\textbf{1.33} & \cellcolor[HTML]{EFEFEF}\textbf{2.00} & \cellcolor[HTML]{EFEFEF}\textbf{3.33} & \cellcolor[HTML]{EFEFEF}\textbf{3.33} & \cellcolor[HTML]{EFEFEF}\textbf{4.67} & \cellcolor[HTML]{EFEFEF}\textbf{2.00} & \cellcolor[HTML]{EFEFEF}\textbf{2.00} & \cellcolor[HTML]{EFEFEF}\textbf{2.71} \\
& \multicolumn{1}{c|}{sEMA} & 10.67 & 14.67 & 10.00 & 4.00 & 10.00 & 7.33 & 6.00 & 6.00 & 8.67 & 8.00 & 6.67 & 6.67 & 10.67 & 6.00 & 8.67 & 8.27 \\
& \multicolumn{1}{c|}{\cellcolor[HTML]{EFEFEF}+MedBN} & \cellcolor[HTML]{EFEFEF}\textbf{2.00} & \cellcolor[HTML]{EFEFEF}\textbf{2.00} & \cellcolor[HTML]{EFEFEF}\textbf{2.00} & \cellcolor[HTML]{EFEFEF}\textbf{0.67} & \cellcolor[HTML]{EFEFEF}\textbf{2.00} & \cellcolor[HTML]{EFEFEF}\textbf{0.00} & \cellcolor[HTML]{EFEFEF}\textbf{0.00} & \cellcolor[HTML]{EFEFEF}\textbf{0.00} & \cellcolor[HTML]{EFEFEF}\textbf{2.00} & \cellcolor[HTML]{EFEFEF}\textbf{2.00} & \cellcolor[HTML]{EFEFEF}\textbf{2.00} & \cellcolor[HTML]{EFEFEF}\textbf{2.00} & \cellcolor[HTML]{EFEFEF}\textbf{2.00} & \cellcolor[HTML]{EFEFEF}\textbf{0.00} & \cellcolor[HTML]{EFEFEF}\textbf{0.00} & \cellcolor[HTML]{EFEFEF}\textbf{1.24} \\
& \multicolumn{1}{c|}{mDIA} & 15.33 & 18.00 & 22.00 & 13.33 & 14.00 & 17.33 & 12.00 & 16.00 & 18.00 & 24.00 & 10.00 & 24.00 & 16.00 & 12.00 & 17.33 & 16.62 \\
& \multicolumn{1}{c|}{\cellcolor[HTML]{EFEFEF}+MedBN} & \cellcolor[HTML]{EFEFEF}\textbf{2.00} & \cellcolor[HTML]{EFEFEF}\textbf{4.00} & \cellcolor[HTML]{EFEFEF}\textbf{6.00} & \cellcolor[HTML]{EFEFEF}\textbf{2.00} & \cellcolor[HTML]{EFEFEF}\textbf{4.00} & \cellcolor[HTML]{EFEFEF}\textbf{0.00} & \cellcolor[HTML]{EFEFEF}\textbf{2.00} & \cellcolor[HTML]{EFEFEF}\textbf{3.33} & \cellcolor[HTML]{EFEFEF}\textbf{0.00} & \cellcolor[HTML]{EFEFEF}\textbf{0.00} & \cellcolor[HTML]{EFEFEF}\textbf{4.00} & \cellcolor[HTML]{EFEFEF}\textbf{4.00} & \cellcolor[HTML]{EFEFEF}\textbf{2.00} & \cellcolor[HTML]{EFEFEF}\textbf{0.00} & \cellcolor[HTML]{EFEFEF}\textbf{0.00} & \cellcolor[HTML]{EFEFEF}\textbf{2.22} \\

\midrule
\multirow{14}{*}{\rotatebox[origin=c]{90}{ImageNet-C}}

& \multicolumn{1}{c|}{TeBN} & 100.00 & 100.00 & 100.00 & 100.00 & 100.00 & 100.00 & 96.00 & 96.00 & 94.67 & 100.00 & 97.33 & 100.00 & 100.00 & 88.00 & 92.00 & 97.60 \\
& \multicolumn{1}{c|}{\cellcolor[HTML]{EFEFEF}+MedBN} & \cellcolor[HTML]{EFEFEF}\textbf{0.00} & \cellcolor[HTML]{EFEFEF}\textbf{0.00} & \cellcolor[HTML]{EFEFEF}\textbf{0.00} & \cellcolor[HTML]{EFEFEF}\textbf{0.00} & \cellcolor[HTML]{EFEFEF}\textbf{0.00} & \cellcolor[HTML]{EFEFEF}\textbf{0.00} & \cellcolor[HTML]{EFEFEF}\textbf{0.00} & \cellcolor[HTML]{EFEFEF}\textbf{0.00} & \cellcolor[HTML]{EFEFEF}\textbf{0.00} & \cellcolor[HTML]{EFEFEF}\textbf{0.00} & \cellcolor[HTML]{EFEFEF}\textbf{0.00} & \cellcolor[HTML]{EFEFEF}\textbf{4.00} & \cellcolor[HTML]{EFEFEF}\textbf{0.00} & \cellcolor[HTML]{EFEFEF}\textbf{0.00} & \cellcolor[HTML]{EFEFEF}\textbf{0.00} & \cellcolor[HTML]{EFEFEF}\textbf{0.27} \\
& \multicolumn{1}{c|}{TENT} & 84.00 & 78.67 & 81.33 & 88.00 & 84.00 & 84.00 & 85.33 & 81.33 & 84.00 & 94.67 & 84.00 & 97.33 & 94.67 & 60.00 & 73.33 & 83.64 \\
& \multicolumn{1}{c|}{\cellcolor[HTML]{EFEFEF}+MedBN} & \cellcolor[HTML]{EFEFEF}\textbf{0.00} & \cellcolor[HTML]{EFEFEF}\textbf{0.00} & \cellcolor[HTML]{EFEFEF}\textbf{0.00} & \cellcolor[HTML]{EFEFEF}\textbf{0.00} & \cellcolor[HTML]{EFEFEF}\textbf{1.33} & \cellcolor[HTML]{EFEFEF}\textbf{0.00} & \cellcolor[HTML]{EFEFEF}\textbf{1.33} & \cellcolor[HTML]{EFEFEF}\textbf{0.00} & \cellcolor[HTML]{EFEFEF}\textbf{0.00} & \cellcolor[HTML]{EFEFEF}\textbf{0.00} & \cellcolor[HTML]{EFEFEF}\textbf{0.00} & \cellcolor[HTML]{EFEFEF}\textbf{4.00} & \cellcolor[HTML]{EFEFEF}\textbf{0.00} & \cellcolor[HTML]{EFEFEF}\textbf{0.00} & \cellcolor[HTML]{EFEFEF}\textbf{0.00} & \cellcolor[HTML]{EFEFEF}\textbf{0.44} \\
& \multicolumn{1}{c|}{EATA} & 100.00 & 98.67 & 100.00 & 96.00 & 98.67 & 92.00 & 89.33 & 93.33 & 88.00 & 93.33 & 88.00 & 100.00 & 96.00 & 69.33 & 82.67 & 92.36 \\
& \multicolumn{1}{c|}{\cellcolor[HTML]{EFEFEF}+MedBN} & \cellcolor[HTML]{EFEFEF}\textbf{0.00} & \cellcolor[HTML]{EFEFEF}\textbf{0.00} & \cellcolor[HTML]{EFEFEF}\textbf{0.00} & \cellcolor[HTML]{EFEFEF}\textbf{0.00} & \cellcolor[HTML]{EFEFEF}\textbf{1.33} & \cellcolor[HTML]{EFEFEF}\textbf{0.00} & \cellcolor[HTML]{EFEFEF}\textbf{0.00} & \cellcolor[HTML]{EFEFEF}\textbf{0.00} & \cellcolor[HTML]{EFEFEF}\textbf{1.33} & \cellcolor[HTML]{EFEFEF}\textbf{0.00} & \cellcolor[HTML]{EFEFEF}\textbf{0.00} & \cellcolor[HTML]{EFEFEF}\textbf{4.00} & \cellcolor[HTML]{EFEFEF}\textbf{0.00} & \cellcolor[HTML]{EFEFEF}\textbf{0.00} & \cellcolor[HTML]{EFEFEF}\textbf{0.00} & \cellcolor[HTML]{EFEFEF}\textbf{0.44} \\
& \multicolumn{1}{c|}{SAR} & 100.00 & 100.00 & 100.00 & 100.00 & 100.00 & 100.00 & 96.00 & 97.33 & 92.00 & 100.00 & 96.00 & 100.00 & 100.00 & 80.00 & 92.00 & 96.89 \\
& \multicolumn{1}{c|}{\cellcolor[HTML]{EFEFEF}+MedBN} & \cellcolor[HTML]{EFEFEF}\textbf{0.00} & \cellcolor[HTML]{EFEFEF}\textbf{0.00} & \cellcolor[HTML]{EFEFEF}\textbf{0.00} & \cellcolor[HTML]{EFEFEF}\textbf{0.00} & \cellcolor[HTML]{EFEFEF}\textbf{0.00} & \cellcolor[HTML]{EFEFEF}\textbf{0.00} & \cellcolor[HTML]{EFEFEF}\textbf{0.00} & \cellcolor[HTML]{EFEFEF}\textbf{0.00} & \cellcolor[HTML]{EFEFEF}\textbf{0.00} & \cellcolor[HTML]{EFEFEF}\textbf{0.00} & \cellcolor[HTML]{EFEFEF}\textbf{0.00} & \cellcolor[HTML]{EFEFEF}\textbf{0.00} & \cellcolor[HTML]{EFEFEF}\textbf{0.00} & \cellcolor[HTML]{EFEFEF}\textbf{0.00} & \cellcolor[HTML]{EFEFEF}\textbf{0.00} & \cellcolor[HTML]{EFEFEF}\textbf{0.00} \\
& \multicolumn{1}{c|}{SoTTA} & 5.33 & 8.00 & 6.67 & 21.33 & 22.67 & 18.67 & 17.33 & 13.33 & 28.00 & 16.00 & 12.00 & 21.33 & 10.67 & 21.33 & 13.33 & 15.73 \\
& \multicolumn{1}{c|}{\cellcolor[HTML]{EFEFEF}+MedBN} & \cellcolor[HTML]{EFEFEF}\textbf{0.00} & \cellcolor[HTML]{EFEFEF}\textbf{0.00} & \cellcolor[HTML]{EFEFEF}\textbf{0.00} & \cellcolor[HTML]{EFEFEF}\textbf{2.67} & \cellcolor[HTML]{EFEFEF}\textbf{0.00} & \cellcolor[HTML]{EFEFEF}\textbf{0.00} & \cellcolor[HTML]{EFEFEF}\textbf{1.33} & \cellcolor[HTML]{EFEFEF}\textbf{2.67} & \cellcolor[HTML]{EFEFEF}\textbf{2.67} & \cellcolor[HTML]{EFEFEF}\textbf{0.00} & \cellcolor[HTML]{EFEFEF}\textbf{0.00} & \cellcolor[HTML]{EFEFEF}\textbf{2.67} & \cellcolor[HTML]{EFEFEF}\textbf{0.00} & \cellcolor[HTML]{EFEFEF}\textbf{0.00} & \cellcolor[HTML]{EFEFEF}\textbf{0.00} & \cellcolor[HTML]{EFEFEF}\textbf{0.80} \\
& \multicolumn{1}{c|}{sEMA} & 8.00 & 17.33 & 12.00 & 18.67 & 8.00 & 21.33 & 16.00 & 8.00 & 16.00 & 24.00 & 4.00 & 8.00 & 8.00 & 13.33 & 14.67 & 13.16 \\
& \multicolumn{1}{c|}{\cellcolor[HTML]{EFEFEF}+MedBN} & \cellcolor[HTML]{EFEFEF}\textbf{0.00} & \cellcolor[HTML]{EFEFEF}\textbf{0.00} & \cellcolor[HTML]{EFEFEF}\textbf{0.00} & \cellcolor[HTML]{EFEFEF}\textbf{0.00} & \cellcolor[HTML]{EFEFEF}\textbf{0.00} & \cellcolor[HTML]{EFEFEF}\textbf{0.00} & \cellcolor[HTML]{EFEFEF}\textbf{0.00} & \cellcolor[HTML]{EFEFEF}\textbf{0.00} & \cellcolor[HTML]{EFEFEF}\textbf{0.00} & \cellcolor[HTML]{EFEFEF}\textbf{0.00} & \cellcolor[HTML]{EFEFEF}\textbf{0.00} & \cellcolor[HTML]{EFEFEF}\textbf{0.00} & \cellcolor[HTML]{EFEFEF}\textbf{0.00} & \cellcolor[HTML]{EFEFEF}\textbf{0.00} & \cellcolor[HTML]{EFEFEF}\textbf{0.00} & \cellcolor[HTML]{EFEFEF}\textbf{0.00} \\
& \multicolumn{1}{c|}{mDIA} & 33.33 & 26.67 & 40.00 & 32.00 & 33.33 & 33.33 & 40.00 & 16.00 & 44.00 & 38.67 & 20.00 & 57.33 & 21.33 & 17.33 & 20.00 & 31.56 \\
& \multicolumn{1}{c|}{\cellcolor[HTML]{EFEFEF}+MedBN} & \cellcolor[HTML]{EFEFEF}\textbf{8.00} & \cellcolor[HTML]{EFEFEF}\textbf{0.00} & \cellcolor[HTML]{EFEFEF}\textbf{0.00} & \cellcolor[HTML]{EFEFEF}\textbf{0.00} & \cellcolor[HTML]{EFEFEF}\textbf{0.00} & \cellcolor[HTML]{EFEFEF}\textbf{0.00} & \cellcolor[HTML]{EFEFEF}\textbf{0.00} & \cellcolor[HTML]{EFEFEF}\textbf{0.00} & \cellcolor[HTML]{EFEFEF}\textbf{4.00} & \cellcolor[HTML]{EFEFEF}\textbf{0.00} & \cellcolor[HTML]{EFEFEF}\textbf{0.00} & \cellcolor[HTML]{EFEFEF}\textbf{0.00} & \cellcolor[HTML]{EFEFEF}\textbf{0.00} & \cellcolor[HTML]{EFEFEF}\textbf{4.00} & \cellcolor[HTML]{EFEFEF}\textbf{0.00} & \cellcolor[HTML]{EFEFEF}\textbf{1.07} \\
\bottomrule
\end{tabular}
\end{adjustbox}
\end{table*}

\begin{table*}[!ht]
\centering
\caption{Error Rate (\%) of the indiscriminate and cumulative attack scenario over all types of corruptions (full version of Table~\ref{tab:cum:avg}).} \label{tab:cum:indis}
\begin{adjustbox}{width=1.0\textwidth}
\begin{tabular}
{l|l|ccccccccccccccccc}
\toprule
& & \multicolumn{3}{c}{Noise} & \multicolumn{4}{c}{Blur} & \multicolumn{4}{c}{Weather} & \multicolumn{4}{c}{Digital} &     
\\
\cmidrule(lr){3-5} \cmidrule(lr){6-9} \cmidrule(lr){10-13} \cmidrule(lr){14-17}
& \multicolumn{1}{c|}{Method} & Gauss. & Shot & Impul. & Defoc. & Glass & Motion & Zoom & Snow & Frost & Fog & Brit. & Contr. & Elastic & Pixel. & JPEG & Avg. \\ 
\midrule
\multirow{14}{*}{\rotatebox[origin=c]{90}{CIFAR10-C}} 
&
\multicolumn{1}{c|}{TeBN} & 35.65 & 34.44 & 45.67 & 23.77 & 37.92 & 27.74 & 20.58 & 29.86 & 26.85 & 40.66 & 20.77 & 31.01 & 32.96 & 23.65 & 33.69 & 35.30 \\
 & \multicolumn{1}{c|}{\cellcolor[HTML]{EFEFEF}+MedBN} & \cellcolor[HTML]{EFEFEF}\textbf{26.76} & \cellcolor[HTML]{EFEFEF}\textbf{25.07} & \cellcolor[HTML]{EFEFEF}\textbf{35.02} & \cellcolor[HTML]{EFEFEF}\textbf{17.18} & \cellcolor[HTML]{EFEFEF}\textbf{27.58} & \cellcolor[HTML]{EFEFEF}\textbf{20.34} & \cellcolor[HTML]{EFEFEF}\textbf{14.70} & \cellcolor[HTML]{EFEFEF}\textbf{20.88} & \cellcolor[HTML]{EFEFEF}\textbf{18.40} & \cellcolor[HTML]{EFEFEF}\textbf{28.25} & \cellcolor[HTML]{EFEFEF}\textbf{14.17} & \cellcolor[HTML]{EFEFEF}\textbf{20.25} & \cellcolor[HTML]{EFEFEF}\textbf{25.00} & \cellcolor[HTML]{EFEFEF}\textbf{16.74} & \cellcolor[HTML]{EFEFEF}\textbf{24.76} & \cellcolor[HTML]{EFEFEF}\textbf{27.22} \\
& \multicolumn{1}{c|}{TENT} & 34.79 & 32.67 & 47.80 & 25.72 & 38.49 & 29.37 & 23.07 & 29.47 & 28.87 & 32.61 & 23.32 & 29.33 & 34.45 & 25.52 & 34.77 & 35.70 \\
& \multicolumn{1}{c|}{\cellcolor[HTML]{EFEFEF}+MedBN} & \cellcolor[HTML]{EFEFEF}\textbf{24.86} & \cellcolor[HTML]{EFEFEF}\textbf{23.29} & \cellcolor[HTML]{EFEFEF}\textbf{34.99} & \cellcolor[HTML]{EFEFEF}\textbf{15.10} & \cellcolor[HTML]{EFEFEF}\textbf{27.88} & \cellcolor[HTML]{EFEFEF}\textbf{17.85} & \cellcolor[HTML]{EFEFEF}\textbf{13.80} & \cellcolor[HTML]{EFEFEF}\textbf{18.31} & \cellcolor[HTML]{EFEFEF}\textbf{17.71} & \cellcolor[HTML]{EFEFEF}\textbf{22.18} & \cellcolor[HTML]{EFEFEF}\textbf{13.98} & \cellcolor[HTML]{EFEFEF}\textbf{19.67} & \cellcolor[HTML]{EFEFEF}\textbf{24.33} & \cellcolor[HTML]{EFEFEF}\textbf{15.46} & \cellcolor[HTML]{EFEFEF}\textbf{22.99} & \cellcolor[HTML]{EFEFEF}\textbf{25.84} \\
& \multicolumn{1}{c|}{EATA} & 35.84 & 33.09 & 46.68 & 24.61 & 39.32 & 27.88 & 22.51 & 29.11 & 27.64 & 31.70 & 23.47 & 28.50 & 35.94 & 25.10 & 34.02 & 35.30 \\
& \multicolumn{1}{c|}{\cellcolor[HTML]{EFEFEF}+MedBN} & \cellcolor[HTML]{EFEFEF}\textbf{26.52} & \cellcolor[HTML]{EFEFEF}\textbf{24.66} & \cellcolor[HTML]{EFEFEF}\textbf{35.86} & \cellcolor[HTML]{EFEFEF}\textbf{15.72} & \cellcolor[HTML]{EFEFEF}\textbf{27.53} & \cellcolor[HTML]{EFEFEF}\textbf{20.50} & \cellcolor[HTML]{EFEFEF}\textbf{17.15} & \cellcolor[HTML]{EFEFEF}\textbf{19.23} & \cellcolor[HTML]{EFEFEF}\textbf{17.31} & \cellcolor[HTML]{EFEFEF}\textbf{21.63} & \cellcolor[HTML]{EFEFEF}\textbf{14.96} & \cellcolor[HTML]{EFEFEF}\textbf{22.56} & \cellcolor[HTML]{EFEFEF}\textbf{25.34} & \cellcolor[HTML]{EFEFEF}\textbf{16.00} & \cellcolor[HTML]{EFEFEF}\textbf{24.16} & \cellcolor[HTML]{EFEFEF}\textbf{26.84} \\
&  \multicolumn{1}{c|}{SAR} & 30.83 & 29.07 & 38.36 & 21.62 & 33.22 & 24.34 & 20.32 & 25.23 & 24.15 & 27.22 & 19.87 & 25.09 & 29.88 & 21.57 & 28.77 & 31.25 \\
& \multicolumn{1}{c|}{\cellcolor[HTML]{EFEFEF}+MedBN} & \cellcolor[HTML]{EFEFEF}\textbf{23.00} & \cellcolor[HTML]{EFEFEF}\textbf{21.20} & \cellcolor[HTML]{EFEFEF}\textbf{30.92} & \cellcolor[HTML]{EFEFEF}\textbf{14.52} & \cellcolor[HTML]{EFEFEF}\textbf{24.93} & \cellcolor[HTML]{EFEFEF}\textbf{16.82} & \cellcolor[HTML]{EFEFEF}\textbf{13.27} & \cellcolor[HTML]{EFEFEF}\textbf{17.56} & \cellcolor[HTML]{EFEFEF}\textbf{16.17} & \cellcolor[HTML]{EFEFEF}\textbf{19.33} & \cellcolor[HTML]{EFEFEF}\textbf{13.10} & \cellcolor[HTML]{EFEFEF}\textbf{17.38} & \cellcolor[HTML]{EFEFEF}\textbf{22.72} & \cellcolor[HTML]{EFEFEF}\textbf{14.55} & \cellcolor[HTML]{EFEFEF}\textbf{21.88} & \cellcolor[HTML]{EFEFEF}\textbf{24.29} \\
&  \multicolumn{1}{c|}{SoTTA} & 25.35 & 24.59 & 34.17 & 15.47 & 28.35 & 17.77 & 14.32 & 19.52 & 17.75 & 21.34 & 14.69 & 19.55 & 23.92 & 16.27 & 23.71 & 26.10 \\
& \multicolumn{1}{c|}{\cellcolor[HTML]{EFEFEF}+MedBN} & \cellcolor[HTML]{EFEFEF}\textbf{22.52} & \cellcolor[HTML]{EFEFEF}\textbf{20.42} & \cellcolor[HTML]{EFEFEF}\textbf{30.98} & \cellcolor[HTML]{EFEFEF}\textbf{11.61} & \cellcolor[HTML]{EFEFEF}\textbf{23.32} & \cellcolor[HTML]{EFEFEF}\textbf{13.95} & \cellcolor[HTML]{EFEFEF}\textbf{10.94} & \cellcolor[HTML]{EFEFEF}\textbf{16.18} & \cellcolor[HTML]{EFEFEF}\textbf{14.52} & \cellcolor[HTML]{EFEFEF}\textbf{15.87} & \cellcolor[HTML]{EFEFEF}\textbf{11.04} & \cellcolor[HTML]{EFEFEF}\textbf{15.85} & \cellcolor[HTML]{EFEFEF}\textbf{19.84} & \cellcolor[HTML]{EFEFEF}\textbf{13.28} & \cellcolor[HTML]{EFEFEF}\textbf{20.00} & \cellcolor[HTML]{EFEFEF}\textbf{22.52} \\

&  \multicolumn{1}{c|}{sEMA} & 29.10 & 27.45 & 39.19 & 17.51 & 29.48 & 20.45 & 14.63 & 22.25 & 18.96 & 32.03 & 15.15 & 23.96 & 25.52 & 18.20 & 27.08 & 28.79 \\
& \multicolumn{1}{c|}{\cellcolor[HTML]{EFEFEF}+MedBN} & \cellcolor[HTML]{EFEFEF}\textbf{25.35} & \cellcolor[HTML]{EFEFEF}\textbf{23.49} & \cellcolor[HTML]{EFEFEF}\textbf{33.90} & \cellcolor[HTML]{EFEFEF}\textbf{14.97} & \cellcolor[HTML]{EFEFEF}\textbf{25.62} & \cellcolor[HTML]{EFEFEF}\textbf{18.51} & \cellcolor[HTML]{EFEFEF}\textbf{13.31} & \cellcolor[HTML]{EFEFEF}\textbf{19.12} & \cellcolor[HTML]{EFEFEF}\textbf{16.77} & \cellcolor[HTML]{EFEFEF}\textbf{26.26} & \cellcolor[HTML]{EFEFEF}\textbf{12.47} & \cellcolor[HTML]{EFEFEF}\textbf{18.20} & \cellcolor[HTML]{EFEFEF}\textbf{23.09} & \cellcolor[HTML]{EFEFEF}\textbf{14.97} & \cellcolor[HTML]{EFEFEF}\textbf{23.64} & \cellcolor[HTML]{EFEFEF}\textbf{25.62} \\
& \multicolumn{1}{c|}{mDIA} & 38.67 & 37.37 & 50.68 & 16.76 & 34.20 & 23.58 & 14.22 & 23.04 & 20.04 & 38.55 & 13.62 & 36.19 & 26.99 & 18.03 & 27.40 & 32.05 \\
& \multicolumn{1}{c|}{\cellcolor[HTML]{EFEFEF}+MedBN} & \cellcolor[HTML]{EFEFEF}\textbf{26.32} & \cellcolor[HTML]{EFEFEF}\textbf{25.20} & \cellcolor[HTML]{EFEFEF}\textbf{34.67} & \cellcolor[HTML]{EFEFEF}\textbf{12.00} & \cellcolor[HTML]{EFEFEF}\textbf{24.84} & \cellcolor[HTML]{EFEFEF}\textbf{16.00} & \cellcolor[HTML]{EFEFEF}\textbf{9.56} & \cellcolor[HTML]{EFEFEF}\textbf{16.62} & \cellcolor[HTML]{EFEFEF}\textbf{13.64} & \cellcolor[HTML]{EFEFEF}\textbf{22.11} & \cellcolor[HTML]{EFEFEF}\textbf{9.87} & \cellcolor[HTML]{EFEFEF}\textbf{20.54} & \cellcolor[HTML]{EFEFEF}\textbf{19.69} & \cellcolor[HTML]{EFEFEF}\textbf{13.24} & \cellcolor[HTML]{EFEFEF}\textbf{21.36} & \cellcolor[HTML]{EFEFEF}\textbf{23.96} \\

\midrule
\multirow{14}{*}{\rotatebox[origin=c]{90}{CIFAR100-C}}
& \multicolumn{1}{c|}{TeBN} & 59.67 & 58.49 & 61.26 & 42.36 & 55.96 & 49.35 & 39.85 & 51.24 & 48.82 & 59.46 & 45.40 & 59.62 & 50.70 & 45.82 & 52.61 & 52.04 \\
& \multicolumn{1}{c|}{\cellcolor[HTML]{EFEFEF}+MedBN} & \cellcolor[HTML]{EFEFEF}\textbf{43.64} & \cellcolor[HTML]{EFEFEF}\textbf{45.12} & \cellcolor[HTML]{EFEFEF}\textbf{52.97} & \cellcolor[HTML]{EFEFEF}\textbf{30.25} & \cellcolor[HTML]{EFEFEF}\textbf{44.35} & \cellcolor[HTML]{EFEFEF}\textbf{32.25} & \cellcolor[HTML]{EFEFEF}\textbf{28.77} & \cellcolor[HTML]{EFEFEF}\textbf{37.01} & \cellcolor[HTML]{EFEFEF}\textbf{33.22} & \cellcolor[HTML]{EFEFEF}\textbf{49.61} & \cellcolor[HTML]{EFEFEF}\textbf{31.11} & \cellcolor[HTML]{EFEFEF}\textbf{40.82} & \cellcolor[HTML]{EFEFEF}\textbf{40.94} & \cellcolor[HTML]{EFEFEF}\textbf{30.05} & \cellcolor[HTML]{EFEFEF}\textbf{38.15} & \cellcolor[HTML]{EFEFEF}\textbf{38.55} \\

& \multicolumn{1}{c|}{TENT} & 57.07 & 54.94 & 63.90 & 41.09 & 56.35 & 44.32 & 40.18 & 51.72 & 49.54 & 54.25 & 42.06 & 53.19 & 50.77 & 41.94 & 49.65 & 50.06 \\
& \multicolumn{1}{c|}{\cellcolor[HTML]{EFEFEF}+MedBN} & \cellcolor[HTML]{EFEFEF}\textbf{43.82} & \cellcolor[HTML]{EFEFEF}\textbf{40.33} & \cellcolor[HTML]{EFEFEF}\textbf{47.09} & \cellcolor[HTML]{EFEFEF}\textbf{30.12} & \cellcolor[HTML]{EFEFEF}\textbf{46.18} & \cellcolor[HTML]{EFEFEF}\textbf{33.86} & \cellcolor[HTML]{EFEFEF}\textbf{29.49} & \cellcolor[HTML]{EFEFEF}\textbf{37.36} & \cellcolor[HTML]{EFEFEF}\textbf{39.50} & \cellcolor[HTML]{EFEFEF}\textbf{42.25} & \cellcolor[HTML]{EFEFEF}\textbf{28.74} & \cellcolor[HTML]{EFEFEF}\textbf{37.73} & \cellcolor[HTML]{EFEFEF}\textbf{36.41} & \cellcolor[HTML]{EFEFEF}\textbf{33.17} & \cellcolor[HTML]{EFEFEF}\textbf{38.60} & \cellcolor[HTML]{EFEFEF}\textbf{37.64} \\
 
& \multicolumn{1}{c|}{EATA} & 52.49 & 50.88 & 60.00 & 39.29 & 53.75 & 44.07 & 39.09 & 46.35 & 47.17 & 51.90 & 39.11 & 48.40 & 48.91 & 40.27 & 51.11 & 47.52 \\
& \multicolumn{1}{c|}{\cellcolor[HTML]{EFEFEF}+MedBN} & \cellcolor[HTML]{EFEFEF}\textbf{40.62} & \cellcolor[HTML]{EFEFEF}\textbf{41.12} & \cellcolor[HTML]{EFEFEF}\textbf{48.30} & \cellcolor[HTML]{EFEFEF}\textbf{31.75} & \cellcolor[HTML]{EFEFEF}\textbf{41.37} & \cellcolor[HTML]{EFEFEF}\textbf{31.50} & \cellcolor[HTML]{EFEFEF}\textbf{26.92} & \cellcolor[HTML]{EFEFEF}\textbf{35.45} & \cellcolor[HTML]{EFEFEF}\textbf{34.81} & \cellcolor[HTML]{EFEFEF}\textbf{40.87} & \cellcolor[HTML]{EFEFEF}\textbf{28.43} & \cellcolor[HTML]{EFEFEF}\textbf{38.06} & \cellcolor[HTML]{EFEFEF}\textbf{39.45} & \cellcolor[HTML]{EFEFEF}\textbf{31.39} & \cellcolor[HTML]{EFEFEF}\textbf{40.91} & \cellcolor[HTML]{EFEFEF}\textbf{36.73} \\
& \multicolumn{1}{c|}{SAR} & 51.39 & 50.32 & 55.06 & 40.06 & 51.30 & 41.36 & 36.46 & 43.92 & 47.64 & 50.51 & 39.70 & 45.07 & 46.44 & 39.18 & 47.05 & 45.70 \\
& \multicolumn{1}{c|}{\cellcolor[HTML]{EFEFEF}+MedBN} & \cellcolor[HTML]{EFEFEF}\textbf{42.29} & \cellcolor[HTML]{EFEFEF}\textbf{41.47} & \cellcolor[HTML]{EFEFEF}\textbf{47.93} & \cellcolor[HTML]{EFEFEF}\textbf{27.83} & \cellcolor[HTML]{EFEFEF}\textbf{40.45} & \cellcolor[HTML]{EFEFEF}\textbf{33.24} & \cellcolor[HTML]{EFEFEF}\textbf{25.26} & \cellcolor[HTML]{EFEFEF}\textbf{35.55} & \cellcolor[HTML]{EFEFEF}\textbf{35.15} & \cellcolor[HTML]{EFEFEF}\textbf{42.26} & \cellcolor[HTML]{EFEFEF}\textbf{29.75} & \cellcolor[HTML]{EFEFEF}\textbf{39.21} & \cellcolor[HTML]{EFEFEF}\textbf{36.61} & \cellcolor[HTML]{EFEFEF}\textbf{29.75} & \cellcolor[HTML]{EFEFEF}\textbf{41.14} & \cellcolor[HTML]{EFEFEF}\textbf{36.53} \\
& \multicolumn{1}{c|}{SoTTA} & 44.10 & 42.92 & 50.09 & 28.53 & 43.43 & 34.78 & 29.47 & 41.13 & 38.85 & 42.48 & 27.67 & 40.34 & 38.16 & 32.75 & 38.37 & 38.21 \\
& \multicolumn{1}{c|}{\cellcolor[HTML]{EFEFEF}+MedBN} & \cellcolor[HTML]{EFEFEF}\textbf{41.07} & \cellcolor[HTML]{EFEFEF}\textbf{38.67} & \cellcolor[HTML]{EFEFEF}\textbf{45.63} & \cellcolor[HTML]{EFEFEF}\textbf{26.71} & \cellcolor[HTML]{EFEFEF}\textbf{37.17} & \cellcolor[HTML]{EFEFEF}\textbf{29.54} & \cellcolor[HTML]{EFEFEF}\textbf{25.07} & \cellcolor[HTML]{EFEFEF}\textbf{33.97} & \cellcolor[HTML]{EFEFEF}\textbf{34.90} & \cellcolor[HTML]{EFEFEF}\textbf{33.85} & \cellcolor[HTML]{EFEFEF}\textbf{27.05} & \cellcolor[HTML]{EFEFEF}\textbf{32.43} & \cellcolor[HTML]{EFEFEF}\textbf{35.52} & \cellcolor[HTML]{EFEFEF}\textbf{29.23} & \cellcolor[HTML]{EFEFEF}\textbf{36.71} & \cellcolor[HTML]{EFEFEF}\textbf{33.84} \\
& \multicolumn{1}{c|}{sEMA} & 41.95 & 44.95 & 47.57 & 28.43 & 46.33 & 35.59 & 28.93 & 36.75 & 37.20 & 46.40 & 29.37 & 42.28 & 37.73 & 28.75 & 41.43 & 38.24 \\
& \multicolumn{1}{c|}{\cellcolor[HTML]{EFEFEF}+MedBN} & \cellcolor[HTML]{EFEFEF}\textbf{36.40} & \cellcolor[HTML]{EFEFEF}\textbf{39.76} & \cellcolor[HTML]{EFEFEF}\textbf{46.11} & \cellcolor[HTML]{EFEFEF}\textbf{27.47} & \cellcolor[HTML]{EFEFEF}\textbf{40.11} & \cellcolor[HTML]{EFEFEF}\textbf{29.85} & \cellcolor[HTML]{EFEFEF}\textbf{25.90} & \cellcolor[HTML]{EFEFEF}\textbf{33.25} & \cellcolor[HTML]{EFEFEF}\textbf{31.49} & \cellcolor[HTML]{EFEFEF}\textbf{43.04} & \cellcolor[HTML]{EFEFEF}\textbf{27.56} & \cellcolor[HTML]{EFEFEF}\textbf{36.48} & \cellcolor[HTML]{EFEFEF}\textbf{35.65} & \cellcolor[HTML]{EFEFEF}\textbf{27.17} & \cellcolor[HTML]{EFEFEF}\textbf{36.34} & \cellcolor[HTML]{EFEFEF}\textbf{34.44} \\
& \multicolumn{1}{c|}{mDIA} & 52.12 & 52.81 & 60.54 & 37.40 & 50.35 & 41.74 & 29.54 & 40.60 & 39.53 & 64.64 & 29.67 & 72.37 & 43.61 & 37.01 & 41.84 & 46.25 \\
& \multicolumn{1}{c|}{\cellcolor[HTML]{EFEFEF}+MedBN} & \cellcolor[HTML]{EFEFEF}\textbf{45.33} & \cellcolor[HTML]{EFEFEF}\textbf{42.47} & \cellcolor[HTML]{EFEFEF}\textbf{53.33} & \cellcolor[HTML]{EFEFEF}\textbf{27.42} & \cellcolor[HTML]{EFEFEF}\textbf{41.30} & \cellcolor[HTML]{EFEFEF}\textbf{30.27} & \cellcolor[HTML]{EFEFEF}\textbf{25.97} & \cellcolor[HTML]{EFEFEF}\textbf{32.96} & \cellcolor[HTML]{EFEFEF}\textbf{31.94} & \cellcolor[HTML]{EFEFEF}\textbf{43.51} & \cellcolor[HTML]{EFEFEF}\textbf{24.66} & \cellcolor[HTML]{EFEFEF}\textbf{50.98} & \cellcolor[HTML]{EFEFEF}\textbf{32.61} & \cellcolor[HTML]{EFEFEF}\textbf{30.66} & \cellcolor[HTML]{EFEFEF}\textbf{35.82} & \cellcolor[HTML]{EFEFEF}\textbf{36.62} \\

\midrule
\multirow{14}{*}{\rotatebox[origin=c]{90}{ImageNet-C}}

& \multicolumn{1}{c|}{TeBN} & 97.10 & 94.53 & 96.50 & 93.42 & 93.36 & 86.07 & 76.11 & 80.95 & 83.20 & 69.33 & 50.19 & 96.64 & 69.92 & 63.12 & 71.59 & 81.47 \\
& \multicolumn{1}{c|}{\cellcolor[HTML]{EFEFEF}+MedBN} & \cellcolor[HTML]{EFEFEF}\textbf{86.18} & \cellcolor[HTML]{EFEFEF}\textbf{85.34} & \cellcolor[HTML]{EFEFEF}\textbf{85.22} & \cellcolor[HTML]{EFEFEF}\textbf{83.63} & \cellcolor[HTML]{EFEFEF}\textbf{84.72} & \cellcolor[HTML]{EFEFEF}\textbf{75.52} & \cellcolor[HTML]{EFEFEF}\textbf{64.36} & \cellcolor[HTML]{EFEFEF}\textbf{68.40} & \cellcolor[HTML]{EFEFEF}\textbf{69.96} & \cellcolor[HTML]{EFEFEF}\textbf{52.86} & \cellcolor[HTML]{EFEFEF}\textbf{37.36} & \cellcolor[HTML]{EFEFEF}\textbf{77.96} & \cellcolor[HTML]{EFEFEF}\textbf{59.88} & \cellcolor[HTML]{EFEFEF}\textbf{52.73} & \cellcolor[HTML]{EFEFEF}\textbf{61.42} & \cellcolor[HTML]{EFEFEF}\textbf{69.70} \\
& \multicolumn{1}{c|}{TENT} & 85.81 & 85.02 & 85.58 & 86.80 & 86.71 & 77.39 & 65.66 & 71.18 & 74.67 & 58.31 & 43.58 & 90.76 & 60.86 & 53.75 & 60.09 & 72.41 \\
& \multicolumn{1}{c|}{\cellcolor[HTML]{EFEFEF}+MedBN} & \cellcolor[HTML]{EFEFEF}\textbf{84.72} & \cellcolor[HTML]{EFEFEF}\textbf{83.66} & \cellcolor[HTML]{EFEFEF}\textbf{83.88} & \cellcolor[HTML]{EFEFEF}\textbf{82.21} & \cellcolor[HTML]{EFEFEF}\textbf{83.55} & \cellcolor[HTML]{EFEFEF}\textbf{74.88} & \cellcolor[HTML]{EFEFEF}\textbf{61.96} & \cellcolor[HTML]{EFEFEF}\textbf{66.28} & \cellcolor[HTML]{EFEFEF}\textbf{68.89} & \cellcolor[HTML]{EFEFEF}\textbf{50.84} & \cellcolor[HTML]{EFEFEF}\textbf{36.43} & \cellcolor[HTML]{EFEFEF}\textbf{76.70} & \cellcolor[HTML]{EFEFEF}\textbf{57.30} & \cellcolor[HTML]{EFEFEF}\textbf{50.40} & \cellcolor[HTML]{EFEFEF}\textbf{59.96} & \cellcolor[HTML]{EFEFEF}\textbf{68.11} \\
& \multicolumn{1}{c|}{EATA} & 95.64 & 92.59 & 94.10 & 90.44 & 91.20 & 80.29 & 68.27 & 74.44 & 76.41 & 61.84 & 44.06 & 91.86 & 62.01 & 54.25 & 63.00 & 76.03 \\
& \multicolumn{1}{c|}{\cellcolor[HTML]{EFEFEF}+MedBN} & \cellcolor[HTML]{EFEFEF}\textbf{85.45} & \cellcolor[HTML]{EFEFEF}\textbf{84.16} & \cellcolor[HTML]{EFEFEF}\textbf{84.89} & \cellcolor[HTML]{EFEFEF}\textbf{82.79} & \cellcolor[HTML]{EFEFEF}\textbf{83.64} & \cellcolor[HTML]{EFEFEF}\textbf{74.59} & \cellcolor[HTML]{EFEFEF}\textbf{61.71} & \cellcolor[HTML]{EFEFEF}\textbf{66.32} & \cellcolor[HTML]{EFEFEF}\textbf{68.85} & \cellcolor[HTML]{EFEFEF}\textbf{51.13} & \cellcolor[HTML]{EFEFEF}\textbf{35.81} & \cellcolor[HTML]{EFEFEF}\textbf{76.21} & \cellcolor[HTML]{EFEFEF}\textbf{57.08} & \cellcolor[HTML]{EFEFEF}\textbf{50.96} & \cellcolor[HTML]{EFEFEF}\textbf{59.61} & \cellcolor[HTML]{EFEFEF}\textbf{68.21} \\
& \multicolumn{1}{c|}{SAR} & 96.27 & 93.21 & 95.61 & 91.90 & 90.68 & 81.10 & 70.97 & 74.60 & 77.28 & 62.92 & 46.45 & 93.22 & 64.24 & 57.85 & 65.52 & 77.46 \\
& \multicolumn{1}{c|}{\cellcolor[HTML]{EFEFEF}+MedBN} & \cellcolor[HTML]{EFEFEF}\textbf{86.36} & \cellcolor[HTML]{EFEFEF}\textbf{85.60} & \cellcolor[HTML]{EFEFEF}\textbf{85.58} & \cellcolor[HTML]{EFEFEF}\textbf{83.55} & \cellcolor[HTML]{EFEFEF}\textbf{84.67} & \cellcolor[HTML]{EFEFEF}\textbf{75.50} & \cellcolor[HTML]{EFEFEF}\textbf{63.93} & \cellcolor[HTML]{EFEFEF}\textbf{68.05} & \cellcolor[HTML]{EFEFEF}\textbf{70.22} & \cellcolor[HTML]{EFEFEF}\textbf{52.67} & \cellcolor[HTML]{EFEFEF}\textbf{36.93} & \cellcolor[HTML]{EFEFEF}\textbf{78.27} & \cellcolor[HTML]{EFEFEF}\textbf{58.96} & \cellcolor[HTML]{EFEFEF}\textbf{52.05} & \cellcolor[HTML]{EFEFEF}\textbf{61.13} & \cellcolor[HTML]{EFEFEF}\textbf{69.56} \\
& \multicolumn{1}{c|}{SoTTA} & 81.20 & 80.61 & 81.10 & 81.95 & 83.69 & 71.89 & 61.56 & 65.08 & 68.78 & 53.28 & 39.38 & 77.87 & 56.13 & 50.47 & 56.33 & 67.29 \\
& \multicolumn{1}{c|}{\cellcolor[HTML]{EFEFEF}+MedBN} & \cellcolor[HTML]{EFEFEF}\textbf{82.59} & \cellcolor[HTML]{EFEFEF}\textbf{81.93} & \cellcolor[HTML]{EFEFEF}\textbf{81.49} & \cellcolor[HTML]{EFEFEF}\textbf{78.26} & \cellcolor[HTML]{EFEFEF}\textbf{79.83} & \cellcolor[HTML]{EFEFEF}\textbf{69.16} & \cellcolor[HTML]{EFEFEF}\textbf{57.51} & \cellcolor[HTML]{EFEFEF}\textbf{61.71} & \cellcolor[HTML]{EFEFEF}\textbf{64.97} & \cellcolor[HTML]{EFEFEF}\textbf{48.00} & \cellcolor[HTML]{EFEFEF}\textbf{34.70} & \cellcolor[HTML]{EFEFEF}\textbf{71.45} & \cellcolor[HTML]{EFEFEF}\textbf{53.00} & \cellcolor[HTML]{EFEFEF}\textbf{46.96} & \cellcolor[HTML]{EFEFEF}\textbf{53.94} & \cellcolor[HTML]{EFEFEF}\textbf{64.37} \\
& \multicolumn{1}{c|}{sEMA} & 88.99 & 87.84 & 88.90 & 88.24 & 88.35 & 80.55 & 70.83 & 74.97 & 76.28 & 63.30 & 44.18 & 86.87 & 65.04 & 57.49 & 66.37 & 75.21 \\
& \multicolumn{1}{c|}{\cellcolor[HTML]{EFEFEF}+MedBN} & \cellcolor[HTML]{EFEFEF}\textbf{89.27} & \cellcolor[HTML]{EFEFEF}\textbf{87.18} & \cellcolor[HTML]{EFEFEF}\textbf{87.41} & \cellcolor[HTML]{EFEFEF}\textbf{84.06} & \cellcolor[HTML]{EFEFEF}\textbf{84.59} & \cellcolor[HTML]{EFEFEF}\textbf{75.55} & \cellcolor[HTML]{EFEFEF}\textbf{64.47} & \cellcolor[HTML]{EFEFEF}\textbf{69.00} & \cellcolor[HTML]{EFEFEF}\textbf{69.98} & \cellcolor[HTML]{EFEFEF}\textbf{53.85} & \cellcolor[HTML]{EFEFEF}\textbf{37.89} & \cellcolor[HTML]{EFEFEF}\textbf{79.76} & \cellcolor[HTML]{EFEFEF}\textbf{59.85} & \cellcolor[HTML]{EFEFEF}\textbf{53.29} & \cellcolor[HTML]{EFEFEF}\textbf{61.88} & \cellcolor[HTML]{EFEFEF}\textbf{70.54} \\
& \multicolumn{1}{c|}{mDIA} & 93.83 & 92.73 & 93.24 & 87.04 & 87.59 & 80.50 & 73.73 & 77.39 & 76.62 & 67.27 & 43.72 & 90.02 & 65.67 & 61.64 & 67.53 & 77.24 \\
& \multicolumn{1}{c|}{\cellcolor[HTML]{EFEFEF}+MedBN} & \cellcolor[HTML]{EFEFEF}\textbf{87.90} & \cellcolor[HTML]{EFEFEF}\textbf{86.92} & \cellcolor[HTML]{EFEFEF}\textbf{86.53} & \cellcolor[HTML]{EFEFEF}\textbf{79.83} & \cellcolor[HTML]{EFEFEF}\textbf{81.59} & \cellcolor[HTML]{EFEFEF}\textbf{73.44} & \cellcolor[HTML]{EFEFEF}\textbf{65.02} & \cellcolor[HTML]{EFEFEF}\textbf{68.13} & \cellcolor[HTML]{EFEFEF}\textbf{67.81} & \cellcolor[HTML]{EFEFEF}\textbf{52.41} & \cellcolor[HTML]{EFEFEF}\textbf{36.44} & \cellcolor[HTML]{EFEFEF}\textbf{78.04} & \cellcolor[HTML]{EFEFEF}\textbf{58.90} & \cellcolor[HTML]{EFEFEF}\textbf{55.15} & \cellcolor[HTML]{EFEFEF}\textbf{60.13} & \cellcolor[HTML]{EFEFEF}\textbf{69.22} \\

\bottomrule
\end{tabular}
\end{adjustbox}
\end{table*}


\clearpage

\section{Error Rates without Attacks} 
To evaluate the performance of the model under a normal TTA setup, we utilize ER on benign samples without attacks in Table~\ref{tab:ins:bf}. 
It provides an understanding of how the model behaves in a non-adversarial environment, i.e., the model's baseline effectiveness, establishing a fundamental metric for comparison against scenarios involving attacks.

\label{appendix:ins:bf-attack}

\begin{table*}[htb!]\centering
\caption{Error Rate (\%) on benign samples without attacks.} \label{tab:ins:bf}
\begin{adjustbox}{width=0.8\textwidth}
\begin{tabular}{c|c|c|ccccccc}
\toprule
\multicolumn{1}{l|}{} & \multicolumn{1}{l|}{} & \multicolumn{1}{l|}{} & \multicolumn{7}{c}{Method} \\ \cmidrule{4-10} 
\multicolumn{1}{c|}{\multirow{-2}{*}{ER (\%)}} & \multicolumn{1}{c|}{\multirow{-2}{*}{$B$ / $m$}} & \multicolumn{1}{c|}{\multirow{-2}{*}{Normalization}} & TeBN & TENT & ETA & SAR & SoTTA & sEMA & mDIA \\

\midrule
 & & BatchNorm & 14.92 & 13.68 & 13.14 & 13.28 & 13.73 & 14.87 & 15.31 \\ \cline{4-4}
\multirow{-2}{*}{CIFAR10-C} & \multirow{-2}{*}{\begin{tabular}[c]{@{}c@{}}200 / 40 \\ (20\%)\end{tabular}} & \cellcolor[HTML]{EFEFEF}Ours (MedBN) & \cellcolor[HTML]{EFEFEF}15.19 & \cellcolor[HTML]{EFEFEF}14.12 & \cellcolor[HTML]{EFEFEF}13.67 & \cellcolor[HTML]{EFEFEF}13.35 & \cellcolor[HTML]{EFEFEF}14.06 & \cellcolor[HTML]{EFEFEF}15.14 & \cellcolor[HTML]{EFEFEF}15.20 \\ \midrule
 & & BatchNorm & 40.08 & 37.74 & 37.44 & 39.30 & 41.22 & 39.72 & 41.72 \\
\multirow{-2}{*}{CIFAR100-C} & \multirow{-2}{*}{\begin{tabular}[c]{@{}c@{}}200 / 40\\ (20\%)\end{tabular}} & \cellcolor[HTML]{EFEFEF}Ours (MedBN) & \cellcolor[HTML]{EFEFEF}40.77 & \cellcolor[HTML]{EFEFEF}39.66 & \cellcolor[HTML]{EFEFEF}39.62 & \cellcolor[HTML]{EFEFEF}41.32 & \cellcolor[HTML]{EFEFEF}42.26 & \cellcolor[HTML]{EFEFEF}40.47 & \cellcolor[HTML]{EFEFEF}41.79 \\ \midrule
 & & BatchNorm & 66.62 & 61.08 & 59.13 & 62.13 & 60.87 & 68.35 & 66.62 \\
\multirow{-2}{*}{ImageNet-C}& \multirow{-2}{*}{\begin{tabular}[c]{@{}c@{}}200 / 20\\ (10\%)\end{tabular}} & \cellcolor[HTML]{EFEFEF}Ours (MedBN) & \cellcolor[HTML]{EFEFEF}69.55 & \cellcolor[HTML]{EFEFEF}68.38 & \cellcolor[HTML]{EFEFEF}66.20 & \cellcolor[HTML]{EFEFEF}66.65 & \cellcolor[HTML]{EFEFEF}64.39 & \cellcolor[HTML]{EFEFEF}70.18 & \cellcolor[HTML]{EFEFEF}68.27 \\ 
\bottomrule
\end{tabular}
\end{adjustbox}
\end{table*}

\section{Examples of Malicious Samples}\label{appendix:vis:malicious}

\begin{figure*}[!h]
    \centering
    \begin{subfigure}[b]{0.6\textwidth}
         \centering
         \includegraphics[trim={5em 9em 5em 9em}, clip, width=\textwidth]{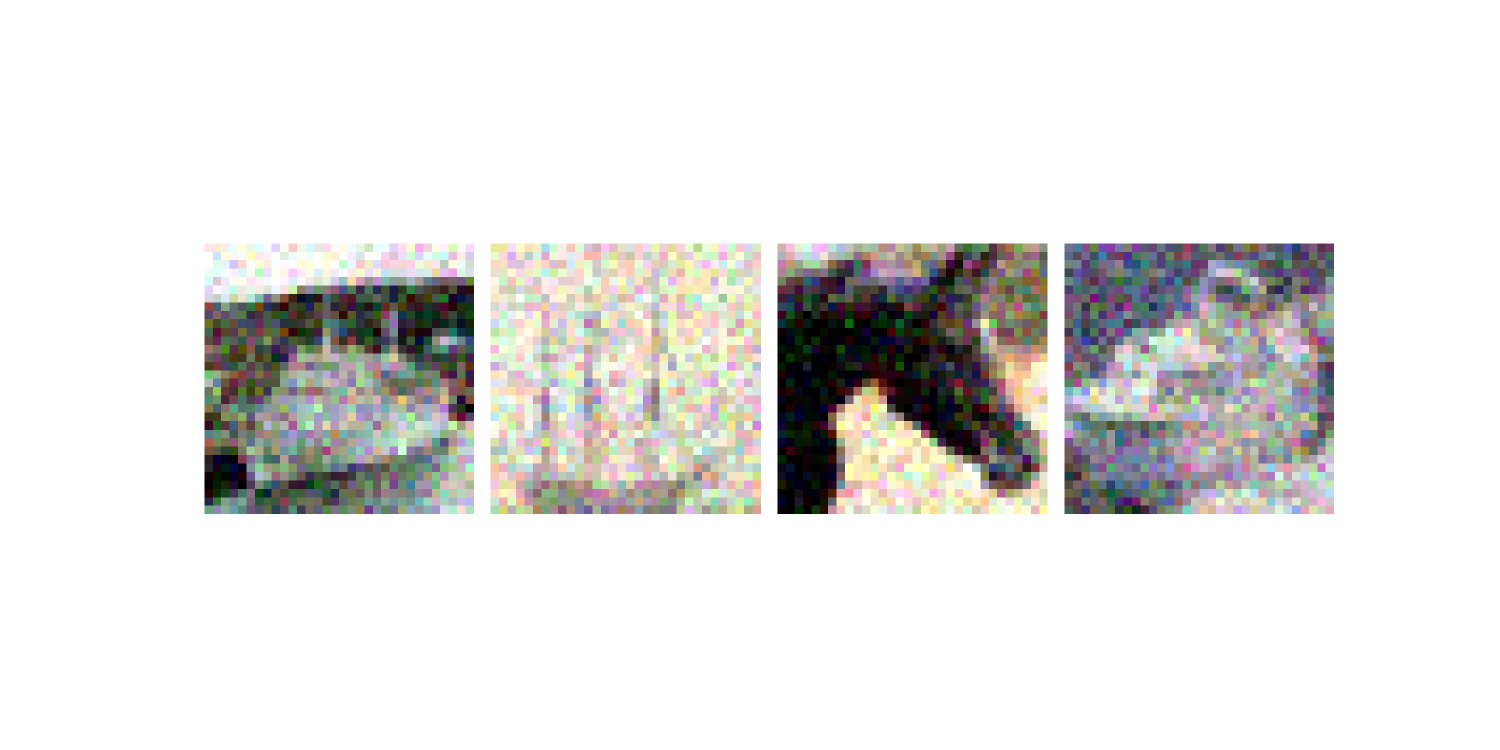}
         \caption{Visualization of benign samples.}
         \label{fig:clean_example}
     \end{subfigure}
     \begin{subfigure}[b]{0.6\textwidth}
         \centering
         \includegraphics[trim={5em 9em 5em 9em}, clip, width=\textwidth]{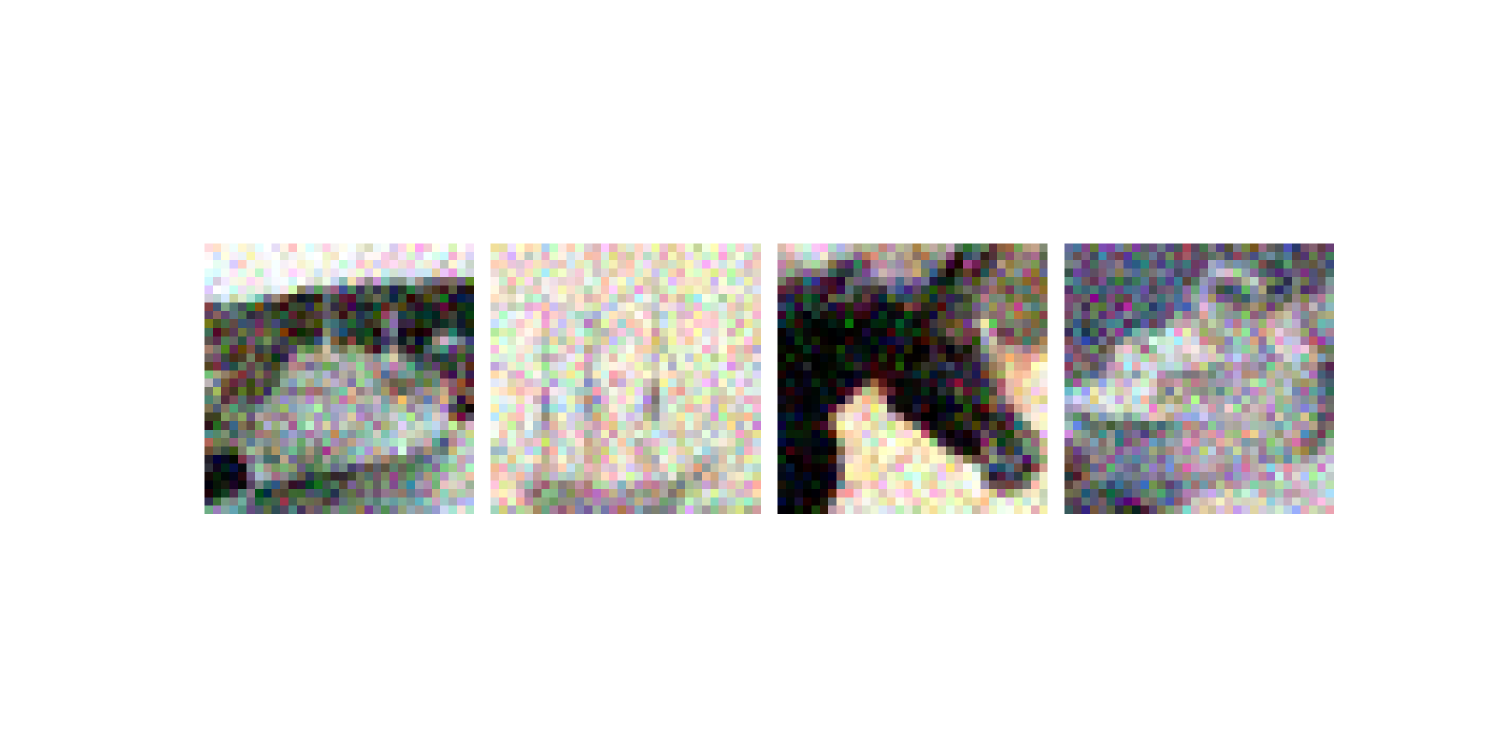}
         \caption{Visualization of malicious samples ($\varepsilon=8/255$).}
         \label{fig:adv_example}
     \end{subfigure}
     \caption{Visualization of test samples from CIFAR10-C benchmark with Gaussian noise and severity level 5. Malicious samples are hardly distinguished from benign samples.}
\end{figure*}


\end{document}